\documentclass{article}
\usepackage[dvipsnames]{xcolor}
\PassOptionsToPackage{numbers, sort&compress}{natbib}
\usepackage{natbib}
\usepackage[final]{neurips_2016}

\usepackage{graphicx}
\usepackage{booktabs}
\usepackage{microtype}
\usepackage{makecell}
\usepackage{pdflscape}
\usepackage{arydshln}
\usepackage{longtable}
\usepackage[misc]{ifsym}

\usepackage{hyperref}
\usepackage{mathtools}
\usepackage{amsmath,amsfonts,amssymb,amsthm}
\usepackage[font=small]{caption}
\usepackage[T1]{fontenc}
\usepackage{float}
\usepackage[section]{placeins}

\renewcommand{\floatpagefraction}{0.85}
\usepackage{afterpage}
\usepackage{tikz}
\usepackage{multirow}
\usepackage{array}
\usepackage{fancyhdr}
\usepackage{algorithm}
\usepackage{algorithmic}
\usepackage{bbm}
\usepackage{lineno}

\usepackage{cleveref}

\crefname{edfig}{Extended Data Fig.}{Extended Data Figs.}
\Crefname{edfig}{Extended Data Fig.}{Extended Data Figs.}
\crefname{suppfig}{Supplementary Fig.}{Supplementary Figs.}
\Crefname{suppfig}{Supplementary Fig.}{Supplementary Figs.}
\crefname{supptab}{Supplementary Table}{Supplementary Tables}
\Crefname{supptab}{Supplementary Table}{Supplementary Tables}
\crefname{suppalg}{Supplementary Algorithm}{Supplementary Algorithms}
\Crefname{suppalg}{Supplementary Algorithm}{Supplementary Algorithms}
\crefname{appendix}{Supplementary Information}{Supplementary Information}
\Crefname{appendix}{Supplementary Information}{Supplementary Information}
\crefname{subappendix}{Supplementary Section}{Supplementary Sections}
\Crefname{subappendix}{Supplementary Section}{Supplementary Sections}
\crefname{subsubappendix}{Supplementary Section}{Supplementary Sections}
\Crefname{subsubappendix}{Supplementary Section}{Supplementary Sections}

\renewcommand{\headrulewidth}{0pt}
\renewcommand{\footrulewidth}{0pt}
\fancypagestyle{plain}{\fancyhf{}\renewcommand{\headrulewidth}{0pt}%
  \renewcommand{\footrulewidth}{0pt}\fancyfoot[R]{\thepage}}
\makeatletter
\def\@notice{}
\makeatother
\title{Transcriptome-informed multi-modal AI for predicting\\neoadjuvant therapy response from breast cancer biopsies}

\newcommand{\affnum}[1]{\textsuperscript{#1}}

\author{%
  \begin{minipage}{\textwidth}
  \centering
  {\bfseries
Jungkyu Park\affnum{1},
Dhruva Biswas\affnum{1},
Joseph Cappadona\affnum{1},
Cerise Tang\affnum{1},
Ken G.~Zeng\affnum{1},
Bartosz Machura\affnum{1},
Chuwen Liu\affnum{1},
Paolo Tarantino\affnum{2,3},
Coral Omene\affnum{4},
Francisco J.~Esteva\affnum{5},
Rohit Bhargava\affnum{6,7},
Marcin Braun\affnum{8},
Kamila Paździerz\affnum{8},\\
Jakub Czerwiński\affnum{8},
Hanna Romańska-Knight\affnum{8},
Albert Grinshpun\affnum{9,10},
Bareket Daniel\affnum{9,10},
Michele Buchinger\affnum{9,10},
Frederick Howard\affnum{11},
Piotr Wysocki\affnum{12},
Brie Chun\affnum{13,14},
Freya Schnabel\affnum{15},
Rich Caruana\affnum{16},
Jan Witowski\affnum{1},\\
Krzysztof J.~Geras\affnum{1,$\dagger$}
  }\\[1.2em]
  {\normalfont\footnotesize
\affnum{1}Ataraxis AI, New York, NY, USA.\\
\affnum{2}Breast Oncology Program, Dana-Farber Cancer Institute, Boston, MA, USA.\\
\affnum{3}Harvard Medical School, Boston, MA, USA.\\
\affnum{4}Division of Medical Oncology, Rutgers Cancer Institute, New Brunswick, NJ, USA.\\
\affnum{5}Northwell Health, New York, NY, USA.\\
\affnum{6}Department of Pathology, University of Pittsburgh, Pittsburgh, PA, USA.\\
\affnum{7}Magee-Womens Hospital of UPMC, Pittsburgh, PA, USA.\\
\affnum{8}Department of Pathology, Chair of Oncology, Medical University of Łódź, Łódź, Poland.\\
\affnum{9}Shaare Zedek Medical Center, Jerusalem, Israel.\\
\affnum{10}The Hebrew University of Jerusalem, Jerusalem, Israel.\\
\affnum{11}University of Chicago, Chicago, IL, USA.\\
\affnum{12}Department of Oncology, Jagiellonian University Medical College, Kraków, Poland.\\
\affnum{13}Fred Hutchinson Cancer Center, Seattle, WA, USA.\\
\affnum{14}University of Washington, Seattle, WA, USA.\\
\affnum{15}NYU Grossman School of Medicine, New York, NY, USA.\\
\affnum{16}Intelligible, Inc., Seattle, WA, USA.\\
\affnum{$\dagger$}Correspondence to: Krzysztof J.~Geras \texttt{<krzysztof.geras@ataraxis.ai>}.
  }
\end{minipage}%
}

\date{}

\hypersetup{%
    colorlinks = true,
    citecolor  = blue,
    linkcolor  = blue,
    urlcolor   = red,
}

\begin{document}

\newgeometry{
  textheight=9in,
  textwidth=6.9in,
  top=1in,
  headheight=12pt,
  headsep=13pt,
  footskip=25pt
}
\setlength{\headwidth}{\textwidth}

\maketitle
\thispagestyle{fancy}

\begin{abstract}
Scarcity of labeled data limits development of deep learning biomarkers in oncology.
We develop a two-stage AI model predicting pathological complete response (pCR) to neoadjuvant therapy in breast cancer.
The first stage learns the transcriptome from histopathology using 8,742 patients across 32 cancer types, corroborated by pathologist review and spatial agreement with measured expression.
This simplifies the second stage to predicting pCR from inferred expression and clinical variables.
Developed using 1,080 patients (five cohorts) and evaluated in 1,412 patients (nine cohorts), the model achieves a pooled AUROC of 0.79 (95\% CI, 0.73--0.85), discriminating responders within molecular subtypes.
It outperforms histopathological biomarkers, remaining stable across intratumoral sampling and with minimal biopsy tissue.
Ablations show transcriptome-wide inference improves discrimination over clinical variables alone or one-stage pathology models, and robustness by avoiding genomic assays' gene selection constraints.
These results indicate that biologically informed compression may generalize to data-sparse applications in precision oncology.

\end{abstract}

\newpage

\section{Introduction}

Neoadjuvant therapy (NAT) is becoming the standard of care for early triple-negative breast cancer (TNBC) and human epidermal growth factor receptor 2 (HER2)--positive patients, and is an established treatment option for selected patients with hormone receptor (HR)--positive/HER2-negative disease~\citep{korde2021neoadjuvant,loibl2024early}.
NAT allows response to be assessed at surgery, informs post-neoadjuvant treatment, and facilitates de-escalation of breast and axillary surgery~\citep{thompson2012neoadjuvant,korde2021neoadjuvant} as well as adjuvant radiation therapy~\citep{mamounas2025omitting}.
In addition, response-adapted de-escalation of chemotherapy~\citep{perez20243,esteva2024tailoring} and omission of breast surgery in exceptional responders~\citep{kuerer2022eliminating} are being investigated.
Advances in neoadjuvant therapy have substantially improved pathological complete response (pCR) rates, with recent trials reporting improvements from 51.2\% to 64.8\% in early TNBC and from 56.3\% to 67.3\% in high-risk HER2$+$ disease~\citep{schmid2020pembrolizumab,harbeck2025neoadjuvant}.
Despite these advances, approximately one third of patients still do not achieve pCR. For these patients, NAT entails months of systemic therapy, including cytotoxic chemotherapy~\citep{korde2021neoadjuvant}, and delayed definitive surgery, with the possibility of disease progression during treatment.
In HR$+$/HER2$-$ disease, pCR is less common and its implications for long-term outcomes are less firmly established~\citep{spring2020pathologic}.
This highlights the need to identify, before treatment, patients at high risk of residual disease.

However, accurate prediction at the patient level faces several challenges.
First, the available data are limited: many existing datasets are small~\citep{sammut2022multi,aswolinskiy2023proacting,huang2023artificial,farahmand2022deep,ogier2023federated}, which has prevented models from reaching the robustness and accuracy required for routine clinical use.
Second, existing biomarkers may lack sufficient expressivity to capture the biology that determines NAT response.
These include genomic assays (e.g., HER2DX~\citep{prat2022development}), receptor status (estrogen receptor (ER), progesterone receptor (PR), HER2)~\citep{korde2021neoadjuvant}, and histological features such as tumor-infiltrating lymphocytes (TILs)~\citep{denkert2018tumour,denkert2010tumor,hagenaars2021tumor,jung2016histomorphological,mao2016prognostic,stanton2016clinical,luen2019prognostic}.
Within HR$+$/HER2$-$ patients, pCR rates vary with gene-expression scores~\citep{Freeman_npjBreastCancer_2023}, motivating richer characterization of response-relevant biology beyond receptor status alone.
Transcriptome-wide analyses indicate that molecular read-outs of the tumor phenotype may be informative for predicting NAT response~\cite{cantore2026predicting,fumagalli2017rna}, but such research-grade assays are not clinically deployable at scale, and clinical RNA-based tests measure only small lists of targets (typically <100 genes~\citep{prat2022development}).
Histological features fare no better: TILs track response across subtypes but stratify coarsely~\citep{denkert2018tumour,denkert2010tumor,hagenaars2021tumor,jung2016histomorphological,mao2016prognostic,stanton2016clinical,luen2019prognostic}.
Third, assessments themselves can be unstable, for two reasons: tumor sampling bias and observer variability.
Needle biopsies typically sample less than one percent of the tumor bulk~\cite{litchfield2020representative}.
As breast tumors are heterogeneous ecosystems of malignant cells and their microenvironment~\citep{marusyk2020intratumor}, this gives rise to tumor sampling bias~\citep{biswas2019clonal}.
When the genes of five gene expression assays were measured across matched regions of the same tumor, a quarter of patients crossed a risk-category boundary depending on which region was assayed~\citep{gyanchandani2016intratumor}.
Additionally, manual assessments add observer variability: scoring of TILs and Ki-67 is only moderately reproducible between observers and laboratories~\citep{kos2020pitfalls,polley2013international}.
Stability to sampling is a prerequisite for a biopsy-based biomarker, yet it is seldom reported.

Deep learning applied to hematoxylin and eosin (H\&E) images offers a route to prediction that requires no additional tissue, molecular assay, or manual read.
Foundation models---neural networks pretrained by self-supervised learning on millions of unlabeled tissue images~\citep{cappadona2024squeezing, biswas2026causal, chen_naturemed_2024, xu_nature_2024, wang_nature_2024}---convert any H\&E slide into rich morphological representations (embeddings) without large annotated datasets.
Turning those representations into a predictor of a clinical endpoint, however, still requires many labeled patients, and NAT cohorts with recorded response are small.
Due to the large quantity and high dimensionality of pathology embeddings and the small NAT cohorts, we reasoned that dividing this learning problem into two stages could help: relating morphology to tumor biology, then relating that biology to treatment response.
Existing pan-cancer cohorts pair each patient's slides with a measured transcriptome---thousands of quantitative measurements per patient rather than one binary label.
These paired data can teach a model how morphology relates to molecular features.
Gene expression is also a useful intermediate representation because it both shapes treatment response and leaves visible traces in tissue morphology~\citep{schmauch2020deep,hoang2024deep,pizurica2024digital}.
Learning to infer expression from these larger cohorts could leave the smaller NAT cohorts with the simpler task of relating inferred molecular features to response.
Therefore, we built an AI model to infer gene expression from morphology and another AI model to predict pCR from the inferred expression~\cite{hoang2024deep,cantore2026predicting}, rather than a single model to predict response directly from the slide.
This design draws on broad transcriptional information without requiring an additional molecular assay or relying on a narrowly selected gene list, and yields an interpretable representation that can be related to NAT response biology.

\begin{figure}[!htbp]
    \centering
    \includegraphics[trim={0cm 0cm 0cm 0.3cm}, clip, width=0.89\linewidth]{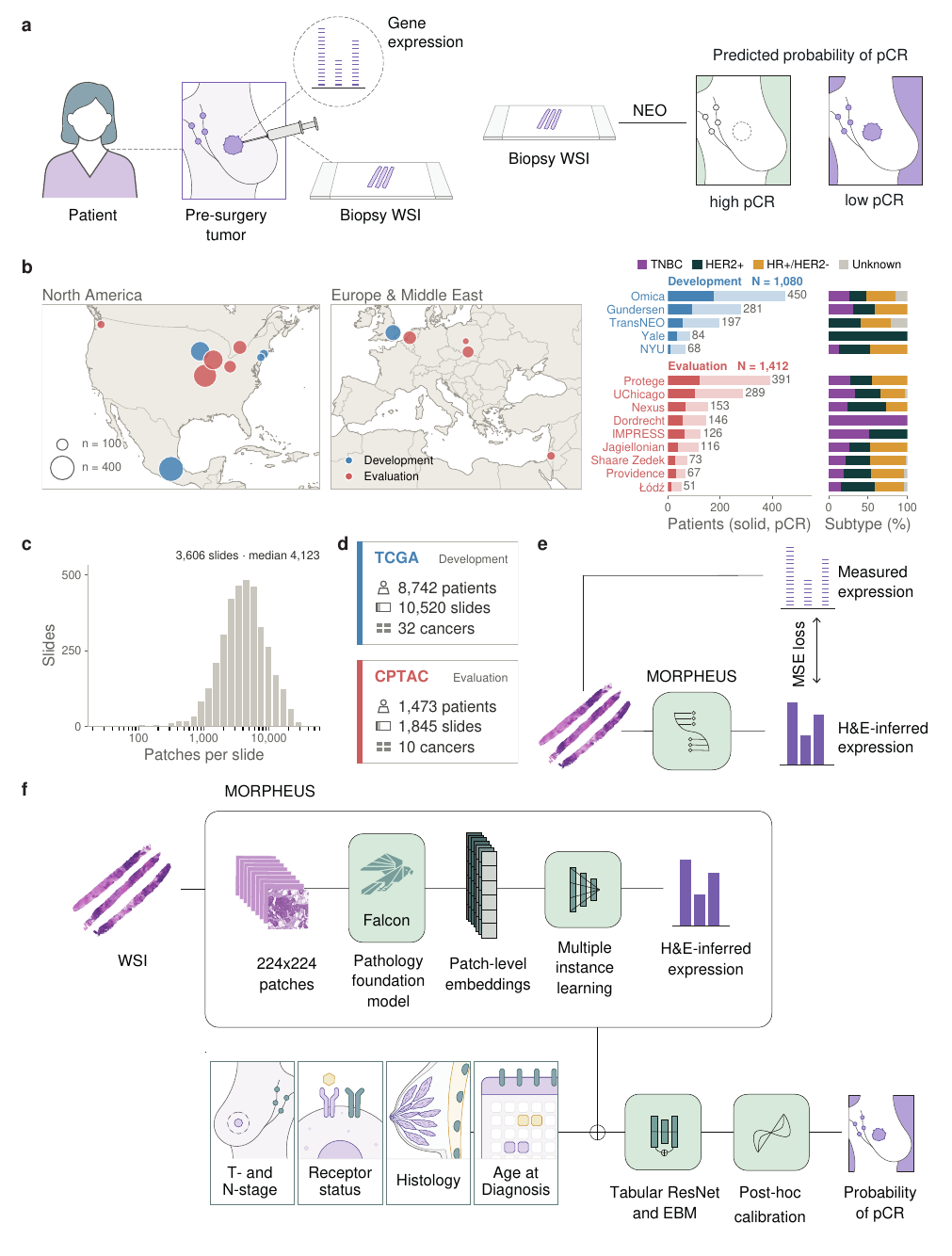}
    \vspace{-3mm}
    \caption{
    \textbf{Overview of the study and NEO architecture.}
    \textbf{a}, Clinical setting and prediction task.
    The tumor's gene expression reflects its biological state, and that state also manifests in tissue morphology.
    The patient's tumor is sampled by core needle biopsy and digitized as an H\&E-stained whole-slide image (WSI).
    From this biopsy WSI, it is possible to infer gene expression, as a substrate for predicting the probability of pathological complete response (pCR) to neoadjuvant therapy (NAT).
    \textbf{b}, Development and evaluation cohorts. Left, the contributing institutions, with marker area proportional to cohort size. Right, two bar charts: cohort size and molecular subtype composition. In the size chart, bar length is the number of patients and the solid segment is those achieving pCR. Throughout, blue denotes the development set (5 cohorts, N\,=\,1,080) and red denotes the held-out evaluation set (9 cohorts, N\,=\,1,412). ``Unknown'' subtype covers patients whose HER2 status is equivocal or whose receptor status is not recorded.
    \textbf{c}, Distribution of the number of 224$\times$224-pixel patches per slide across all 3,606 slides in both splits (log scale; median 4,123 per slide).
    \textbf{d}, Cohorts used to build and evaluate MORPHEUS. TCGA supplied the paired H\&E slides and bulk RNA-seq on which MORPHEUS was developed; CPTAC was held out and used only for evaluation. Counts are patients, whole-slide images and distinct cancer types.
    \textbf{e}, MORPHEUS training and inference. Trained on TCGA cases with paired H\&E slides and bulk RNA-seq, MORPHEUS infers expression from WSIs alone, requiring no transcriptomic assay at inference.
    \textbf{f}, NEO inference pipeline. MORPHEUS converts the biopsy WSI into H\&E-inferred expression, which is combined with baseline clinical variables and mapped to a calibrated probability of pCR (\Cref{sec:model-description}).
    }
    \label{fig:overview}
\end{figure}

Here we present Ataraxis Breast NEO (hereafter NEO), a multi-modal AI model that estimates the probability of pCR from a pre-treatment biopsy H\&E slide together with baseline clinical variables (\Cref{fig:overview}a).
Its first step, MORPHEUS (Morphological Oncology Representation and Prediction of High-dimensional Expression Using Slides), is a pan-cancer model developed on 8,742 patients from The Cancer Genome Atlas (TCGA)~\citep{hoadley2018cell} spanning 32 projects to infer the expression of 14,773 genes from morphology alone.
Its second step predicts pCR from that inferred expression with the standard clinical variables.
NEO was developed on 1,080 patients from five cohorts and evaluated on 1,412 patients from nine independent cohorts.
We assessed four aspects: discrimination relative to benchmarks, added value beyond the clinical variables, explainability of the predictions, and stability to biopsy sampling.

\section{Results}

\subsection{Multi-national development and evaluation datasets}

We assembled a multi-national dataset of digitized H\&E biopsy whole-slide images (WSIs) and clinical data from 14 breast cancer cohorts spanning seven countries across North America, Latin America, Europe, and the Middle East (\Cref{fig:overview}b).
Five cohorts from three countries (Gundersen, Omica, TransNEO~\citep{sammut2022multi}, Yale~\citep{farahmand2022her2rois} and NYU; N\,=\,1,080) were used for development and nine independent cohorts from five countries (Protege, UChicago~\citep{howard2025integration}, Nexus~\citep{nexus}, Dordrecht, IMPRESS~\citep{huang2023artificial}, Jagiellonian, Shaare Zedek, Providence~\citep{brca-psj-path}, and \L{}\'od\'z; N\,=\,1,412) for evaluation, spanning diverse molecular subtypes and geographies.
Of the development cohorts, Omica, Gundersen, and NYU were used for cross-validation (hereafter cross-validation cohorts), and TransNEO and Yale for training only (\Cref{subsec:mscv-main}).
Cohort characteristics are given in \Cref{tab:cohorts_breakdown}.
The evaluation set (\Cref{tab:cohorts_breakdown_per_cohort}), used for all reported pCR performance estimates, had an overall pCR rate of 37.3\% and a near-even split across the molecular subtypes (36.7\% TNBC, 31.5\% HR$+$/HER2$-$, 30.8\% HER2$+$).

The label used for model development and evaluation was pCR, assessed at surgery as the absence of residual invasive disease in both the breast and regional lymph nodes, consistent with the NeoSTEEP definition~\cite{litton2023standardized} (\Cref{sec:clinical_inputs}).
Where available, response was additionally characterized by residual cancer burden (RCB) class~\cite{symmans2007measurement}, from RCB-0 (equivalent to pCR) to RCB-III (extensive residual disease), which we used only as an exploratory secondary endpoint.

As the evaluation cohorts differ in size, subtype composition, and pCR rate (\Cref{fig:overview}b), performance is pooled with a random-effects model throughout (\Cref{sec:method_eval}).
This heterogeneity reflects real population differences, enabling a stronger assessment of generalizability.
In total, the imaging data comprise 3,606 pre-treatment biopsy slides carrying 19.6 million 224$\times$224-pixel tissue patches (median 4,123 per slide; \Cref{fig:overview}c).
The overall sample size of our study substantially exceeds that of most previously reported histopathology-based NAT response prediction studies~\citep{sammut2022multi,aswolinskiy2023proacting,huang2023artificial,farahmand2022deep,ogier2023federated}.

\subsection{NEO: a multi-modal AI model for predicting neoadjuvant therapy response}
\label{subsec:neo}

We developed NEO (\Cref{fig:overview,sec:model-description}), a multi-modal AI model that integrates routine pre-treatment biopsy H\&E WSIs and baseline clinical variables to estimate the probability of achieving pCR.
The model operates in two steps.
First, MORPHEUS infers expression values for 14,773 protein-coding genes from the WSIs, referred to here as H\&E-inferred expression (\Cref{fig:overview}e).
Falcon~\cite{biswas2026causal}, a pathology foundation model trained by self-supervised learning, extracts a morphological embedding for every 224$\times$224-pixel patch.
A model trained on paired H\&E and bulk RNA sequencing data from TCGA then aggregates these embeddings over the slide by multiple-instance learning (MIL) into expression values.
Second, the H\&E-inferred expression is compressed to five components (\Cref{sec:ica_readout}) and combined with baseline clinical variables (\Cref{sec:clinical_inputs}) to produce a calibrated probability of pCR (\Cref{fig:overview}f).
This is the only step trained on response data: explainable boosting machines (EBMs)~\citep{nori2019interpretml,caruana2015intelligible} and tabular ResNets~\citep{gorishniy2021revisiting} were fitted on the 1,080 development patients, calibrated post hoc (\Cref{sec:calibration}), and ensembled (\Cref{sec:response_model}).

\subsection{External evaluation and benchmarking of NEO}

\begin{figure}[!htbp]
    \centering
    \includegraphics[width=\linewidth]{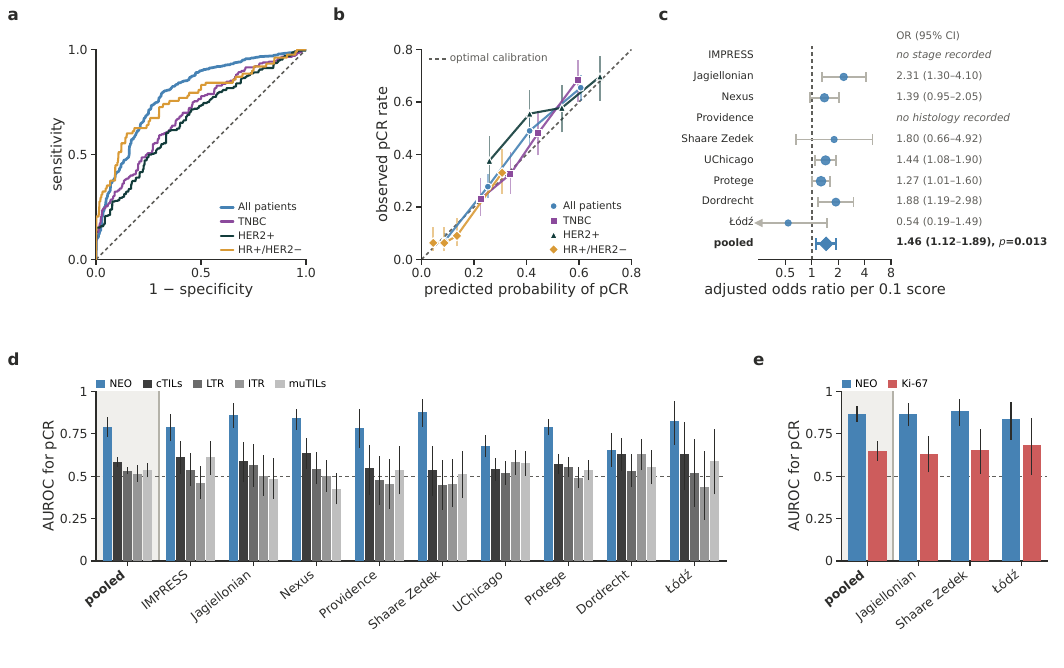}
    \caption{
    \scriptsize
    \textbf{NEO performance on independent evaluation cohorts ($N=1{,}412$).}
    \textbf{a}, Receiver operating characteristic curves for pCR in all patients (blue), TNBC (purple), HER2$+$ (green), and HR$+$/HER2$-$ (gold); each curve is the pooled average of the per-cohort curves, and the per-cohort estimates behind it are shown in \Cref{fig:cohort_forests}.
    \textbf{b}, Observed pCR rate against mean predicted probability in four equal-count bins of predicted probability, for all patients (blue circles), TNBC (purple squares), HER2$+$ (green triangles), and HR$+$/HER2$-$ (gold diamonds).
    \textbf{c}, Adjusted odds ratio of pCR per 0.1-unit increase in the score (\Cref{sec:adjusted_analysis}), from a Firth-penalized logistic regression within each cohort: one blue circle per cohort, area proportional to its weight, with the pooled estimate as a blue diamond beneath. $n=1{,}031$ patients in seven cohorts; IMPRESS and Providence do not record a required covariate, contribute no complete case, and appear as excluded rows. The axis is logarithmic, and arrowheads mark intervals continuing beyond the axis; exact values are given on the right.
    \textbf{d}, AUROC of NEO (blue bars) and of four computational TIL biomarkers, the computational tumor-infiltrating lymphocytes, the lymphocytes-to-tumor ratio, the inflamed tumor ratio~\citep{aswolinskiy2023proacting}, and muTILs~\citep{liu2024panoptic} (gray bars, darkest to lightest in that order), by cohort and pooled, over the slides on which all five methods return a value ($n=1{,}386$ patients; \Cref{sec:shared_slides}).
    \textbf{e}, AUROC of NEO (blue bars) and pathologist-assessed Ki-67 (red bars), by cohort and pooled, in the $n=237$ patients for whom Ki-67 was recorded.
    Statistics: pooled AUROCs and odds ratios are random-effects estimates over the contributing cohorts (\Cref{sec:method_eval}). Error bars are 95\% confidence intervals: bootstrap for cohort AUROCs in \textbf{d},\textbf{e}; Wilson in \textbf{b}; Wald for odds ratios within cohort in \textbf{c}; and Hartung--Knapp for pooled estimates in \textbf{c}--\textbf{e}. Dashed lines mark the null reference: chance discrimination in \textbf{a}, optimal calibration in \textbf{b}, OR $=1$ in \textbf{c}, AUROC $=0.5$ in \textbf{d}, \textbf{e}.
    The $p$ value in \textbf{c} is from the random-effects model; paired comparisons between methods in \textbf{d} and \textbf{e}, with their $p$ values, are given in \Cref{fig:paired_comparisons}a--e.
    }
    \label{fig:performance}
\end{figure}

We evaluated NEO on nine independent external cohorts comprising 1,412 patients (\Cref{fig:performance}).
NEO achieved a pooled area under the receiver operating characteristic curve (AUROC) of 0.792 (95\% CI, 0.732--0.852; $p = 3.4\times10^{-6}$ versus a null AUROC of 0.5), with substantial heterogeneity across cohorts ($I^2$ = 74\%, $\tau^2$ = 0.0040; both defined in \Cref{sec:supp_evaluation}).
Discrimination was above chance in every cohort: per-cohort AUROCs ranged from 0.658 to 0.880, and all 95\% confidence intervals excluded 0.5 (\Cref{fig:cohort_forests}).

Discrimination held within each molecular subtype (\Cref{fig:performance}a):
0.699 (95\% CI, 0.643--0.755; $I^2$ = 0\%, $\tau^2$ = 0; seven cohorts) for TNBC, 0.669 (0.576--0.763; $I^2$ = 58\%, $\tau^2$ = 0.0077; eight cohorts) for HER2$+$, and 0.762 (0.686--0.837; $I^2$ = 0\%, $\tau^2$ = 0; six cohorts) for HR$+$/HER2$-$ disease, each pooled over the cohorts with sufficient samples of that subtype (\Cref{sec:method_eval}; per-cohort estimates in \Cref{fig:cohort_forests}b--d).

We next assessed calibration, the agreement between predicted and observed pCR rates (\Cref{fig:performance}b, \Cref{sec:calibration_assessment}).
The calibration slope did not differ significantly from 1 in any stratum (all patients: 1.11, 95\% CI, 0.97--1.25; TNBC: 1.21, 95\% CI, 0.89--1.52; HER2$+$: 0.75, 95\% CI, 0.47--1.04; HR$+$/HER2$-$: 1.17, 95\% CI, 0.84--1.51), consistent with appropriately dispersed predictions.
The intercept indicated under-prediction overall (0.18, 95\% CI, 0.06--0.30) and in HER2$+$ disease (0.36, 95\% CI, 0.16--0.56) and did not differ significantly from zero in TNBC (0.13, 95\% CI, -0.05--0.32) and HR$+$/HER2$-$ disease (-0.06, 95\% CI, -0.35--0.23).

To assess whether NEO does not merely restate the clinical variables available at diagnosis, we fitted a Firth-penalized logistic regression of pCR on the score within each cohort, adjusted for baseline clinical variables NEO consumes (\Cref{sec:clinical_inputs}) together with histologic grade, and pooled the log odds ratios (\Cref{fig:performance}c, \Cref{sec:adjusted_analysis}).
Among complete cases (1,031 of the 1,412 patients, seven cohorts), the NEO score remained associated with pCR, with a pooled odds ratio of 1.46 per 0.1-unit increase (95\% CI, 1.12--1.89; $p = 0.013$; $I^2$ = 31\%, $\tau^2$ = 0.020), so it carries information the clinical variables do not.

We next benchmarked NEO against four computational TIL biomarkers (\Cref{fig:performance}d): computational tumor-infiltrating lymphocytes (cTILs), the lymphocytes-to-tumor ratio (LTR), the inflamed tumor ratio (ITR)~\citep{aswolinskiy2023proacting}, and muTILs~\citep{liu2024panoptic}.
We re-implemented the first three from their published definitions and ran muTILs with the authors' released models (\Cref{sec:benchmark_biomarkers}).
On IMPRESS, our cTILs implementation reproduced the published results, whereas we were not able to reproduce the results reported for LTR and ITR (\Cref{tab:impress_reproduction}).
These methods score a slide only when their models detect the tissue compartments required by the metric, which failed on 108 of the 2,489 evaluation slides.
The comparison therefore uses the shared-slide set, the 1,386 of 1,412 patients scored by all five methods (\Cref{sec:shared_slides}).
NEO reached a pooled AUROC of 0.791 (95\% CI, 0.732--0.850) against 0.516--0.583 for the four biomarkers.
NEO's advantage, expressed as the pooled paired AUROC difference, ranged from $+0.192$ AUROC over cTILs to $+0.279$ over ITR, all $p < 5\times10^{-4}$ (\Cref{fig:paired_comparisons}a--d).
Within subtypes, all twelve paired differences directionally favored NEO and eleven reached $p < 0.05$.
The exception was NEO versus ITR for TNBC patients ($+0.075$, 95\% CI, $-0.001$--$0.152$; $p = 0.052$; \Cref{fig:subtype_biomarkers}).

Against pathologist-assessed Ki-67, recorded in three cohorts for 237 patients, NEO reached 0.868 (95\% CI, 0.821--0.914) versus 0.648 (95\% CI, 0.591--0.706), a paired difference of $+0.216$ (95\% CI, $+0.115$--$+0.317$; $p = 0.012$; \Cref{fig:performance}e; \Cref{fig:paired_comparisons}e).

To assess whether NEO captures gradations of response beyond binary pCR, we also evaluated the same score against RCB class in the 361 patients across four cohorts with recorded RCB (inclusion criteria in \Cref{sec:method_eval}).
The NEO score decreased across increasing RCB from RCB-0 to RCB-III, with medians of 0.486, 0.312, 0.259, and 0.139 ($p = 4.2\times10^{-25}$, one-sided cohort-stratified Jonckheere--Terpstra test; \Cref{sec:rcb_ordinal}).
It also separated RCB-0/I from RCB-II/III (AUROC 0.792, 95\% CI, 0.656--0.928; \Cref{fig:secondary_endpoints}b).
Finally, the same score also ranked nodal response, without retraining, in the 536 clinically node-positive patients across six cohorts with determinable nodal outcomes (pooled AUROC 0.736, 95\% CI, 0.644--0.829; \Cref{sec:nodal_pcr}).
For context, AUROC for overall pCR was 0.775 (95\% CI, 0.632--0.918) in the same 536 patients.
Both secondary endpoints overlap with the primary one: RCB-0 is pCR by definition, and an overall pCR implies nodal pCR, so neither contrast is fully independent of the primary result.
We regard both as exploratory, as RCB was assessed for only a quarter of the evaluation patients and the nodal outcome is non-randomly missing for 16\% of clinically node-positive patients.

Taken together, NEO discriminated pCR in every evaluation cohort and molecular subtype, added information beyond the standard clinical variables, and outperformed the computational TIL biomarkers and Ki-67 on the same patients.

\subsection{MORPHEUS: H\&E-inferred expression recovers biologically meaningful programs}
\label{subsec:morpheus}

The information passed to NEO from the biopsy slide is purely mediated via H\&E-inferred gene expression, so what it can learn about tumor biology is bounded by what that expression captures.
Here, we examined how faithfully gene expression is inferred, which biological programs are recovered, and whether the inferred expression is spatially and morphologically grounded in the tissue.

MORPHEUS was developed on paired H\&E and bulk RNA-seq from 32 TCGA projects, comprising 8,742 patients and 10,520 slides.
We used a subset of this TCGA cohort (n = 1,748 patients) as the validation set, and the Clinical Proteomic Tumor Analysis Consortium (CPTAC)~\cite{ellis2013connecting} cohort (n = 1,473 patients, providing 1,566 primary-tumor samples across 10 cancer types) as an evaluation cohort (\Cref{fig:overview}d; \Cref{tab:morpheus_breakdown_tcga,tab:morpheus_breakdown_cptac}).
We define fidelity as the per-gene Pearson correlation $r$ between inferred and measured expression within each cancer type (\Cref{sec:fidelity_metric}).
On the TCGA validation subset the mean per-type median $r$ was 0.27 (range 0.16--0.39) across the 23 projects with at least 20 validation patients, rising to 0.59 (0.44--0.76) over each type's 1,000 best-predicted genes (\Cref{fig:ed_morpheus_fidelity}).
That subset of TCGA also drove model selection, so these values are likely optimistic.
In evaluation on CPTAC cases, the corresponding values were 0.20 (0.13--0.25) and 0.52 (0.44--0.60), with every cohort above zero.
Thus, expression is recovered in every cancer type and transfers to data never seen during training or validation.

\begin{figure}[!htbp]
    \centering
    \includegraphics[trim={0cm 0cm 0cm 0.32cm}, clip, width=0.94\linewidth]{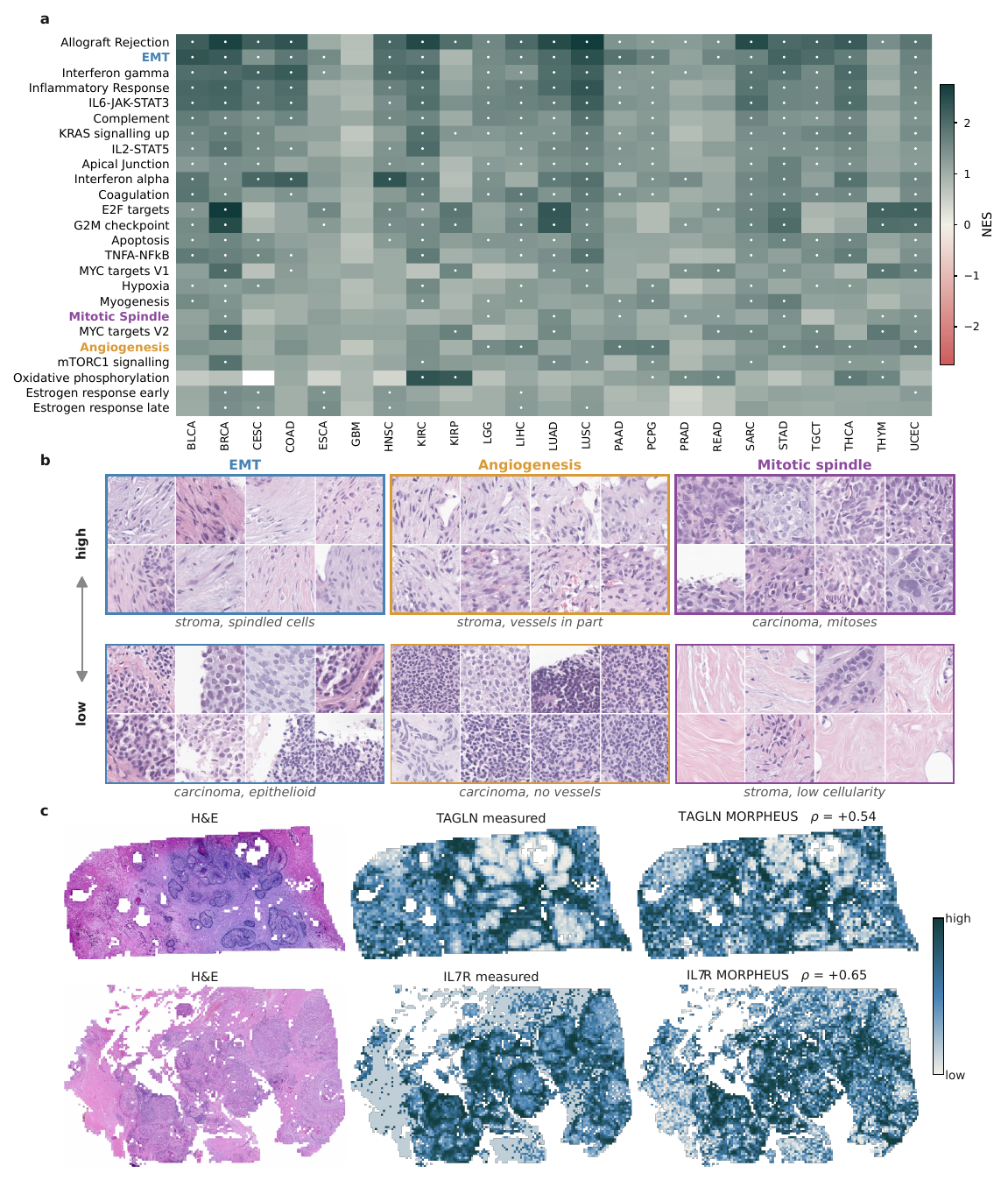}
\caption{
    \scriptsize
    \textbf{Transcriptional programs recovered from H\&E by MORPHEUS.}
    \textbf{a}, Hallmark pathway enrichment of MORPHEUS prediction fidelity within each TCGA project. Columns, the 23 projects with at least 20 validation patients (\Cref{tab:morpheus_breakdown_tcga}); rows, the 25 hallmark sets of 50 that reached significance in at least six of the 23 projects.
    Color, normalized enrichment score (NES) from genes ranked by the within-project Pearson correlation between inferred and measured expression; a positive NES indicates that the set's genes are predicted better than the transcriptome-wide background. White dots, Benjamini--Hochberg adjusted $p < 0.05$ within that project.
    \textbf{b}, Highest- and lowest-scoring patches among the 100 sampled at random from each slide of the Jagiellonian evaluation cohort, for three hallmark programs: epithelial-mesenchymal transition (EMT), angiogenesis, and the mitotic spindle. Each patch was scored by the mean H\&E-inferred expression of that hallmark set's members (\Cref{sec:bio_correlates_methods}). Eight patches per block; high and low are the two ends of the same ranking. The italic label below each block summarizes what two pathologists reported for it (\Cref{sec:bio_correlates_methods}).
    \textbf{c}, Spatially resolved examples on two publicly available imaging-based spatial transcriptomic slides of human breast cancer, neither seen by MORPHEUS at any stage: one Xenium v1 slide and one Xenium Prime 5K slide, one tissue section each (\Cref{sec:spatial_methods}). One gene per program, each among the best-inferred members of its set (\Cref{sec:spatial_methods}): \textit{TAGLN} (EMT; Xenium v1, 3,614 patches) and \textit{IL7R} (inflammatory response; Xenium Prime, 8,068 patches). Each slide was tiled into 224$\times$224-pixel patches; the measurement paired with a patch is log-transformed transcript abundance normalized to the total transcript count across all measured genes in that patch (\Cref{sec:spatial_methods}).
    Left to right, H\&E; measured expression; MORPHEUS prediction.
    Predictions were generated per patch with no spatial supervision, retraining, or fine-tuning of any kind.
    Both maps show percentile rank within the slide, one cell per patch; $\rho$ is the Spearman correlation between prediction and measurement across those patches (\Cref{sec:spatial_methods}).
    }
    \label{fig:morpheus}
\end{figure}

Next, we asked whether certain gene expression pathways are more or less learnable from morphology.
For example, TILs are known to be readily readable from pathology images by human~\cite{salgado2015evaluation} and rule-based computer systems~\cite{aswolinskiy2023proacting,liu2024panoptic}, so we might expect associated gene expression pathways to be readily encoded by MORPHEUS.
We performed a transcriptome-wide analysis for each of 23 TCGA cancer types, first ranking genes from high to low accuracy of inferred gene expression, which was used as input to gene set enrichment analysis (\Cref{fig:morpheus}a; \Cref{sec:bio_correlates_methods}).
Several pathways were commonly enriched across multiple cancer types (adjusted $p < 0.05$): immune pathways, including allograft rejection (21 cancer types), interferon gamma response (19), IL6--JAK--STAT3 signaling (17) and inflammatory response (17); stromal pathways, including epithelial-mesenchymal transition (EMT; 19), apical junction (16) and coagulation (15).
Within cancer types, tissue-specific patterns of enrichment were also observed.
Proliferation pathways --- including E2F targets and the G2M checkpoint --- were enriched in 14 cancer types, consistent with mitotic activity being a grading feature at some sites and not others.
In addition, estrogen response pathways reached significance in 6 cancer types, with a highly significant signal in breast (adjusted $p < 0.002$).

As a sanity check, we asked whether the morphological correlates of the inferred expression were discernible on the corresponding H\&E slide.
We randomly sampled 100 H\&E image patches per slide from the Jagiellonian cohort, a representative cohort from the evaluation set, and applied MORPHEUS to score each patch by the mean H\&E-inferred expression of each gene expression pathway.
We focused on three gene expression pathways (EMT, angiogenesis, and mitotic spindle) which may have morphological correlates, and asked two board-certified pathologists with subspecialty expertise in breast pathology to review the highest- and lowest-scoring patches (\Cref{fig:morpheus}b; \Cref{sec:bio_correlates_methods}).
For EMT, both reviewers found the contrast consistent with the program: the high-scoring patches contained cells that were more mesenchymal in appearance and predominantly stromal; the low-scoring patches contained cells that displayed more epithelioid and carcinoma-like features.
For angiogenesis, vessels were identified in the high-scoring set, with none seen in the low-scoring set.
Similarly, for mitotic spindle, high-scoring patches exhibited several mitotic figures, with none visible in the low-scoring set.

Lastly, we subjected MORPHEUS to a more stringent test, asking whether predictions learned at the bulk tumor level recapitulated expression measured at patch-level resolution by spatial transcriptomics.
The imaging-based platforms used here localize individual transcripts from a targeted gene list, so the transcripts of a given gene detected within a patch can be counted and set against the value inferred for that same patch.
We used two publicly available breast cancer spatial transcriptomics datasets, one Xenium v1~\citep{janesick2025biomarker} and one Xenium Prime 5K (\Cref{fig:morpheus}c; \Cref{sec:spatial_methods}).
Neither slide is in TCGA, and both were profiled with a different assay from the bulk RNA-seq used in training.
MORPHEUS was applied zero-shot --- without retraining, fine-tuning, or spatial supervision of any kind --- focusing on the highest fidelity pathways.
We evaluated at the gene-level, with each gene drawn from the members driving its set's enrichment in \Cref{fig:morpheus}a.
For \textit{TAGLN}, a myofibroblast marker in the EMT set, the inferred and measured maps were positively correlated (Spearman $\rho = +0.54$ across the 3,614 patches of the Xenium v1 slide). The same held for \textit{IL7R}, a lymphocyte surface receptor in the inflammatory response set ($\rho = +0.65$ across the 8,068 patches of the Xenium Prime slide).
In these examples, patch-level predictions of MORPHEUS track spatial variation within a slide.

Collectively, these data indicate that the H\&E-inferred gene expression is quantitatively faithful across tumor types, recovering biologically coherent immune, stromal, and proliferative pathways; that it aligns with what human experts see; and that, for the genes and sections examined, patch-level predictions track measured spatial variation in gene expression.

\subsection{Integration of H\&E-inferred expression in pCR prediction: performance, explainability, and design}

\begin{figure}[!htbp]
    \centering
    \includegraphics[width=\linewidth]{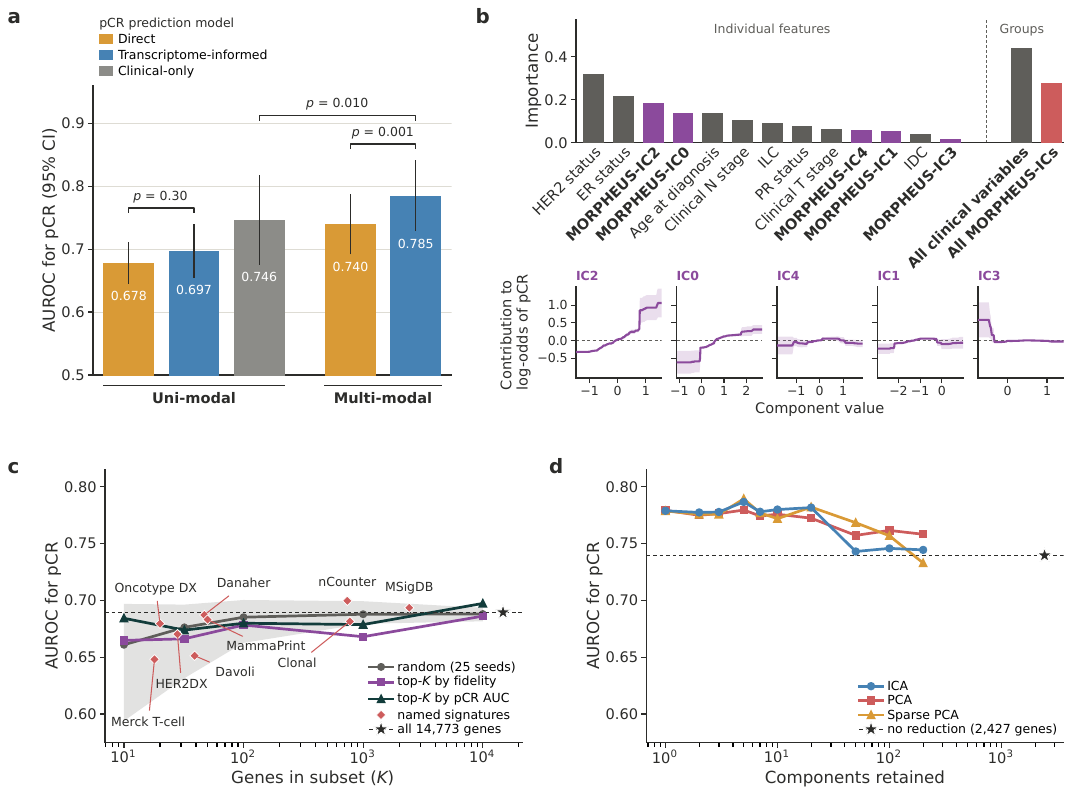}
\caption{
    \scriptsize
    \textbf{Interpretability of the H\&E-inferred expression, and ablations of its contribution to pCR prediction.}
    \textbf{a}, Pooled AUROC for pCR of the transcriptome-informed and direct pCR prediction models, uni-modal (H\&E only), and multi-modal (H\&E with the clinical variables), alongside a pCR prediction model using the clinical variables alone. Transcriptome-informed pCR prediction models use the H\&E-inferred expression to predict pCR. Direct pCR prediction models receive the same foundation-model embeddings that MORPHEUS uses and predict pCR from them directly (\Cref{sec:direct_path_model}). Error bars are 95\% confidence intervals.
    Brackets give two-sided $p$ values for pooled paired AUROC differences (per-cohort differences in \Cref{fig:paired_comparisons}f--h).
    Panel \textbf{a} ablates NEO's neural-network-based pCR prediction model rather than its full ensemble, so that the transcriptome-informed and direct pCR prediction models are neural networks matched in shared hyperparameter ranges and tuning budget.
    \textbf{b}, Global feature importance of the EBM trained on the clinical variables and the five MORPHEUS-ICs of the H\&E-inferred expression (top), and the shape functions of the five MORPHEUS-ICs (bottom). Past the divider, the two bars give the importance of the five MORPHEUS-ICs taken as a group and of the eight clinical variables taken as a group (\Cref{sec:ebm_interpretation}). Each shape function shows the amount each MORPHEUS-IC adds to the predicted log-odds of pCR as the component varies, with all other features held fixed. The band is $\pm$1 s.d. across folds.
    \textbf{c}, Pooled AUROC of pCR prediction models trained on H\&E-inferred expression alone, for gene subsets of increasing size selected four ways: at random (gray; mean and range over 25 seeds), by expression-prediction fidelity (purple), by per-gene pCR discrimination in the cross-validation cohorts (green), and as published clinical or curated signatures (red). Dashed line with star, all 14,773 genes.
    \textbf{d}, Pooled AUROC of the multi-modal pCR prediction model against the number of components the 2,427-gene H\&E-inferred expression is reduced to, for three reduction methods, each fitted on the TCGA validation subset (\Cref{sec:ica_readout}): ICA (purple), principal component analysis (green), and sparse principal component analysis (gold); the star and the dashed line mark the unreduced representation, all 2,427 standardized genes.
    AUROCs in \textbf{a},\textbf{c},\textbf{d} pool the nine evaluation cohorts ($N=1{,}412$) with a random-effects model (\Cref{sec:method_eval}).
    }
    \label{fig:tech_interpretability}
\end{figure}

NEO predicts pCR through H\&E-inferred expression rather than directly from the image.
That intermediate step yields a set of expression values for named genes, so, unlike an image embedding, it can be interpreted directly, and its composition can be chosen: which genes the model is given, and how compactly they are summarized, are design decisions rather than assay constraints.
Here, we examined what this design provides: whether it improves discrimination (\Cref{fig:tech_interpretability}a), what it reveals about the biology behind each prediction (\Cref{fig:tech_interpretability}b, \Cref{fig:neo_bio_correlates}), and how performance depends on the genes supplied and on the size of the summary (\Cref{fig:tech_interpretability}c, d).

We first tested whether that intermediate step improves discrimination.
We refer to models that predict pCR from H\&E-inferred expression as transcriptome-informed.
For comparison, we trained MIL models that predict pCR directly from the foundation-model embeddings MORPHEUS takes as input.
We refer to these as direct pCR prediction models.
We scored them on identical patients in all nine evaluation cohorts (\Cref{fig:tech_interpretability}a; \Cref{sec:direct_path_model}).
Transcriptome-informed and direct pCR prediction models were each trained in two settings: uni-modal, and with the clinical variables added (\Cref{fig:model_components}).
The comparison used NEO's neural-network-based pCR prediction model rather than its full ensemble, matching the direct models in architecture family (\Cref{sec:direct_path_model}).
In the multi-modal setting in which NEO operates, the transcriptome-informed pCR prediction model reached a pooled AUROC of 0.785 (95\% CI, 0.728--0.841) against 0.740 (95\% CI, 0.692--0.788) for the direct pCR prediction model, with a pooled paired difference of $+0.038$ AUROC (95\% CI, $+0.020$--$+0.056$; $p = 1.4\times10^{-3}$; \Cref{fig:paired_comparisons}h).
Without the clinical variables the ordering was the same but the difference was not statistically significant (0.697 versus 0.678; $+0.017$, 95\% CI, $-0.018$--$+0.053$; $p = 0.30$; \Cref{fig:paired_comparisons}g).
H\&E-inferred expression also improved on the clinical variables by themselves (0.785 versus 0.746; $+0.028$, 95\% CI, $+0.009$--$+0.047$; $p = 9.6\times10^{-3}$; \Cref{fig:paired_comparisons}f).

The pCR prediction model consumes the H\&E-inferred expression not gene by gene but as five independent components (ICs), MORPHEUS-IC0 to MORPHEUS-IC4, collectively the MORPHEUS-ICs.
The H\&E-inferred expression of a fixed set of 2,427 genes drawn from the Molecular Signatures Database (MSigDB) gene families~\citep{subramanian2005gene,liberzon2011molecular} is reduced by independent component analysis (ICA), fitted once on the TCGA validation subset and then frozen (\Cref{sec:ica_readout}).

Because the H\&E-inferred expression enters the model as five tabular components, its contribution to the prediction is inspectable with an EBM.
In NEO's EBM, the five MORPHEUS-ICs carried a group importance of 0.28 in log-odds units compared to 0.44 for the eight clinical variables combined, a ratio of 0.63.
Individually, MORPHEUS-IC2 and MORPHEUS-IC0 ranked third and fourth of thirteen input features respectively and both contributed nearly monotonically to the predicted log-odds of pCR, whereas the other three components were mostly flat (\Cref{fig:tech_interpretability}b; \Cref{sec:ebm_interpretation}).
We annotated MORPHEUS-IC2 and MORPHEUS-IC0 using gene set enrichment analysis over genes ranked by their weight within each component (\Cref{fig:ic_pathways}; \Cref{sec:bio_correlates_methods}).

We next examined molecular correlates of NEO predictions, relating the score to measured gene expression in the TCGA breast cancer cohort (TCGA-BRCA) (\Cref{fig:neo_bio_correlates}; \Cref{sec:neo_correlates_methods}).
These analyses used the full TCGA-BRCA cohort, since the validation subset alone (212 patients) was too small to estimate enrichment stably.
No TCGA patient carried a pCR label or entered the training of the pCR prediction models, but because inferred expression may track measured expression more closely on MORPHEUS's own training-set patients, the association strengths below may be somewhat overstated.

Pathway analysis of NEO scores found proliferative programs (E2F targets, G2M checkpoint, MYC targets, and mitotic spindle) to be positively associated with predicted pCR probability, consistent with the greater chemosensitivity of highly proliferative tumors~\cite{stover2016role} (\Cref{fig:neo_bio_correlates}a).
Immune and inflammatory signaling (interferon alpha and gamma signaling, IL6-JAK-STAT3, and complement) was also positively associated with NEO scores, consistent with the immune-dependent mechanisms engaged by anti-HER2 antibodies, immune checkpoint blockade, and chemotherapy-induced immunogenic cell death. In contrast, both early and late estrogen response were negatively associated with NEO scores. This finding is consistent with the low pCR rates of ER-driven tumors~\cite{yau2022residual}.

To characterize the tumor immune microenvironment, we performed immune cell-type signature enrichment using single-sample GSEA, scoring each tumor's expression profile against the 23 immune signatures of \citet{bindea2013spatiotemporal}, and tested each signature's association with the NEO score. B cells, activated dendritic cells, and helper T cells were positively associated with NEO scores (\Cref{fig:neo_bio_correlates}b).
These associations are compatible with lymphoid immune activity.
In contrast, tissue-resident cell types (including eosinophils and mast cells) were negatively associated with NEO score.

Within PAM50 subtypes, NEO scores were highest in basal-like and HER2-enriched tumors and lowest in luminal A, reflecting the established hierarchy of pCR rates across molecular subtypes~\cite{ohara2019pam50,stover2016role} (\Cref{fig:neo_bio_correlates}c).

We next asked how much of the transcriptome the response signal requires, and whether choosing genes deliberately improves on an unselected list.
We trained EBMs on H\&E-inferred expression alone, reduced to five MORPHEUS-ICs as in NEO (\Cref{sec:ica_readout}), and for subsets ranging from 10 genes to the 14,773 genes MORPHEUS predicts.
Subsets were chosen four ways: at random (25 draws per size), by MORPHEUS fidelity, by per-gene pCR discrimination on the cross-validation cohorts, and as published signatures (\Cref{fig:tech_interpretability}c; \Cref{sec:gene_list_resolution}).
\Cref{fig:tech_interpretability}c should be interpreted for the trends across length of gene list and selection strategy rather than for differences between individual points.

Two trends emerged from the random draws.
As the gene list size grew, mean performance rose, from 0.661 at 10 genes to 0.688 at 1,000, against 0.689 at all 14,773 genes.
The range in performance across draws also narrowed, from roughly 0.10 in AUROC at 10 genes to under 0.02 at 1,000.
Of the 25 random 10-gene lists, 10 performed below 0.66 and the worst reached only 0.594, whereas every 1,000-gene draw fell within 0.679--0.699.
Every random 1{,}000-gene draw thus had a pooled AUROC close to that of all 14,773 genes, whichever genes it happened to contain.

Deliberate selection might be expected to improve on an unselected gene list.
For example, the 10 genes most associated with pCR on the cross-validation cohorts are \textit{AIM2}, \textit{GZMB}, \textit{APOBEC3A}, \textit{LAMP3}, \textit{TNIP3}, \textit{CTLA4}, \textit{ADAMDEC1}, \textit{CYB5R4}, \textit{TAP2}, and \textit{APOL4}, most of which are immune-related.
Yet, these strategies showed no consistent advantage over random selection across gene list sizes.
Ranking by per-gene pCR discrimination reached 0.684 with only 10 genes but changed little thereafter (0.679 at 1,000), and ranking by MORPHEUS fidelity fell below every random draw at 1,000 genes (0.668).
The nine published signatures spanned 0.648 to 0.700, within the range of the random draws (\Cref{fig:tech_interpretability}c).
NEO's 2,427-gene MSigDB-family gene list is of the broad kind these trends favor, and its genes were no more correlated with one another than those of a random draw of its size (mean absolute pairwise correlation 0.254; \Cref{fig:gene_redundancy}).
NEO's gene list was fixed before these experiments (\Cref{sec:ica_readout}), which did not inform its choice.

Finally, we examined the other design parameter: the number of components the gene list is reduced to.
We fixed this at five during development, where it gave the best validation performance.
A post hoc sweep on the evaluation cohorts is consistent with this choice: with clinical variables, pooled AUROC declines as the representation grows beyond five components, for all three reduction methods tested---ICA, PCA and sparse PCA (\Cref{fig:tech_interpretability}d).
The three methods performed comparably at five components; we chose ICA because it targets statistically independent components, a property suited to interpreting them individually.

Together, these results indicate that prediction through H\&E-inferred gene expression outperforms direct prediction from the same embeddings in the multi-modal setting and adds information beyond clinical variables alone. The representation is explainable, and its biology matches the established correlates of neoadjuvant response.
They also indicate how to parameterize that representation: a broad, heterogeneous gene list, reduced to a small number of components.
No selection strategy showed a consistent advantage over random selection, and five components sufficed to carry the signal.

\subsection{Robustness of NEO predictions to tumor sampling bias}

\begin{figure}[!htbp]
    \centering
    \includegraphics[width=0.95\linewidth]{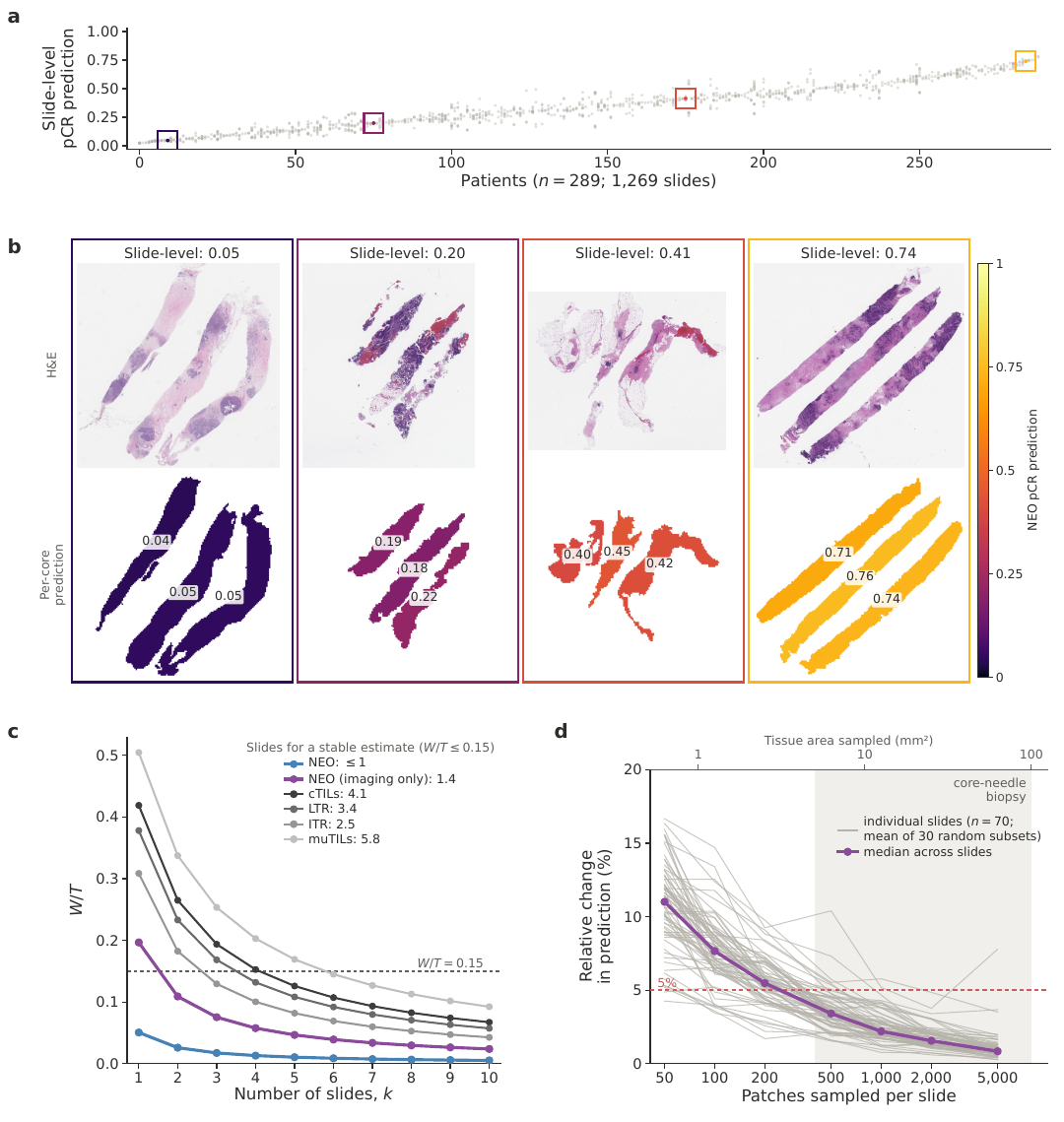}
    \caption{
    \scriptsize
    \textbf{Robustness of NEO to tumor sampling bias.}
    \textbf{a}, Whole-slide predictions for every UChicago slide, patients ordered by their mean; the four examples in \textbf{b} are the boxed points, outlined in the same colors.
    \textbf{b}, Per-core predictions on example UChicago slides along with their whole-slide prediction (given above each column). \emph{Top}: H\&E slide. \emph{Bottom}: the same field with each core scored independently by NEO, colored on the shared pCR scale with its score overlaid. Each column is outlined in the color of its own whole-slide prediction, keying it to panel \textbf{a}.
    \textbf{c}, Fraction of total variance attributable to sampling, $W/T$, against the number of slides averaged, $k$, for NEO, an imaging-only variant of NEO, and the four computational TIL biomarkers, on the UChicago multi-slide patients within the shared-slide set (\Cref{sec:shared_slides}; $n=242$ patients with $\ge 2$ pre-treatment slides). The dashed line is the stability threshold $W/T=0.15$~\citep{bachtiary2006gene}; the slides needed to reach it are given in the legend.
    \textbf{d}, Dependence of the prediction on the amount of tissue sampled, for the imaging-only variant of NEO shown in \textbf{c} (\Cref{sec:imaging_only_stability}). For each of the 70 evaluation slides with $\ge$10,000 patches, the model was re-run on random subsets of 50 to 5,000 of the slide's patches (30 subsets per size). For each subset, prediction change was calculated as the absolute difference from the all-patch predicted pCR probability, divided by the all-patch prediction. Thin lines: one per slide, the mean change across its 30 random subsets at each subset size; purple line, median of these per-slide means across slides; dashed line, a 5\% change shown for reference. The shaded span marks the patch count of a typical core-needle biopsy slide in our cohorts, its lower edge (400 patches) being about the 1st percentile of per-slide patch counts across our cohorts. The top axis gives the corresponding tissue area (0.0125\,mm$^2$ per patch).
    }
    \label{fig:sampling_bias}
\end{figure}

Intratumor heterogeneity gives rise to sampling bias, which can confound genomic assays by giving different results depending on which part of the tumor bulk is sampled~\cite{gyanchandani2016intratumor,biswas2019clonal}. The problem of sampling bias is particularly relevant to prediction of pCR, as needle biopsy typically samples less than 1\% of the tumor bulk~\cite{litchfield2020representative}.
We therefore quantified how much a patient's NEO prediction depends on which slide is used, and benchmarked the computational TIL biomarkers on the same tissue. For this experiment, we used the UChicago cohort which contributes the most slides per patient (289 patients, 1,269 slides; \Cref{fig:sampling_bias}).

When NEO was applied to individual slides, predictions were highly consistent within patients (\Cref{fig:sampling_bias}a).
Across the 247 patients with more than one slide (median 4 slides per patient, up to 15), the intraclass correlation coefficient ICC(1,1) was 0.934 (95\% CI, 0.909--0.954).

As an illustration at the sub-slide level, we applied NEO separately to the individual biopsy cores of four example slides (\Cref{fig:sampling_bias}b).
In each example the independently scored cores gave similar predictions, both to one another and to the whole-slide prediction.

We then compared NEO against the computational TIL biomarkers using an established variance-decomposition metric~\citep{bachtiary2006gene,biswas2025prospective}.
The metric is $W/T$, the fraction of a biomarker's total variance attributable to tumor sampling when $k$ slides are averaged.
$W$ is the within-patient variance of the biomarker across slides of the same tumor, $B$ the variance between patients, and $T = W + B$ the total.
We take $W/T \le 0.15$ as the stability threshold adopted in the original study (\Cref{sec:sampling_robustness}).
Across the 242 multi-slide patients within the shared-slide set, NEO met the stability threshold from a single slide ($W/T=0.05$).
Every computational TIL biomarker required substantially more, from 2.5 slides for ITR to 5.8 for muTILs (\Cref{fig:sampling_bias}c).
Noting that NEO's clinical inputs are identical across a patient's slides and contribute no within-patient variance, we repeated the analysis on an imaging-only variant (\Cref{sec:imaging_only_stability}).
It required 1.4 slides, still fewer than any of the other biomarkers, so NEO was the most stable of the scores with and without its clinical inputs.

Finally, in a descriptive analysis, we asked whether predictions remain stable when far less tissue is sampled.
We randomly subsampled patches from the 70 evaluation slides containing at least 10,000 patches ($\approx$125\,mm$^2$ of tissue) and re-ran the imaging-only model on each subset (\Cref{sec:imaging_only_stability}).
Because the imaging-only model has no clinical inputs fixed across subsamples, any stability must come from the tissue-derived signal itself.
For each slide and subset size, we computed the absolute difference between each subsampled prediction and the all-patch prediction, divided by the all-patch prediction.
We averaged this relative change across 30 random subsamples per slide, and then took the median of per-slide averages.
Predictions remained stable even with substantial subsampling.
This median relative change was 3.4\% with 500 patches ($\approx$6\,mm$^2$) and 0.8\% with 5{,}000 patches, and remained below 5\% at all subset sizes of 500 patches or more.
Only at 50 patches ($\approx$0.6\,mm$^2$) did the median relative change rise to 11\% (\Cref{fig:sampling_bias}d).

Together, these data indicate that NEO yields a stable estimate from a single needle biopsy slide, a vital condition for biopsy-based biomarkers that existing histopathological markers do not meet. The imaging-only variant also remained stable under patch subsampling, supporting robustness to limited tissue.

\section{Discussion}\label{sec:discussion}

Neoadjuvant therapy commits a patient to months of systemic treatment before pathological response can be assessed at surgery.
Its principal endpoint, pCR, is strongly associated with long-term outcomes~\citep{cortazar2014pathological,spring2020pathologic}, and predicting pCR accurately offers the potential for de-escalation of chemotherapy~\citep{perez20243}, alternative regimens, or trial enrollment.
We developed NEO to estimate pCR from the two inputs already acquired at diagnosis---the biopsy H\&E slide and the baseline clinical variables---adding no tissue, assay, or manual read to the workflow.
Its first step, MORPHEUS, is a pan-cancer model inferring transcriptome-wide expression from morphology.
NEO then predicts pCR from the H\&E-inferred gene expression together with the clinical variables.
Across 1,412 patients from nine independent cohorts, NEO discriminated responders within the three molecular subtypes, and added information beyond the clinical variables already available at diagnosis.

The prediction is also interpretable in terms of the molecular signal it uses.
Rather than predicting pCR directly from an image embedding, NEO operates on five MORPHEUS-ICs, each a weighted combination of the 2,427-gene list.
The two MORPHEUS-ICs the model relies on most contribute nearly monotonically to the predicted probability, and the five together carry just under two-thirds the group importance of the eight clinical variables (\Cref{fig:tech_interpretability}b).
Consistent with this dependence on transcriptomic features, NEO's transcriptomic and immune-cell correlates are the established correlates of neoadjuvant response (\Cref{fig:neo_bio_correlates}a,b).
Grounding a prediction in established response biology, rather than in an opaque embedding, may ease its path to clinical adoption.

These correlates recover the axes that a multi-omic characterization of the same endpoint identified.
Previous analyses have profiled pre-treatment biopsies via DNA- and RNA-sequencing, together with digital pathology, finding that tumors attaining pCR were jointly high in proliferation and in immune activation~\citep{sammut2022multi}.
Here, we reproduce this result (\Cref{fig:neo_bio_correlates}a,b), notably only using the diagnostic H\&E slide.

No tool in routine clinical use supplies an estimate of pCR likelihood across breast cancer subtypes.
Multigene assays measure small targeted lists and estimate recurrence risk rather than response to NAT~\citep{paik2004multigene,sparano2020Clinical,OShaughnessy2023}.
HER2DX predicts pCR, but only in HER2$+$ disease and only with an RNA assay~\citep{prat2022development}.
Histopathological markers reproduce poorly~\citep{kos2020pitfalls,polley2013international} and are vulnerable to tumor sampling bias from needle biopsy~\citep{gyanchandani2016intratumor}.
Machine-learning predictors of NAT response, including those that predict through inferred expression~\cite{hoang2024deep,cantore2026predicting}, have meanwhile been trained on small cohorts with limited external validation~\citep{sammut2022multi,huang2023artificial,ogier2023federated,aswolinskiy2023proacting}.
NEO, by contrast, was developed on five multi-institutional cohorts and evaluated across nine.
It outperformed the computational TIL biomarkers and pathologist-assessed Ki-67 and remained associated with pCR after adjustment for clinical variables.
It is not confined to a targeted gene list either: inference supplies 14,773 genes from the slide alone.
Breadth is also what removes the need to choose well.
Small gene lists varied widely in performance depending on their composition, whereas larger random lists approached full-transcriptome performance more reliably (\Cref{fig:tech_interpretability}c).
One reason may be redundancy: ranked lists accumulate mutually correlated genes (\Cref{fig:gene_redundancy}), so growing such a gene list adds little new signal.
Building a gene list from each gene's association with pCR requires NAT labels and, when estimated on small cohorts, risks capturing spurious associations.
Prediction fidelity also does not help --- selecting the genes MORPHEUS infers most faithfully never beat random selection --- so a wide, imprecisely inferred gene list is a better substrate than a narrow, accurately inferred one.
This makes breadth a variance-reduction strategy, and one available without response labels.
However, breadth of input did not require breadth of representation: pooled AUROC did not improve beyond five components (\Cref{fig:tech_interpretability}d).
A plausible explanation is that individual inferred gene values are noisy and that reducing the gene list to a few components averages that noise away while preserving the programs that carry response signal.
A wider representation may also dilute the contribution of the clinical variables, which are few but individually strong.
NEO is also fully automated, which eliminates inter-observer variability by construction, and its prediction is stable under the biopsy sampling that confounds the histopathological markers.
End to end, scoring a patient takes approximately five minutes on average on a single modern GPU, requiring no reagent or tissue beyond the diagnostic block.

A needle core captures less than 1\% of a heterogeneous tumor, which makes stability of predictions under sampling a prerequisite.
NEO reached a stable estimate from a single slide, whereas computational TIL biomarkers required 2.5 to 5.8 slides (\Cref{fig:sampling_bias}c).
Illustrative examples suggest the same at the sub-slide level.
Independently scored cores of one slide gave similar predictions (\Cref{fig:sampling_bias}b), and the imaging-only prediction changed little under patch subsampling down to ${\approx}$6\,mm$^2$ of tissue (\Cref{fig:sampling_bias}d).
NEO predictions are robust to tumor sampling bias, whether from spatial heterogeneity or from limited tissue quantity, remaining stable within the range of variation seen with pretreatment core biopsies.

Prior work has established that gene expression may be learned from pathology slides~\cite{dey2025generating, hoang2024deep}.
Here, we expand on these efforts by establishing exactly which genomic programs are learnable from H\&E slides, and whether this varies between cancer types.
We discover that molecular signals of immune infiltration and stromal remodeling could be inferred from morphology in nearly every cancer type.
In addition, tissue-specific signals were identified: proliferation pathways were detectable in cancer types where mitotic activity is part of routine grading; estrogen response was measurable most strongly in breast cancer, and also in other hormone-driven cancer types.
Unlike previous work, we also confirm correspondence between predicted molecular signals and associated morphological features through annotation by board-certified pathologists with subspecialty expertise in breast cancer.
Several of these biological signals already inform computational or clinical approaches in pathology.
TILs, which pathologists can identify on H\&E slides, have already been the subject of numerous computational approaches~\citep{aswolinskiy2023proacting,liu2024panoptic,rijthoven2026analysis}.
The proliferation and hormone-driven features identified here have likewise been used in multivariable models predicting neoadjuvant chemotherapy response in ER$+$/HER2$-$ tumors~\citep{jerome2026magee,bhargava2021magee,farrugia2017magee}, although those models require reliable semiquantitative immunohistochemistry.
By identifying which molecular programs can be inferred from morphology and in which cancer types, our findings may help prioritize biological targets and disease settings for future AI tools in pathology.

Prior studies have also demonstrated the added utility of transcriptome-informed predictions from H\&E slides for modeling treatment responses --- versus direct pathology predictions --- in a pan-cancer cohort~\citep{hoang2024deep}, and on small datasets with dozens of breast tumors~\citep{cantore2026predicting}.
Here, we extend these efforts by testing the hypothesized advantage of transcriptome-informed pCR prediction at scale: developed using 1,080 patients (five cohorts) and evaluated in 1,412 patients (nine cohorts).
A plausible explanation is that MORPHEUS simplifies the learning problem by first compressing the image embeddings into biologically meaningful features.
Developed on 8,742 TCGA patients to predict the expression of 14,773 genes, MORPHEUS learned to relate morphology to gene expression.
The model trained on the NAT cohorts is then left with the simpler task of relating the inferred expression to response.
In the direct pCR prediction model, morphology must be related directly to response from a thousand response labels.
The strategy generalizes: pretrain the image-to-biology step on abundant paired data, then fit the clinical endpoint on the compact representation it yields, so that the endpoint model can be estimated from a modest cohort.
However, this strategy may not remain necessary as training data accumulate.
With a cohort large enough to identify which morphology matters from response labels alone, we expect the advantage of the intermediate step to narrow.

This study has limitations.
First, the evaluation cohorts are observational, with patients treated according to local practice.
They validate NEO as a predictor of pCR, however, based on the results presented in this study alone, cannot inform specific treatment decisions. This can be explored using data from clinical trials.
The PHERGain trials used response-adapted strategies to omit chemotherapy in patients with HER2$+$ early breast cancer who responded to neoadjuvant trastuzumab and pertuzumab alone, in both the randomized PHERGain trial~\citep{perez2021chemotherapy,perez20243} and the single-arm PHERGain-2 trial~\citep{garrigos2026chemotherapy}.
Our companion study validates NEO in these trials, finding that model predictions maintain performance in the setting of chemotherapy de-escalation~\citep{phergain2026neo}.
Second, three of the four computational TIL biomarkers (cTILs, LTR and ITR) rest on our own re-implementation.
We built them from the published definitions of \citet{aswolinskiy2023proacting} with a different segmentation backbone because the reference implementation could not be obtained (\Cref{sec:benchmark_biomarkers}).
Although the original authors indicated that alternative segmentation backbones could be used, differences between our reproduced and published LTR and ITR performance warrant caution when interpreting these comparisons.
muTILs, however, runs on the authors' released models and performed no better, suggesting the computational TIL biomarkers' limited discrimination is not solely due to re-implementation.
Third, NEO was benchmarked only against pathology-based computational TIL biomarkers and Ki-67, as these were the sole benchmarks available in the NAT cohorts. Future work will compare performance against genomic assays.
Fourth, the tissue-sampling analysis (\Cref{fig:sampling_bias}d) subsampled patches at random from large slides rather than evaluating genuinely small specimens.
Finally, both secondary endpoints are exploratory: RCB is sparsely assessed, and the nodal outcome is non-randomly missing.

Future work could strengthen the learned representation at its source, for example by fine-tuning the pathology foundation model on the paired gene expression data or by including gene expression in its pretraining~\citep{vaidya2025molecular,jaume2024transcriptomics,PanShi_Teaching_MICCAI2025}.
In addition, the bulk transcriptome-informed approach developed here may be extended to spatially resolved expression inference~\citep{shulman2026ai}, building on the zero-shot examples in \Cref{fig:morpheus}c.
With larger datasets, NEO could be extended with a causal module that predicts pCR under each candidate treatment while adjusting for non-random treatment allocation~\citep{biswas2026causal}.
Beyond the retrospective data presented here, prospective evaluation is the necessary next step to establish clinical utility and impact on outcomes~\citep{biswas2025prospective}.
A silent-deployment study of NEO, in which the model runs prospectively without influencing care, is registered with ClinicalTrials.gov (NCT07327970) and has begun enrollment.

\newpage
\section{Methods}
\label{sec:methods}

\subsection{Neoadjuvant cohorts and data}\label{sec:clinical_inputs}

To develop and evaluate a model for pre-treatment prediction of response to NAT, we assembled a multi-national dataset of H\&E-stained breast cancer biopsy WSIs paired with baseline clinical variables and pCR outcomes.
It spans fourteen cohorts: five used for model development ($N=1{,}080$) and nine reserved for evaluation ($N=1{,}412$; \Cref{fig:overview}b).
Available evaluation-cohort periods were 2018--2024 for Protege, 2000--2021 for Nexus, 2012--2024 for Dordrecht, 2011--2016 for IMPRESS, 2021--2024 for Jagiellonian, 2024--2026 for Shaare Zedek, 2017--2022 for Providence, and 2023--2025 for \L{}\'od\'z, based on the reported inclusion period for IMPRESS~\citep{huang2023artificial} and recorded diagnosis years for the remaining cohorts.
All patients were women with non-metastatic breast cancer who received NAT.
Patients with distant metastases at diagnosis were not included.
Every patient contributed at least one pre-treatment H\&E biopsy WSI together with a pCR outcome.
This study was conducted in accordance with the ethical principles of the Declaration of Helsinki and applicable regulatory requirements, including ICH-GCP, relevant US/Canadian human research protection regulations, and TCPS2, as applicable. Informed consent was waived due to the minimal risk to the subjects, the retrospective collection of data, and the use of only de-identified data.

The primary endpoint was pCR following NAT.
Where the source institution recorded pCR status explicitly, it was used as reported.
In the remaining cohorts, pCR was inferred from post-treatment pathological staging.
That is, a patient was classified as achieving pCR if the pathological tumor stage was ypT0/Tis and the pathological nodal stage was ypN0, and as non-pCR if residual disease was documented in the breast or the lymph nodes (ypT $>$ Tis and/or ypN $>$ 0).

Model inputs were restricted to information available before treatment.
NEO requires pre-treatment biopsy WSIs as input; resection specimens were excluded.
The clinical inputs are eight variables: receptor status (ER, PR, and HER2), clinical T stage, clinical N stage, age at diagnosis, ductal histology, and lobular histology, the last two recorded as separate indicators. Pathological stages were not used as input, as they incorporate post-surgical information and reflect NAT response.
The distribution of demographic and clinical characteristics is provided in \Cref{tab:cohorts_breakdown}.

\subsection{NEO overview}\label{sec:model-description}

NEO is a multi-modal AI model that takes pre-treatment biopsy WSIs and baseline clinical variables as inputs and estimates the probability of pCR following NAT.
In the first stage, MORPHEUS infers the expression of 14,773 genes from the WSIs.
This H\&E-inferred expression is reduced by independent component analysis (ICA) to five components, MORPHEUS-IC0 to MORPHEUS-IC4 (\Cref{sec:ica_readout}).
In the second stage, explainable boosting machines (EBMs)~\cite{nori2019interpretml,caruana2015intelligible} and tabular ResNets~\cite{gorishniy2021revisiting} map these components together with baseline clinical variables to a probability of pCR (\Cref{sec:response_model}).
Only these pCR prediction models were trained on the NAT cohorts.
Their probabilities were calibrated post hoc and averaged to produce the final NEO score (\Cref{sec:calibration,sec:backbone_combination}).

\subsection{Slide preprocessing and Falcon foundation models}\label{sec:falcon}

Each biopsy WSI was preprocessed prior to feature extraction.
A custom DeepLabV3~\citep{chen2017rethinking} segmentation model identified the tissue regions, and the tissue was divided into non-overlapping 224$\times$224-pixel patches at 0.5\,$\mu$m per pixel, yielding a median of 4,123 patches per slide (\Cref{fig:overview}c).
Each patch was then encoded by Falcon into a fixed-length embedding summarizing its morphology.
Falcon is a family of pan-cancer digital pathology foundation models trained by self-supervised learning, so that generic morphological representations could be learned from a large unlabeled H\&E corpus and re-used without task-specific annotation~\citep{song2023artificial,moor2023foundation}.
The Falcon models are successors of Kestrel~\citep{cappadona2024squeezing,witowski2026multi}, trained on a larger H\&E corpus with DINOv2~\citep{oquab2023dinov2}.

Two variants were used in this study.
First, Falcon-g is a ViT-g architecture (1.1 billion parameters, 1,536-dimensional embeddings) trained on approximately 2 billion patches from 180,000 H\&E WSIs spanning diverse cancer types and institutions.
Second, Falcon-L is a ViT-L architecture (300 million parameters, 1,024-dimensional embeddings) trained on 2 billion patches from 38,000 WSIs.
No patient contributing to the fourteen development or evaluation cohorts is represented in the self-supervised pretraining corpus of either variant, and no fine-tuning was performed on NAT data.
TCGA and CPTAC, the cohorts on which MORPHEUS was developed and evaluated, likewise did not contribute to Falcon training.
Falcon parameters were frozen throughout, so both are fixed feature extractors for everything that follows.

\subsection{Gene expression prediction dataset}\label{subsec:rna_expression_prediction_dataset}

MORPHEUS was developed on a pan-cancer dataset spanning 32 TCGA projects and evaluated on 10 CPTAC cancer types, leveraging the matched gene expression profiles and histopathology WSIs available in both cohorts.
The cohort breakdowns are provided in \Cref{tab:morpheus_breakdown_tcga,tab:morpheus_breakdown_cptac}.
We kept only the RNA-seq profiles corresponding to primary tumor and discarded other profiles corresponding to normal tissues, recurrent tumor, or metastatic cancer.
We then paired each bulk RNA-seq profile with the WSIs of the tissue it came from, at the finest level of attribution that we resolved for each cohort.
In CPTAC, every WSI records the specimen that was sequenced, so pairing was done per RNA-seq sample. The 10 cancer types contribute 1,566 primary-tumor samples from 1,473 patients (41 patients contributed 2 to 5 samples each), and those samples carry 1,845 WSIs.
For TCGA, we did not resolve an equivalent attribution when assembling the dataset, so patients with more than one primary-tumor RNA-seq sample were excluded and pairing is done per patient (\Cref{sec:fidelity_metric}).
After an 80:20 patient-level training-validation split, 6,994 TCGA patients (8,431 WSIs) were used for training MORPHEUS and the remaining 1,748 patients (2,089 WSIs) were used for hyperparameter tuning and model selection, together 8,742 patients and 10,520 WSIs across 32 projects (\Cref{fig:overview}d).
Gene expression processing, comprising expression-level filtering, a variance-stabilizing transformation (VST), and per-gene standardization, is described in \Cref{sec:expression_preprocessing}; it left 14,773 protein-coding genes for the analyses reported here.

\subsection{MORPHEUS: gene expression prediction from H\&E}\label{sec:gene_expression_model}

MORPHEUS consists of two separate gene expression prediction models, one trained with each Falcon variant (Falcon-g and Falcon-L).
Falcon outputs thousands of patch-level embeddings per slide, and which of them carry the signal is not known in advance: the labels are attached to the slide, not to any region within it.
We therefore aggregate them into a fixed-dimensional slide-level representation with a MIL aggregator built on gated attention~\citep{ilse2018attention} (\Cref{sec:gated_attention_detail}; \Cref{fig:model_components}).
The model then predicts all 14,773 gene expression values jointly.
The full training objective, including its regularization term and balancing weights, is given in \Cref{sec:morpheus_objective}, and the optimizer settings, model-selection rule and hyperparameter search in \Cref{sec:optimization_detail,sec:hyperparameters}.
The fidelity of the expression MORPHEUS infers is quantified as described in \Cref{sec:fidelity_metric} and reported per gene and per tumor type in \Cref{fig:ed_morpheus_fidelity}; the transcriptional programs it recovers are characterized in \Cref{fig:morpheus}.

\subsection{Use of H\&E-inferred expression}\label{sec:ica_readout}

MORPHEUS was applied to histopathology slides from patients in the NAT cohort to generate slide-level H\&E-inferred expression for 14,773 genes.
Multiple slide-level H\&E-inferred expressions belonging to the same patient were averaged to obtain patient-level H\&E-inferred expression.
To reduce dimensionality, we restricted the input to the gene list curated by~\citet{chen2021multimodal}, which comprises six gene families from MSigDB~\citep{subramanian2005gene,liberzon2011molecular}: tumor suppressor genes, oncogenes, protein kinases, cell differentiation markers, transcription factors, and cytokines/growth factors.
This gene list was fixed before the main experiments.
Of the 3,047 unique genes in this list, 2,427 were present in our 14,773-gene set after expression-level filtering.
The H\&E-inferred expression for the 2,427 genes was transformed into a five-dimensional representation, using an ICA model fitted on the TCGA validation subset with the FastICA algorithm~\citep{hyvarinen2000independent}.
Because the ICA reconstruction is linear, each component has a direct interpretation in gene space.
The corresponding column of the mixing matrix holds the weight with which that component reconstructs each gene's H\&E-inferred expression, so ranking genes by the magnitude of these weights lists the genes the component most strongly represents.
The pathway analysis of the two most important MORPHEUS-ICs ranked genes by this weight vector (\Cref{sec:ic_pathways}).
The five components, together with the baseline clinical variables, form the input to the pCR prediction model (\Cref{sec:response_model}).
What each component contributes to the prediction is read off in \Cref{fig:tech_interpretability}b, and the choice of reduction method and of the number of components is examined in \Cref{fig:tech_interpretability}d (\Cref{sec:reduction_ablation}).

\subsection{Multiple-source cross-validation}\label{subsec:mscv-main}
All development of the pCR prediction model (hyperparameter tuning, model selection, and calibration) used a multiple-source cross-validation (MSCV) protocol~\citep{geras2013multiple} designed for developing models with heterogeneous multi-cohort data.
Development cohorts were partitioned into a set of \emph{train-only} cohorts $\mathcal{D}^T = \{D_1, \ldots, D_J\}$ used only for training, and a set of \emph{cross-validation} cohorts $\mathcal{D}^V = \{D_{J+1}, \ldots, D_{J+K}\}$, each of which was held out in turn.
Each cross-validation cohort defines one fold: for $k = 1, \ldots, K$, models are fit on $\mathcal{D}^{\mathrm{tr}}_{k} = (\mathcal{D}^T \cup \mathcal{D}^V) \setminus D_{J+k}$ and validated on the single held-out cohort $\mathcal{D}^{\mathrm{val}}_{k} = D_{J+k}$.
On each fold, the held-out cohort is therefore unseen during training and drawn from exactly one source.

Relative to a single random patient-level split, this prevents a model from inflating validation performance in two ways: by memorizing cohort-specific artifacts, and by scoring well through discriminating between cohorts of differing prevalence.
Neither shortcut is available when the held-out cohort is both unseen and drawn from a single source.
For NEO, the train-only set $\mathcal{D}^T$ comprises the TransNEO and Yale cohorts ($J = 2$), which enter the training set of every fold, and the cross-validation set $\mathcal{D}^V$ comprises the Omica, NYU, and Gundersen cohorts ($K = 3$), each of which is held out in turn.
Model development therefore uses three folds.
\Cref{alg:mscv} describes how models are trained, selected, ensembled and calibrated according to the MSCV protocol.

\subsection{pCR prediction model}\label{sec:response_model}

The final NEO pCR probability prediction is produced by an ensemble of two model classes: EBMs and tabular ResNets.
Throughout, a trained tabular ResNet or MIL model that outputs a pCR probability is referred to as a neural-network-based pCR prediction model.
Both take the same input variables: the baseline clinical variables (\Cref{sec:clinical_inputs}) together with the five MORPHEUS-ICs of the H\&E-inferred expression, formed separately for each Falcon variant.
Clinical-input encoding and missing-value handling are described in \Cref{sec:clinical_encoding}.
Thus, every model built on top of Falcon was built twice, once on Falcon-g features and once on Falcon-L features (\Cref{sec:backbone_combination}).
For each Falcon variant, model class, and fold, the fitted model was calibrated on its own held-out cohort (\Cref{sec:calibration}), and the calibrated predictions are averaged over folds, model classes, and Falcon variants to give the final NEO output.
That output is the NEO score reported throughout: its discrimination and calibration in \Cref{fig:performance}, its stability under biopsy sampling in \Cref{fig:sampling_bias}, and the two secondary endpoints in \Cref{fig:secondary_endpoints}.
\Cref{alg:atxneo} states the procedure end to end. The architecture diagram for the tabular ResNet is drawn in \Cref{fig:model_components}.

The tabular ResNet~\citep{gorishniy2021revisiting} was trained to predict pCR jointly with auxiliary targets for post-treatment pathological stage---ypT, ypN, and ypT/ypN downstaging. These auxiliary targets give a finer-grained training signal, since pCR dichotomizes residual disease and discards the gradation that ypT and ypN record. They serve as training signal only: their predictions are neither reported nor evaluated here. Definitions of the auxiliary targets and the full multitask objective are given in \Cref{sec:multitask_detail}.
The same formulation was used by every neural network trained on the pCR outcome in this work, including the two direct pCR prediction models (\Cref{sec:direct_path_model,fig:model_components}).

EBMs~\citep{nori2019interpretml,caruana2015intelligible} are interpretable by construction and competitive with black-box models in their performance, and were trained to predict pCR alone, on the same input features and MSCV folds. The mechanism that enforces the MSCV train/validation split inside EBM training is described in \Cref{sec:ebm_mscv_integration}.
The two summaries of a fitted EBM, feature importance and shape functions (\Cref{fig:tech_interpretability}b), are defined in \Cref{sec:ebm_interpretation}.

Optimizer settings, the hyperparameter search protocol, and the model-selection rule for both classes are given in \Cref{sec:optimization_detail} and \Cref{tab:hyperparameters}.

\subsection{Model evaluation}\label{sec:method_eval}
Model performance was quantified primarily using the area under the receiver operating characteristic curve (AUROC), which summarizes discriminative ability across all classification thresholds.
AUROCs are reported for each cohort individually and pooled across cohorts with a random-effects model.
We pool performances across cohorts with a random-effects model rather than pooling patient predictions for the same reason that MSCV validates on single-source cohorts (\Cref{subsec:mscv-main}).
An AUROC computed over patients pooled across cohorts could reward separating cohorts with different pCR prevalence, whereas a random-effects model measures discrimination within each cohort and treats the cohort as the unit of replication.
Performance was further stratified by molecular subtype.
Each subtype estimate is itself the result of a random-effects model over cohorts.
This procedure produces every pooled AUROC reported in this work: the discrimination of NEO overall and within subtype (\Cref{fig:performance}a, with the per-cohort estimates in \Cref{fig:cohort_forests}), the comparisons against the computational TIL biomarkers and Ki-67 (\Cref{fig:performance}d,e and \Cref{fig:subtype_biomarkers}), the ablations of the representation given to the pCR prediction model (\Cref{fig:tech_interpretability}a,c,d), and the two secondary endpoints (\Cref{fig:secondary_endpoints}b,c).

A cohort entered a pooled estimate only if it contributed at least thirty positive--negative pairs for the endpoint being evaluated.
For pCR, each pair contains one patient who achieved pCR with one who did not.
Below that, an AUROC rests on too few comparisons to be stable, and when the score separates the few patients perfectly, its bootstrap interval can degenerate to an interval of 1 to 1.
The threshold was fixed before these analyses and applied identically to every score, so the cohorts behind a pooled estimate are the same for NEO and for each benchmark biomarker.
Which cohorts the rule excludes from each subtype-specific estimate, and how many therefore enter it, are given with the per-cohort estimates in \Cref{fig:cohort_forests}.

Per-cohort AUROC confidence intervals were computed by nonparametric bootstrap over patients, and cohort estimates were pooled using a DerSimonian--Laird random-effects model~\cite{dersimonian1986meta} with the Hartung--Knapp adjustment~\cite{inthout2014hartung}.
How each pooled quantity supplies its standard error, and why the adjustment is used, are given in \Cref{sec:pooling_detail}.

\subsection{Calibration assessment}\label{sec:calibration_assessment}
Discrimination measures only whether the score orders patients correctly.
Calibration asks the separate question of whether the predicted probabilities, on average, match observed outcome rates.
Calibration of NEO predictions was assessed over the combined evaluation set (all nine cohorts, $N=1{,}412$), and within each molecular subtype.
Patients were grouped into equal-count (quantile) bins of predicted risk.
We report four bins in \Cref{fig:performance}b.
For each bin we plot the mean predicted probability against the observed pCR rate, with a Wilson score interval~\citep{wilson1927probable} on the observed rate.

We report the standard logistic recalibration summaries~\citep{vancalster2019calibration}. For outcome $Y\in\{0,1\}$ and predicted pCR probability $\hat{p}$, the calibration slope is the coefficient $b$ in $\mathrm{logit}\{\Pr(Y=1\mid\hat{p})\} = a + b\,\mathrm{logit}(\hat{p})$: $b<1$ indicates predictions that are too extreme, and $b>1$ predictions that are too compressed. The calibration intercept is refitted with the slope held at 1, making $a$ a shift in level. $a>0$ indicates under-prediction of response.
Both were fitted by maximum likelihood on the combined patient set of each stratum, with Wald 95\% confidence intervals.

\subsection{Adjustment for baseline clinical variables}\label{sec:adjusted_analysis}
To test whether the NEO score carries information beyond the standard clinical variables, we fitted a Firth-penalized multivariable logistic regression~\citep{firth1993bias,heinze2002solution} of pCR on the score, adjusted for the baseline clinical variables listed below, separately within each evaluation cohort.
The Firth penalty reduces the small-sample bias of maximum likelihood, which matters here because several cohorts have fewer than five events per fitted parameter.
The resulting log odds ratios were pooled across cohorts using a DerSimonian--Laird random-effects model with the Hartung--Knapp adjustment (\Cref{sec:method_eval}). \Cref{fig:performance}c shows each cohort's estimate and the pooled result.
The adjustment set was pre-specified and clinically motivated, with no data-driven variable selection: the clinical inputs to NEO (\Cref{sec:clinical_inputs}) as well as histologic grade.
Age entered as a continuous variable; grade, T stage, and N stage as ordinals; and receptor status and ductal or lobular histology as binary indicators.

A patient entered the analysis only if all nine covariates were observed.
As a result, 1,031 of 1,412 patients were analyzed, and two cohorts that contributed no complete cases at all were excluded (IMPRESS records no clinical stage, and Providence no histology).
Covariate coding, the handling of rank-deficient covariates within a cohort, and the fitting software are given in \Cref{sec:adjusted_detail}.
Odds ratios are expressed per 0.1-unit increase in the score. Because the score is a calibrated probability, this unit corresponds to a 10-percentage-point difference in predicted pCR probability and means the same thing in every cohort, so the pooled estimate has a single interpretation.

\subsection{Measuring fidelity of H\&E-inferred expression}\label{sec:fidelity_metric}
For per-gene expression-prediction fidelity (\Cref{fig:ed_morpheus_fidelity}, \Cref{fig:morpheus}a), the unit of analysis is the case: a patient in TCGA and a primary-tumor RNA-seq sample in CPTAC (\Cref{subsec:rna_expression_prediction_dataset}).
Fidelity was computed per gene as the Pearson correlation between predicted and measured expression across the cases of a single cancer type (\Cref{fig:ed_morpheus_fidelity}).
Following~\citet{hoang2024deep}, it is summarized in two steps: the median over genes within each type, then the mean of those medians across types.
Types with fewer than 20 evaluated cases are omitted from the per-type summaries, as a per-gene correlation over so few cases is too unstable to summarize.
This excluded 9 of the 32 TCGA projects represented in the validation subset, leaving 23 projects and 1,642 of 1,748 patients (\Cref{tab:morpheus_breakdown_tcga}), whereas all 10 CPTAC cohorts clear the threshold.
\Cref{fig:tech_interpretability}c and \Cref{fig:gene_redundancy} use fidelity differently.
For gene ranking, fidelity is Kendall's $\tau$ over all cancer types combined in the TCGA validation subset, matching the model-selection criterion used in MORPHEUS training itself.

\subsection{Biological correlates}\label{sec:bio_correlates_methods}

To characterize which transcriptional programs are learnable from morphology (\Cref{fig:morpheus}a), slide-level predictions were first averaged to one profile per patient.
This analysis was restricted to TCGA projects with at least 20 cases (\Cref{sec:fidelity_metric}).
Within each such project, we computed the Pearson correlation between predicted and measured expression across that project's patients, separately for each of the 14{,}773 genes, and ranked the genes by that correlation.
This ranking was used for GSEA with the fgsea R package, evaluating enrichment against the hallmark gene set collection. Here, a positive normalized enrichment score (NES) indicates that a pathway's genes were predicted more accurately than the transcriptome-wide background.

To relate the recovered programs to morphology (\Cref{fig:morpheus}b), we selected the Jagiellonian cohort as a representative cohort from the evaluation set.
MORPHEUS was applied separately to 100 randomly sampled 224$\times$224-pixel patches (\Cref{sec:falcon}) from each slide.
A patch's score for a hallmark set is the mean H\&E-inferred expression over all set members present in the 14{,}773-gene output.
Two board-certified pathologists with subspecialty expertise in breast pathology independently reviewed the highest- and lowest-scoring patches for each of three programs (EMT, mitotic spindle, and angiogenesis) and reported whether the high set showed more of that biology than the low set.

\subsection{Robustness to biopsy sampling}\label{sec:sampling_robustness}

We assessed the stability of predictions to biopsy sampling on the UChicago cohort.
This cohort was selected as it contributed the largest number of slides per patient (289 patients, 1,269 slides).
Two nested patient sets were used.
The intraclass-correlation analysis (\Cref{fig:sampling_bias}a) concerned NEO alone, and uses all 247 patients with $\ge 2$ slides.
The variance-decomposition comparison (\Cref{fig:sampling_bias}c) required every method to return a value on every slide, so it was restricted to the 242 of those patients in the shared-slide set (\Cref{sec:shared_slides}), on which NEO and computational TIL biomarkers were scored on identical slides.
We did not impose that restriction on the intraclass correlation: the shared-slide set is defined by where the segmentation or detection pipelines of the computational TIL biomarkers fail, and those failures are not random with respect to tissue content, so a NEO-only estimate is better left unconditioned on them.
For patients with $\ge 2$ slides we computed the one-way random-effects intraclass correlation coefficient ICC(1,1)~\citep{shrout1979intraclass} of the slide-level predictions, treating the patient as the random target and that patient's slides as interchangeable repeated measurements.
The estimator, its correction for unequal numbers of slides per patient, and the bootstrap interval are stated in \Cref{sec:sampling_estimators}.

We additionally applied the variance-decomposition metric developed by \citet{bachtiary2006gene}, which has been used by \citet{biswas2025prospective} to benchmark biomarkers according to their robustness to tumor sampling bias (\Cref{fig:sampling_bias}c).
$W/T$ is the fraction of a biomarker's total variance attributable to sampling when $k$ slides are averaged.
The number of slides required for a stable estimate is the value of $k$ at which $W/T$ reaches the threshold of $0.15$ adopted by \citet{bachtiary2006gene}, reported as a fractional number of slides.
The repeatedly sampled unit here is the H\&E slide, in place of the spatially distinct multi-region biopsies of the original studies.
A patient's slides derive from routine diagnostic sampling and are not guaranteed to be spatially independent tumor regions.
The estimators of the two variance components and the exact crossing $k^{*}$ are given in \Cref{sec:sampling_estimators}.
Every score entering the decomposition is a slide-level score, so NEO was run end-to-end on each slide independently rather than through its multi-slide aggregation mechanism, which averages a patient's H\&E-inferred expression vectors before the ICA projection.

\subsection{Imaging-only variant and patch subsampling}\label{sec:imaging_only_stability}
The imaging-only variant is identical to NEO in architecture, protocol and ensembling, but was trained without the clinical variables, so that it is scored from the image alone.
This enables a like-with-like comparison to the computational TIL biomarkers.
This imaging-only baseline enters the variance decomposition of \Cref{fig:sampling_bias}c and the patch-subsampling analysis of \Cref{fig:sampling_bias}d, and is reported nowhere else.

To characterize how the prediction depends on the amount of tissue analyzed (\Cref{fig:sampling_bias}d), we selected every evaluation slide with at least 10,000 patches of 224$\times$224 pixels retained after tissue segmentation (70 slides from 62 patients across the Shaare Zedek, IMPRESS, \L{}\'od\'z, Jagiellonian, Dordrecht, UChicago, and Providence cohorts).
For each slide and each size $n \in \{50, 100, 200, 500, 1000, 2000, 5000\}$ we drew 30 random subsets of $n$ patches without replacement, using identical subsets under both foundation-model embeddings, matched by patch coordinates.
For each subset we recorded the absolute value of the relative change in predicted probability of pCR from that slide's all-patch prediction.
These were averaged over the 30 subsets, giving one value per slide and size, and we report the median of these values across slides at each size (\Cref{fig:sampling_bias}d).
We ran the imaging-only variant on each subset exactly as on a whole slide, and computed the relative change of the calibrated pCR probability with respect to the prediction obtained from all of the slide's patches.

\subsection{Statistical analysis}\label{sec:reporting}
For the clinical-performance analyses, two families of $p$ values are reported: two-sided Hartung--Knapp $t$ tests and a one-sided Jonckheere--Terpstra test.
Tests for pathway enrichment and molecular correlates are described separately in \Cref{sec:bio_correlates_methods,sec:neo_correlates_methods}.
Pooled AUROCs were tested against chance (0.5), pooled paired differences between two scores evaluated on the same patients (\Cref{sec:paired_diff}) against zero, and pooled odds ratios against 1, by the Hartung--Knapp $t$ test on $k-1$ degrees of freedom described in \Cref{sec:pooling_detail}.
These are the $p$ values quoted for the adjusted odds ratio (\Cref{fig:performance}c), for the comparisons against the computational TIL biomarkers and Ki-67 (\Cref{fig:performance}d,e and \Cref{fig:paired_comparisons}a--e), and for the ablations comparing transcriptome-informed with direct and clinical-only pCR prediction models (\Cref{fig:tech_interpretability}a and \Cref{fig:paired_comparisons}f--h).
The trend in the score across RCB classes was assessed using a cohort-stratified Jonckheere--Terpstra test (\Cref{fig:secondary_endpoints}a; \Cref{sec:rcb_ordinal}).
All are two-sided except the Jonckheere--Terpstra test, which is one-sided because the direction was fixed in advance: the score is built to fall as residual disease increases.
All other quantities are reported as estimates with 95\% confidence intervals, and every bootstrap procedure in this work used 2,000 replicates.
The number of patients entering each analysis is annotated in the corresponding figure, and confidence intervals are reported as (95\% CI, lower--upper) throughout.
As the data were collected retrospectively, no statistical methods were used to predetermine sample size, and no randomization was performed.
Investigators were not blinded to cohort identity during model development.
The evaluation cohorts were held out from training, model selection, and calibration, but were not sequestered from analysis.
Software versions and hardware are listed in \Cref{sec:methods:eval:stats}.

\newpage

\bibliography{references}
\bibliographystyle{unsrtnat}

\newpage
\section*{Data availability}

The datasets analyzed in this study were obtained from 16 cohorts under data use agreements that in most cases preclude public redistribution of individual-level data. Requests for access to each cohort should be directed to the respective data custodians, subject to the terms of the original agreements and applicable ethical and legal frameworks.
Publicly available cohorts can be accessed directly.
TCGA and CPTAC data, including diagnostic H\&E whole-slide images and matched bulk RNA-seq, are available from the NCI Genomic Data Commons (\url{https://portal.gdc.cancer.gov}) and The Cancer Imaging Archive (\url{https://www.cancerimagingarchive.net}).
The Yale cohort~\citep{farahmand2022her2rois} is available from The Cancer Imaging Archive, the TransNEO cohort~\citep{sammut2022multi} from the European Genome-phenome Archive (EGAS00001004582), the IMPRESS cohort~\citep{huang2023artificial} from \url{https://tinyurl.com/IMPRESS-DATA}, the Providence cohort~\citep{brca-psj-path} from the Nightingale Open Science platform (\url{https://doi.org/10.48815/N5159B}), and the Nexus cohort~\citep{nexus} from \url{https://healthdatanexus.ai/content/breastcancerimaging/1.0.0/}.

\section*{Code availability}

The complete codebase underlying the AI model is proprietary and cannot be publicly released. The model and selected components of the code may be made available to academic researchers for non-commercial research purposes, subject to institutional approval and execution of appropriate data use and/or material transfer agreements. Requests will be evaluated on a case-by-case basis to ensure compliance with data privacy, intellectual property, and commercial considerations. To facilitate reproducibility, we provide detailed methodological descriptions in the Methods section, including model architectures, training procedures, and hyperparameter selection strategies, and we refer to all open-source libraries and frameworks used in the implementation. In addition, the code used to generate the main figures and aggregate results presented in this study can be made available upon reasonable request.

\section*{Acknowledgements}

The results published here are in part based upon data generated by the TCGA Research Network (\url{https://www.cancer.gov/tcga}).
This study includes data generated by the National Cancer Institute Clinical Proteomic Tumor Analysis Consortium (CPTAC).
We acknowledge the TransNEO study investigators, the Caldas Laboratory at the Cancer Research UK Cambridge Institute, and the patients who participated in the TransNEO study as the sources of the data used in this work~\citep{sammut2022multi}. Data were obtained from the European Genome-phenome Archive (EGA) under accession EGAS00001004582.
We acknowledge the University of Chicago for providing the UChicago cohort~\citep{howard2025integration}.
We acknowledge Pieter Westenend as a provider of data from Dordrecht.
We acknowledge William Tran as the data holder for Nexus cohort~\citep{nexus}.

\section*{Author contributions}

JP, DB, JW, and KJG contributed to study conception and design.
JP, JCa, KGZ, and KJG contributed to the design of the models used in this study.
JP, DB, CT, CL, and KJG contributed to the evaluation methods used in this study.
PT, CO, FJE, RB, MBr, AG, FH, PW, BC, FS, and JW provided clinical guidance.
RC and CL provided statistical guidance.
JP, JCa, and KGZ wrote code, developed infrastructure and trained models throughout the study.
BM worked on data preparation.
JP and CT performed evaluation and analysis.
JP, DB, JCa, CT, KGZ, and KJG worked on drafting and revising the manuscript.
All authors critically reviewed the paper and the results and approved the final version.

\section*{Competing interests}

JP, DB, JCa, CT, KGZ, BM, CL, FJE, FS, RC, JW and KJG are equity holders of Ataraxis AI.
JP, DB, JCa, KGZ, JW, and KJG are inventors on a US patent application filed corresponding to some of the methodological aspects of this work.
The remaining authors declare no competing interests.

\newpage

\section{Extended Data}
\setcounter{figure}{0}
\renewcommand{\theHfigure}{ed.\arabic{figure}}
\renewcommand{\thefigure}{\arabic{figure}}
\captionsetup[figure]{name=Extended Data Fig.}
\crefalias{figure}{edfig}
\counterwithin{table}{section}
\counterwithin{algorithm}{section}

\begin{figure}[H]
    \centering
    \includegraphics[width=\linewidth]{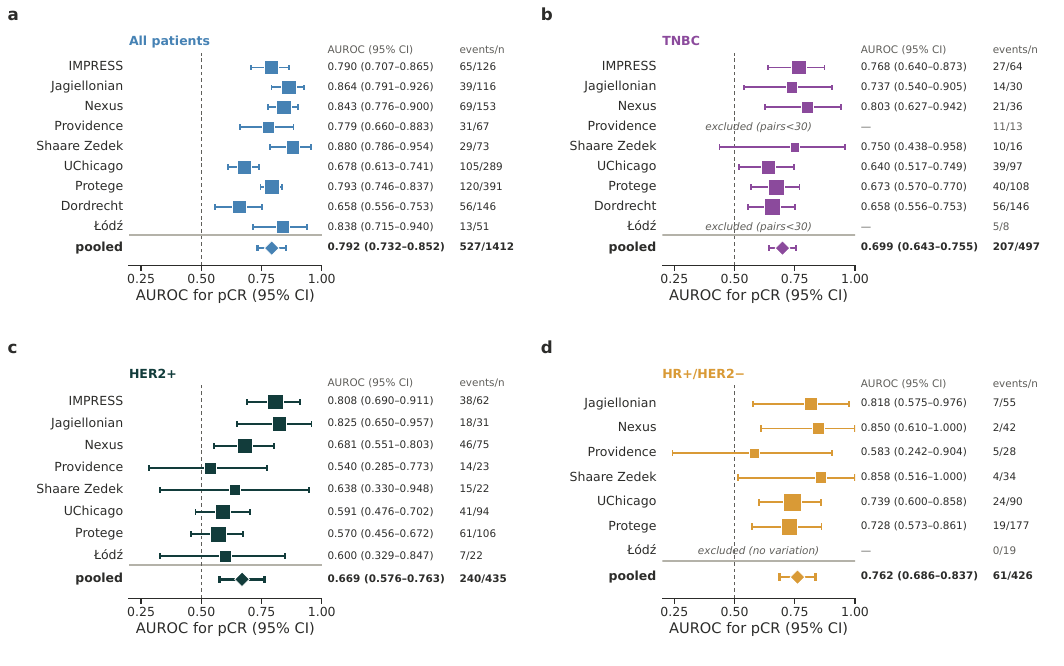}
    \caption{
    \scriptsize
    \textbf{NEO subgroup analyses by cohort and subtype.}
    Model discrimination, measured using area under the receiver operating characteristic curve (AUROC) for pathological complete response (pCR), was assessed within individual evaluation cohorts and molecular subtypes.
    Cohort estimates are shown as squares and pooled estimates as diamonds.
    Square area is proportional to the cohort's random-effects weight; horizontal bars indicate 95\% confidence intervals (CIs).
    The right-hand column reports pCR events and patient counts.
    Cohorts contributed to pooling only if they contained at least 30 positive--negative patient pairs (\Cref{sec:method_eval}); excluded cohorts and reasons are indicated.
    \textbf{a}, All patients (nine cohorts), individual AUROCs 0.658--0.880 and pooled AUROC 0.792 (95\% CI, 0.732--0.852); $I^2$ = 74\%, $\tau^2$ = 0.0040.
    \textbf{b}, Triple-negative breast cancer (TNBC,  seven cohorts), individual AUROCs 0.640--0.803 and pooled AUROC 0.699 (95\% CI, 0.643--0.755); $I^2$ = 0\%, $\tau^2$ = 0. \L{}\'od\'z and Providence were excluded because they contributed only 15 and 22 positive--negative pairs, respectively.
    \textbf{c}, Human epidermal growth factor receptor 2-positive (HER2$+$, eight cohorts), individual AUROCs 0.540--0.825 and pooled AUROC 0.669 (95\% CI, 0.576--0.763); $I^2$ = 58\%, $\tau^2$ = 0.0077. Dordrecht contributed no HER2-positive patients.
    \textbf{d}, Hormone receptor-positive, HER2-negative (HR$+$/HER2$-$, six cohorts), individual AUROCs 0.583--0.858 and pooled AUROC 0.762 (95\% CI, 0.686--0.837); $I^2$ = 0\%, $\tau^2$ = 0. \L{}\'od\'z had no pCR events in this subtype. IMPRESS and Dordrecht contribute no patients of this subtype.
    Statistics: the cohort-level CIs used 2,000 bootstrap resamples. Pooled estimates used random-effects models (\Cref{sec:method_eval}). $I^2$ and $\tau^2$ summarize between-cohort heterogeneity.
    }
    \label{fig:cohort_forests}
\end{figure}

\begin{figure}[H]
    \centering
    \includegraphics[width=\linewidth]{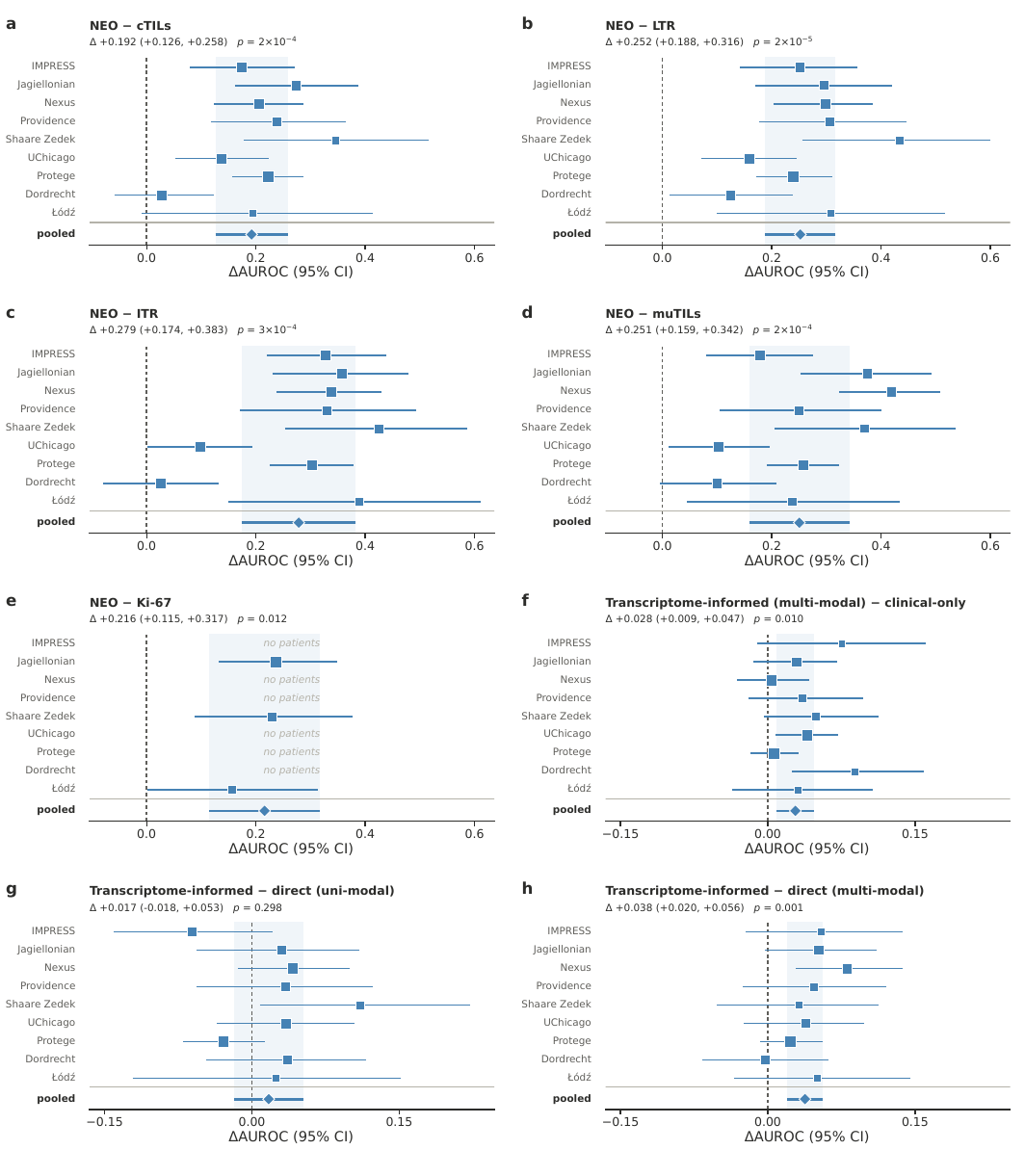}
    \caption{
    \scriptsize
    \textbf{Ablation studies of NEO and benchmark biomarkers.}
    Each panel compares the area under the receiver operating characteristic curve (AUROC) for predicting pathological complete response (pCR) between the two predictors named in the panel title.
    Positive differences favor the first predictor and negative differences favor the second.
    \textbf{a}--\textbf{e}, NEO versus computational tumor-infiltrating lymphocytes (cTILs, \textbf{a}), lymphocytes-to-tumor ratio (LTR, \textbf{b}), inflamed tumor ratio (ITR, \textbf{c}), muTILs (\textbf{d}), and Ki-67 (\textbf{e}).
    \textbf{f}, The multi-modal transcriptome-informed pCR prediction model versus the clinical-only baseline.
    \textbf{g},\textbf{h}, The transcriptome-informed versus direct pCR prediction models in the uni-modal (\textbf{g}) and multi-modal (\textbf{h}) settings.
    Squares show cohort-level differences, with area proportional to the random-effects weight and horizontal bars indicating 95\% confidence intervals (CIs) from 2,000 paired patient-level bootstrap resamples.
    Diamonds show pooled differences from random-effects models across contributing cohorts, with 95\% CIs (\Cref{sec:paired_diff}).
    Shaded bands repeat the pooled CIs, and dashed vertical lines mark zero difference.
    Panel headings report the pooled difference, CI, and two-sided $p$ value for the null hypothesis of no difference (\Cref{sec:paired_diff}).
    Neural-model comparisons (\textbf{f}--\textbf{h}) used 1,412 patients across nine cohorts.
    Comparisons with computational TIL biomarkers (\textbf{a}--\textbf{d}) used the shared-slide population of 1,386 patients across nine cohorts, with all methods scored on identical patients and slides (\Cref{sec:shared_slides}).
    The Ki-67 comparison (\textbf{e}) used 237 patients with recorded Ki-67 across three cohorts (\Cref{sec:Ki-67}).
    }
    \label{fig:paired_comparisons}
\end{figure}

\begin{figure}[H]
    \centering
    \includegraphics[width=\linewidth]{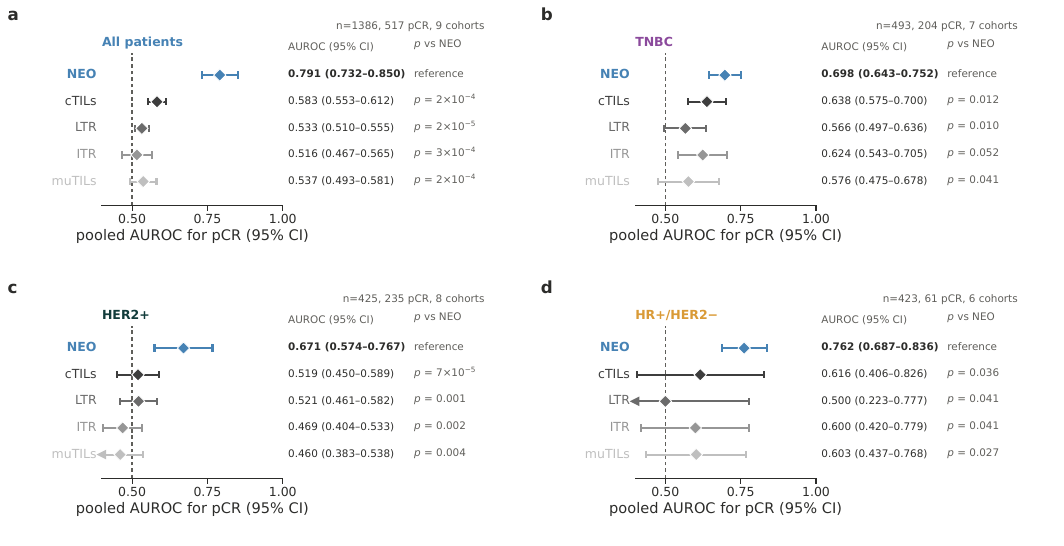}
    \caption{
    \scriptsize
    \textbf{NEO and benchmark biomarker subgroup analyses by molecular subtype.}
    Pooled area under the receiver operating characteristic curve (AUROC) for pathological complete response (pCR),  comparing NEO with computational tumor-infiltrating lymphocytes (cTILs), lymphocytes-to-tumor ratio (LTR), inflamed tumor ratio (ITR), and muTILs.
    All methods were evaluated on identical patients and slides within each panel.
    The overall shared-slide population comprised 1,386 patients (\Cref{sec:shared_slides}).
    Subtype panels include only cohorts meeting the minimum of 30 positive--negative patient pairs (\Cref{sec:method_eval}).
    \textbf{a}, All patients ($n=1,386$, nine cohorts).
    \textbf{b}, Triple-negative breast cancer (TNBC; $n=493$, seven cohorts).
    \textbf{c}, Human epidermal growth factor receptor 2-positive (HER2$+$) disease ($n=425$, eight cohorts).
    \textbf{d}, Hormone receptor-positive, HER2-negative (HR$+$/HER2$-$) disease ($n=423$, six cohorts).
    Points show pooled AUROCs from random-effects models across contributing cohorts, with 95\% confidence intervals (\Cref{sec:method_eval}).
    The right-hand column reports two-sided $p$ values from pooled paired AUROC differences between each biomarker and NEO; cohort-level standard errors used 2,000 paired patient-level bootstrap resamples (\Cref{sec:paired_diff}).
    Arrowheads indicate confidence intervals extending beyond the plotted range; their exact bounds are reported in the value column.
    }
    \label{fig:subtype_biomarkers}
\end{figure}

\begin{figure}[H]
    \centering
    \includegraphics[width=\linewidth]{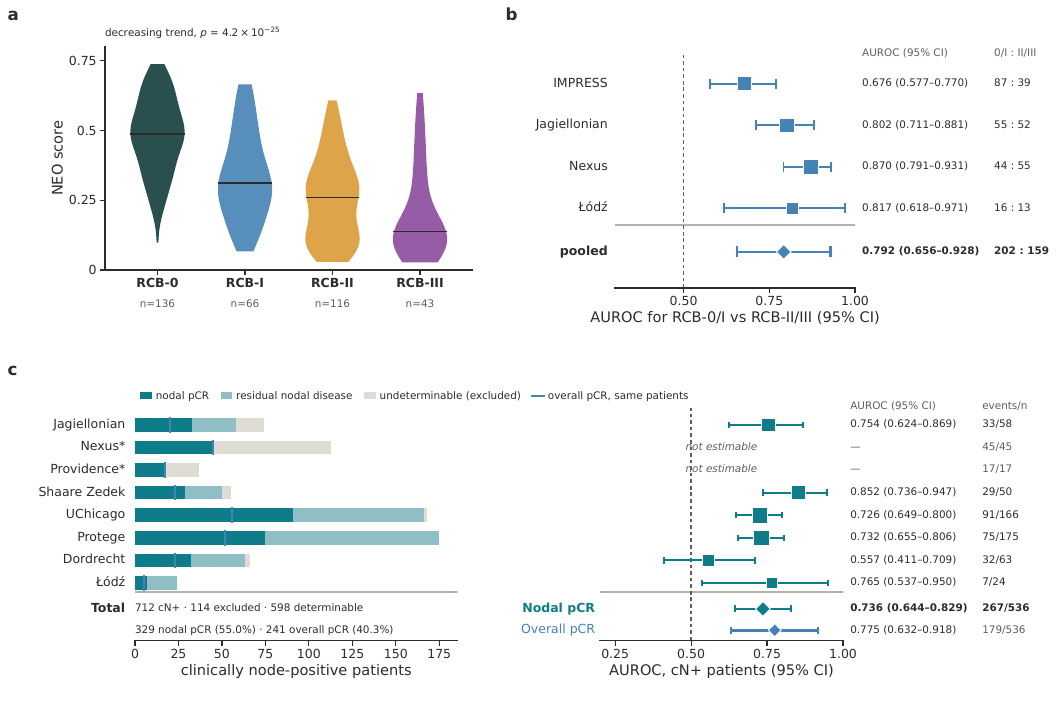}
    \caption{
    \scriptsize
    \textbf{Transfer of NEO to residual cancer burden and nodal pCR prediction.}
    The NEO score trained to predict overall pathological complete response (pCR) was evaluated against secondary endpoints without retraining or recalibration.
    \textbf{a}, NEO score by residual cancer burden (RCB) class in 361 patients across four cohorts.
    Violin plots span the observed range of scores and the horizontal ticks indicate the median: 0.486, 0.312, 0.259 and 0.139 for RCB-0, RCB-I, RCB-II, and RCB-III.
    Patient counts are shown below each class.
    The one-sided $p$ value tests decreasing NEO scores with increasing RCB using a cohort-stratified Jonckheere--Terpstra test (\Cref{sec:rcb_ordinal}).
    \textbf{b}, Area under the receiver operating characteristic curve (AUROC) for RCB-0/I versus RCB-II/III in the same 361 patients.
    Squares show cohort estimates, with area proportional to the random-effects weight.
    The diamond shows the pooled AUROC of 0.792 (95\% CI, 0.656--0.928).
    The right-hand column gives the counts in the two outcome groups.
    \textbf{c}, Nodal-pCR analysis among 712 patients with clinical nodal involvement at baseline.
    Stacked bars show nodal pCR, residual nodal disease, and indeterminate outcomes; blue ticks mark overall pCR among patients with determinable nodal outcomes.
    Nexus ($n=45$) and Providence ($n=17$), marked with asterisks, had only nodal-pCR outcomes and therefore did not meet the minimum of 30 positive--negative patient pairs required for AUROC pooling (\Cref{sec:method_eval}).
    Among the remaining 536 patients across six cohorts, pooled AUROC was 0.736 (95\% CI, 0.644--0.829) for nodal pCR and 0.775 (95\% CI, 0.632--0.918) for overall pCR.
    Nodal pCR was defined by documented ypN0 or overall pCR (\Cref{sec:nodal_pcr}).
    In \textbf{b} and the right side of \textbf{c}, horizontal bars show cohort-level 95\% CIs from 2,000 patient-level bootstrap resamples.
    Diamonds show pooled estimates from random-effects models across contributing cohorts, with 95\% CIs (\Cref{sec:method_eval}).
    }
    \label{fig:secondary_endpoints}
\end{figure}

\begin{figure}[H]
    \centering
    \includegraphics[width=\linewidth]{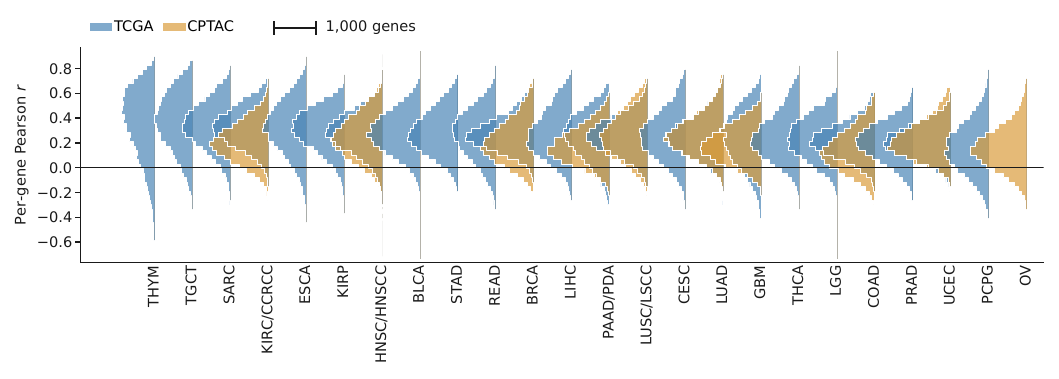}
    \caption{
    \scriptsize
    \textbf{MORPHEUS performance by cancer type.}
    Per-gene fidelity of H\&E-inferred expression, calculated separately within each cancer type.
    Plotted as the distribution of Pearson correlation values between inferred and measured expression across the 14{,}773 genes MORPHEUS predicts.
    Blue distributions represent the TCGA validation subset; orange distributions represent CPTAC evaluation cohort (\Cref{sec:fidelity_metric}).
    Each column corresponds to one cancer type.
    The vertical axis shows Pearson correlation, and histogram width represents the number of genes on a single scale shared by every column (scale bar, 1,000 genes), so widths are comparable across types and between cohorts.
    Columns are labeled by project code, the two joined by a slash where TCGA and CPTAC code one disease differently (KIRC/CCRCC, HNSC/HNSCC, LUSC/LSCC, PAAD/PDA).
    Only cancer types with at least 20 evaluated cases are shown; cohort composition is reported in \Cref{tab:morpheus_breakdown_tcga,tab:morpheus_breakdown_cptac}.
    The TCGA validation subset was used for MORPHEUS model selection, so its fidelity estimates may be optimistic.
    CPTAC was reserved for independent evaluation.
    These distributions summarize gene-level fidelity; the correlations for the TCGA validation subset underlie the pathway enrichment analysis in \Cref{fig:morpheus}a.}
    \label{fig:ed_morpheus_fidelity}
\end{figure}

\newpage
\appendix
\renewcommand{\floatpagefraction}{0.7}

\section{Supplementary Information}
\setcounter{figure}{0}
\setcounter{table}{0}
\setcounter{algorithm}{0}
\renewcommand{\theHfigure}{supp.\arabic{figure}}
\renewcommand{\theHtable}{supp.\arabic{table}}
\renewcommand{\theHalgorithm}{supp.\arabic{algorithm}}
\captionsetup[figure]{name=Supplementary Fig.}
\captionsetup[table]{name=Supplementary Table}
\floatname{algorithm}{Supplementary Algorithm}
\renewcommand{\thefigure}{\arabic{figure}}
\renewcommand{\thetable}{\arabic{table}}
\renewcommand{\thealgorithm}{\arabic{algorithm}}
\crefalias{figure}{suppfig}
\crefalias{table}{supptab}
\crefalias{algorithm}{suppalg}

\subsection{Datasets and cohorts}\label{sec:supp_datasets}

\subsubsection{Neoadjuvant cohort characteristics}
\label{appendix:cohort_characteristics}

\begin{figure}[H]
    \centering
    \includegraphics[width=0.9\linewidth]{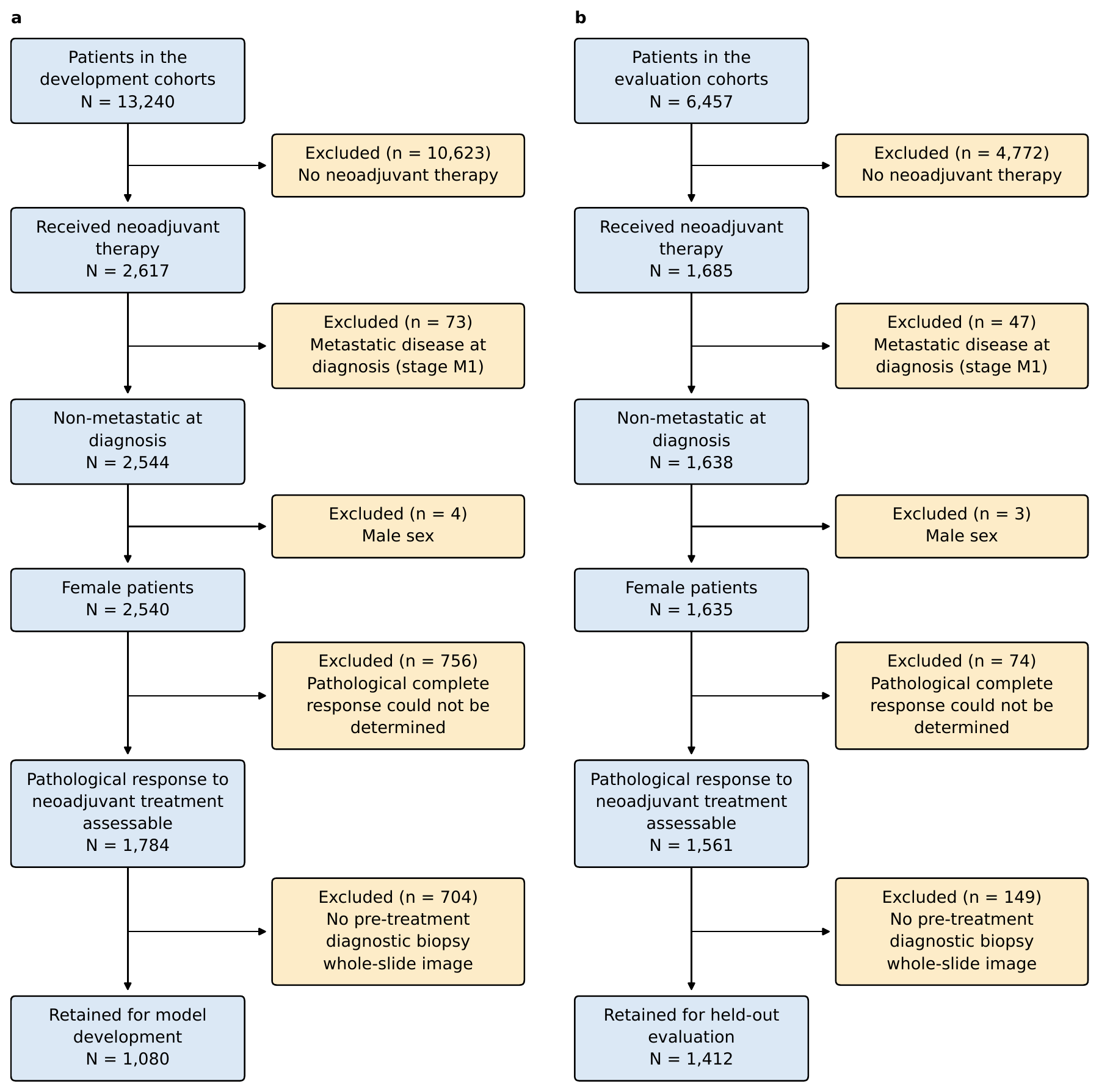}
    \caption{
    \scriptsize
    \textbf{Data filtering flow diagrams.}
    Sequential inclusion and exclusion criteria applied to assemble the development set (\textbf{a}, five cohorts) and the held-out evaluation set (\textbf{b}, nine cohorts), following the eligibility criteria of \Cref{sec:clinical_inputs}. Blue boxes indicate the number of patients retained at each step. Orange boxes indicate the number excluded and the reason for exclusion.
    Of 13,240 patients in the development cohorts, 1,080 were retained for model development: those who received NAT, were non-metastatic at diagnosis, were female, had an assessable pathological response to NAT, and had at least one pre-treatment diagnostic biopsy WSI. Of 6,457 patients in the evaluation cohorts, 1,412 were retained under the identical criteria.
    Most exclusions occurred at the first step in both sets. Several contributing institutions supplied general breast cancer cohorts, so the majority of their patients were treated in the adjuvant setting and are outside the population addressed here.}
    \label{fig:data_filtering}
\end{figure}

The neoadjuvant development and evaluation cohorts were assembled by sequentially applying the eligibility criteria in \Cref{sec:clinical_inputs}. Patient counts and exclusions at each step are shown in \Cref{fig:data_filtering}.
Demographic characteristics, receptor status, stage, histology, pCR outcomes, and treatment regimens are summarized for each split in \Cref{tab:cohorts_breakdown} and for each evaluation cohort in \Cref{tab:cohorts_breakdown_per_cohort}. Cohort sizes and geographic distribution are shown in \Cref{fig:overview}b.

\begin{table}[htbp]
    \centering
    \tiny
    \caption{
    \scriptsize
    \textbf{Characteristics of the development and evaluation cohorts.} Age is reported as median [IQR] in years; categorical variables are reported as n (\%). Percentages use the total number of patients in each column, except for treatment regimens, which use the number in the corresponding molecular subtype. Molecular subtypes (TNBC, HR$+$/HER2$-$, HER2$+$) are derived from receptor status; patients whose subtype cannot be determined are classified as unknown. Treatment regimen groups describe recorded treatment components, not specific protocols or treatment sequences. }
    \label{tab:cohorts_breakdown}

\begin{tabular}{rrr}
\hline
\textbf{Category} & \textbf{Development cohorts (N=1,080)} & \textbf{Evaluation cohorts (N=1,412)} \\
\hline
\textbf{Age median, [IQR]} & 54.00 [45.00--63.00] & 52.29 [44.00--62.00] \\
\hline
\multicolumn{3}{l}{\textbf{Race}}\\

Asian & 7 (0.65\%) & 10 (0.71\%) \\

Black or African American & 15 (1.39\%) & 119 (8.43\%) \\

White & 315 (29.17\%) & 315 (22.31\%) \\

Other/Unknown & 743 (68.80\%) & 968 (68.56\%) \\

\hline
\multicolumn{3}{l}{\textbf{ER receptor status}}\\

Positive & 531 (49.17\%) & 680 (48.16\%) \\

Negative & 513 (47.50\%) & 730 (51.70\%) \\

Unknown & 36 (3.33\%) & 2 (0.14\%) \\
\hline
\multicolumn{3}{l}{\textbf{PR receptor status}}\\

Positive & 421 (38.98\%) & 517 (36.61\%) \\

Negative & 426 (39.44\%) & 890 (63.03\%) \\

Unknown & 233 (21.57\%) & 5 (0.35\%) \\
\hline
\multicolumn{3}{l}{\textbf{HER2 receptor status}}\\

Positive & 368 (34.07\%) & 435 (30.81\%) \\

Negative & 646 (59.81\%) & 964 (68.27\%)  \\

Equivocal & 0 (0.00\%) & 9 (0.64\%) \\

Unknown & 66 (6.11\%) & 4 (0.28\%) \\
\hline
\multicolumn{3}{l}{\textbf{Molecular subtype}}\\

TNBC & 211 (19.54\%) & 518 (36.69\%) \\

HR$+$/HER2$-$ & 392 (36.30\%) & 445 (31.52\%) \\

HER2$+$ & 368 (34.07\%) & 435 (30.81\%) \\

Unknown & 109 (10.09\%) & 14 (0.99\%) \\
\hline
\multicolumn{3}{l}{\textbf{Clinical T stage}}\\

T0/Tis & 7 (0.65\%) & 5 (0.35\%) \\

T1 & 124 (11.48\%) & 189 (13.39\%) \\

T2 & 424 (39.26\%) & 732 (51.84\%) \\

T3 & 218 (20.19\%) & 220 (15.58\%) \\

T4 & 162 (15.00\%) & 73 (5.17\%) \\

Unknown/TX & 145 (13.43\%) & 193 (13.67\%) \\
\hline
\multicolumn{3}{l}{\textbf{Clinical N stage}}\\

N0 & 354 (32.78\%) & 501 (35.48\%)  \\

N1 & 272 (25.19\%) & 549 (38.88\%)  \\

N2 & 103 (9.54\%) & 84 (5.95\%)\\

N3 & 106 (9.81\%) & 61 (4.32\%)  \\

N+ (node positive, specific stage unknown) & 101 (9.35\%) & 18 (1.27\%) \\

Unknown/NX & 144 (13.33\%) & 199 (14.09\%) \\
\hline
\multicolumn{3}{l}{\textbf{Ductal histology}}\\

Yes & 876 (81.11\%) & 1,183 (83.78\%)  \\

No & 78 (7.22\%) & 81 (5.74\%) \\

Unknown & 126 (11.67\%) & 148 (10.48\%)  \\
\hline
\multicolumn{3}{l}{\textbf{Lobular histology}}\\

Yes & 73 (6.76\%) & 77 (5.45\%) \\

No & 881 (81.57\%) & 1,187 (84.07\%) \\

Unknown & 126 (11.67\%) & 148 (10.48\%)  \\
\hline
\multicolumn{3}{l}{\textbf{pCR}}\\

Yes & 367 (33.98\%) & 527 (37.32\%)  \\

No & 713 (66.02\%) & 885 (62.68\%)  \\

Unknown & 0 (0.00\%) & 0 (0.00\%) \\
\hline

\multicolumn{3}{l}{\textbf{TNBC treatment regimen} (\% of the cohort's TNBC patients)}\\

Chemotherapy + immune checkpoint inhibitor (ICI) & 12 (5.7\%) & 48 (9.3\%) \\

Platinum-containing chemotherapy, no ICI & 6 (2.8\%) & 31 (6.0\%) \\

Non-platinum chemotherapy, no ICI & 60 (28.4\%) & 90 (17.4\%) \\

Unknown & 133 (63.0\%) & 349 (67.4\%) \\

\hline
\multicolumn{3}{l}{\textbf{HER2$+$ treatment regimen} (\% of the cohort's HER2$+$ patients)}\\

Trastuzumab + pertuzumab, with anthracycline & 35 (9.5\%) & 46 (10.6\%) \\

Trastuzumab + pertuzumab, with platinum, without anthracycline & 46 (12.5\%) & 110 (25.3\%) \\

Trastuzumab + pertuzumab, without anthracycline or platinum & 23 (6.2\%) & 40 (9.2\%) \\

Trastuzumab, with anthracycline & 74 (20.1\%) & 67 (15.4\%) \\

Trastuzumab, without anthracycline & 47 (12.8\%) & 34 (7.8\%) \\

Chemotherapy without anti-HER2 & 19 (5.2\%) & 30 (6.9\%) \\

Unknown & 124 (33.7\%) & 108 (24.8\%) \\

\hline
\multicolumn{3}{l}{\textbf{HR$+$/HER2$-$ treatment regimen} (\% of the cohort's HR$+$/HER2$-$ patients)}\\

Anthracycline + taxane chemotherapy ($\pm$ endocrine) & 213 (54.3\%) & 180 (40.4\%) \\

Other chemotherapy ($\pm$ endocrine) & 63 (16.1\%) & 120 (27.0\%) \\

Neoadjuvant endocrine therapy, no chemotherapy & 60 (15.3\%) & 11 (2.5\%) \\

Unknown & 56 (14.3\%) & 134 (30.1\%) \\

\hline
\end{tabular}
\end{table}

\begin{table}[htbp]
    \centering
    \tiny
    \caption{
    \scriptsize
    \textbf{Characteristics of the evaluation cohorts, shown separately by cohort.} Age is reported as median [IQR] in years; categorical variables are reported as n (\%). Percentages use the total number of patients in each column, except for treatment regimens, which use the number in the corresponding molecular subtype. Molecular subtypes (TNBC, HR$+$/HER2$-$, HER2$+$) are derived from receptor status; patients whose subtype cannot be determined are classified as unknown. Treatment regimen groups describe recorded treatment components, not specific protocols or treatment sequences.  Age summaries and percentages are rounded to whole numbers; abbreviated row labels are given in full in \Cref{tab:cohorts_breakdown}. H, trastuzumab; P, pertuzumab; ICI, immune checkpoint inhibitor; chemo, chemotherapy; endo, endocrine therapy. }
    \label{tab:cohorts_breakdown_per_cohort}

\setlength{\tabcolsep}{2pt}
\begin{tabular}{rrrrrrrrrr}
\hline
\textbf{Category} & \textbf{Protege} & \textbf{UChicago} & \textbf{Nexus} & \textbf{Dordrecht} & \textbf{IMPRESS} & \textbf{Jagiellonian} & \shortstack{\textbf{Shaare}\\\textbf{Zedek}} & \textbf{Providence} & \textbf{\L{}\'od\'z} \\
\hline
\textbf{Patients} & 391 & 289 & 153 & 146 & 126 & 116 & 73 & 67 & 51 \\
\textbf{Age median [IQR]} & 56 [46--65] & 51 [42--61] & 50 [44--56] & 52 [42--61] & 54 [46--59] & 50 [39--57] & 48 [43--59] & 53 [43--63] & 63 [53--69] \\
\hline
\multicolumn{10}{l}{\textbf{Race}}\\
Asian & 5 (1\%) & 0 (0\%) & 0 (0\%) & 0 (0\%) & 0 (0\%) & 0 (0\%) & 0 (0\%) & 5 (7\%) & 0 (0\%) \\
Black or African American & 117 (30\%) & 0 (0\%) & 0 (0\%) & 0 (0\%) & 0 (0\%) & 0 (0\%) & 0 (0\%) & 2 (3\%) & 0 (0\%) \\
White & 263 (67\%) & 0 (0\%) & 0 (0\%) & 0 (0\%) & 0 (0\%) & 0 (0\%) & 0 (0\%) & 52 (78\%) & 0 (0\%) \\
Other/Unknown & 6 (2\%) & 289 (100\%) & 153 (100\%) & 146 (100\%) & 126 (100\%) & 116 (100\%) & 73 (100\%) & 8 (12\%) & 51 (100\%) \\
\hline
\multicolumn{10}{l}{\textbf{ER receptor status}}\\
Positive & 238 (61\%) & 145 (50\%) & 79 (52\%) & 0 (0\%) & 30 (24\%) & 71 (61\%) & 44 (60\%) & 42 (63\%) & 31 (61\%) \\
Negative & 153 (39\%) & 143 (49\%) & 74 (48\%) & 146 (100\%) & 96 (76\%) & 45 (39\%) & 29 (40\%) & 24 (36\%) & 20 (39\%) \\
Unknown & 0 (0\%) & 1 (0\%) & 0 (0\%) & 0 (0\%) & 0 (0\%) & 0 (0\%) & 0 (0\%) & 1 (1\%) & 0 (0\%) \\
\hline
\multicolumn{10}{l}{\textbf{PR receptor status}}\\
Positive & 182 (47\%) & 89 (31\%) & 79 (52\%) & 0 (0\%) & 19 (15\%) & 55 (47\%) & 33 (45\%) & 34 (51\%) & 26 (51\%) \\
Negative & 209 (53\%) & 199 (69\%) & 74 (48\%) & 146 (100\%) & 107 (85\%) & 61 (53\%) & 39 (53\%) & 30 (45\%) & 25 (49\%) \\
Unknown & 0 (0\%) & 1 (0\%) & 0 (0\%) & 0 (0\%) & 0 (0\%) & 0 (0\%) & 1 (1\%) & 3 (4\%) & 0 (0\%) \\
\hline
\multicolumn{10}{l}{\textbf{HER2 receptor status}}\\
Positive & 106 (27\%) & 94 (33\%) & 75 (49\%) & 0 (0\%) & 62 (49\%) & 31 (27\%) & 22 (30\%) & 23 (34\%) & 22 (43\%) \\
Negative & 285 (73\%) & 187 (65\%) & 78 (51\%) & 146 (100\%) & 64 (51\%) & 85 (73\%) & 50 (68\%) & 42 (63\%) & 27 (53\%) \\
Equivocal & 0 (0\%) & 7 (2\%) & 0 (0\%) & 0 (0\%) & 0 (0\%) & 0 (0\%) & 1 (1\%) & 0 (0\%) & 1 (2\%) \\
Unknown & 0 (0\%) & 1 (0\%) & 0 (0\%) & 0 (0\%) & 0 (0\%) & 0 (0\%) & 0 (0\%) & 2 (3\%) & 1 (2\%) \\
\hline
\multicolumn{10}{l}{\textbf{Molecular subtype}}\\
TNBC & 108 (28\%) & 97 (34\%) & 36 (24\%) & 146 (100\%) & 64 (51\%) & 30 (26\%) & 16 (22\%) & 13 (19\%) & 8 (16\%) \\
HR$+$/HER2$-$ & 177 (45\%) & 90 (31\%) & 42 (27\%) & 0 (0\%) & 0 (0\%) & 55 (47\%) & 34 (47\%) & 28 (42\%) & 19 (37\%) \\
HER2$+$ & 106 (27\%) & 94 (33\%) & 75 (49\%) & 0 (0\%) & 62 (49\%) & 31 (27\%) & 22 (30\%) & 23 (34\%) & 22 (43\%) \\
Unknown & 0 (0\%) & 8 (3\%) & 0 (0\%) & 0 (0\%) & 0 (0\%) & 0 (0\%) & 1 (1\%) & 3 (4\%) & 2 (4\%) \\
\hline
\multicolumn{10}{l}{\textbf{Clinical T stage}}\\
T0/Tis & 2 (1\%) & 1 (0\%) & 0 (0\%) & 0 (0\%) & 0 (0\%) & 0 (0\%) & 1 (1\%) & 0 (0\%) & 1 (2\%) \\
T1 & 53 (14\%) & 54 (19\%) & 11 (7\%) & 31 (21\%) & 0 (0\%) & 1 (1\%) & 9 (12\%) & 21 (31\%) & 9 (18\%) \\
T2 & 204 (52\%) & 155 (54\%) & 92 (60\%) & 96 (66\%) & 0 (0\%) & 72 (62\%) & 47 (64\%) & 41 (61\%) & 25 (49\%) \\
T3 & 67 (17\%) & 56 (19\%) & 38 (25\%) & 9 (6\%) & 0 (0\%) & 28 (24\%) & 12 (16\%) & 4 (6\%) & 6 (12\%) \\
T4 & 17 (4\%) & 15 (5\%) & 11 (7\%) & 9 (6\%) & 0 (0\%) & 9 (8\%) & 4 (5\%) & 1 (1\%) & 7 (14\%) \\
Unknown/TX & 48 (12\%) & 8 (3\%) & 1 (1\%) & 1 (1\%) & 126 (100\%) & 6 (5\%) & 0 (0\%) & 0 (0\%) & 3 (6\%) \\
\hline
\multicolumn{10}{l}{\textbf{Clinical N stage}}\\
N0 & 167 (43\%) & 112 (39\%) & 40 (26\%) & 79 (54\%) & 0 (0\%) & 34 (29\%) & 18 (25\%) & 30 (45\%) & 21 (41\%) \\
N1 & 146 (37\%) & 133 (46\%) & 100 (65\%) & 50 (34\%) & 0 (0\%) & 50 (43\%) & 20 (27\%) & 30 (45\%) & 20 (39\%) \\
N2 & 7 (2\%) & 28 (10\%) & 11 (7\%) & 9 (6\%) & 0 (0\%) & 12 (10\%) & 8 (11\%) & 6 (9\%) & 3 (6\%) \\
N3 & 22 (6\%) & 7 (2\%) & 1 (1\%) & 7 (5\%) & 0 (0\%) & 12 (10\%) & 10 (14\%) & 1 (1\%) & 1 (2\%) \\
N+ (stage unspecified) & 0 (0\%) & 0 (0\%) & 1 (1\%) & 0 (0\%) & 0 (0\%) & 0 (0\%) & 17 (23\%) & 0 (0\%) & 0 (0\%) \\
Unknown/NX & 49 (13\%) & 9 (3\%) & 0 (0\%) & 1 (1\%) & 126 (100\%) & 8 (7\%) & 0 (0\%) & 0 (0\%) & 6 (12\%) \\
\hline
\multicolumn{10}{l}{\textbf{Ductal histology}}\\
Yes & 355 (91\%) & 276 (96\%) & 145 (95\%) & 132 (90\%) & 61 (48\%) & 107 (92\%) & 63 (86\%) & 0 (0\%) & 44 (86\%) \\
No & 33 (8\%) & 5 (2\%) & 8 (5\%) & 8 (5\%) & 1 (1\%) & 9 (8\%) & 10 (14\%) & 0 (0\%) & 7 (14\%) \\
Unknown & 3 (1\%) & 8 (3\%) & 0 (0\%) & 6 (4\%) & 64 (51\%) & 0 (0\%) & 0 (0\%) & 67 (100\%) & 0 (0\%) \\
\hline
\multicolumn{10}{l}{\textbf{Lobular histology}}\\
Yes & 27 (7\%) & 29 (10\%) & 8 (5\%) & 1 (1\%) & 1 (1\%) & 1 (1\%) & 6 (8\%) & 0 (0\%) & 4 (8\%) \\
No & 361 (92\%) & 252 (87\%) & 145 (95\%) & 139 (95\%) & 61 (48\%) & 115 (99\%) & 67 (92\%) & 0 (0\%) & 47 (92\%) \\
Unknown & 3 (1\%) & 8 (3\%) & 0 (0\%) & 6 (4\%) & 64 (51\%) & 0 (0\%) & 0 (0\%) & 67 (100\%) & 0 (0\%) \\
\hline
\multicolumn{10}{l}{\textbf{pCR}}\\
Yes & 120 (31\%) & 105 (36\%) & 69 (45\%) & 56 (38\%) & 65 (52\%) & 39 (34\%) & 29 (40\%) & 31 (46\%) & 13 (25\%) \\
No & 271 (69\%) & 184 (64\%) & 84 (55\%) & 90 (62\%) & 61 (48\%) & 77 (66\%) & 44 (60\%) & 36 (54\%) & 38 (75\%) \\
Unknown & 0 (0\%) & 0 (0\%) & 0 (0\%) & 0 (0\%) & 0 (0\%) & 0 (0\%) & 0 (0\%) & 0 (0\%) & 0 (0\%) \\
\hline
\multicolumn{10}{l}{\textbf{TNBC treatment regimen} (\% of the cohort's TNBC patients)}\\
Chemotherapy + ICI & 27 (25\%) & 1 (1\%) & 0 (0\%) & 0 (0\%) & 0 (0\%) & 5 (17\%) & 15 (94\%) & 0 (0\%) & 0 (0\%) \\
Platinum-containing, no ICI & 1 (1\%) & 4 (4\%) & 0 (0\%) & 0 (0\%) & 0 (0\%) & 25 (83\%) & 1 (6\%) & 0 (0\%) & 0 (0\%) \\
Non-platinum chemo, no ICI & 1 (1\%) & 86 (89\%) & 0 (0\%) & 0 (0\%) & 0 (0\%) & 0 (0\%) & 0 (0\%) & 0 (0\%) & 3 (38\%) \\
Unknown & 79 (73\%) & 6 (6\%) & 36 (100\%) & 146 (100\%) & 64 (100\%) & 0 (0\%) & 0 (0\%) & 13 (100\%) & 5 (62\%) \\
\hline
\multicolumn{10}{l}{\textbf{HER2$+$ treatment regimen} (\% of the cohort's HER2$+$ patients)}\\
H+P, anthracycline & 0 (0\%) & 16 (17\%) & 0 (0\%) & --- & 0 (0\%) & 26 (84\%) & 0 (0\%) & 0 (0\%) & 4 (18\%) \\
H+P, platinum, no anthracycline & 62 (58\%) & 20 (21\%) & 0 (0\%) & --- & 0 (0\%) & 1 (3\%) & 17 (77\%) & 0 (0\%) & 10 (45\%) \\
H+P, no anthracycline/platinum & 23 (22\%) & 12 (13\%) & 0 (0\%) & --- & 0 (0\%) & 0 (0\%) & 4 (18\%) & 0 (0\%) & 1 (5\%) \\
H, anthracycline & 0 (0\%) & 9 (10\%) & 52 (69\%) & --- & 0 (0\%) & 4 (13\%) & 0 (0\%) & 0 (0\%) & 2 (9\%) \\
H, no anthracycline & 3 (3\%) & 27 (29\%) & 0 (0\%) & --- & 0 (0\%) & 0 (0\%) & 0 (0\%) & 0 (0\%) & 4 (18\%) \\
Chemotherapy without anti-HER2 & 0 (0\%) & 6 (6\%) & 23 (31\%) & --- & 0 (0\%) & 0 (0\%) & 1 (5\%) & 0 (0\%) & 0 (0\%) \\
Unknown & 18 (17\%) & 4 (4\%) & 0 (0\%) & --- & 62 (100\%) & 0 (0\%) & 0 (0\%) & 23 (100\%) & 1 (5\%) \\
\hline
\multicolumn{10}{l}{\textbf{HR$+$/HER2$-$ treatment regimen} (\% of the cohort's HR$+$/HER2$-$ patients)}\\
Anthracycline + taxane ($\pm$ endo.) & 5 (3\%) & 70 (78\%) & 42 (100\%) & --- & --- & 23 (42\%) & 28 (82\%) & 0 (0\%) & 12 (63\%) \\
Other chemotherapy ($\pm$ endo.) & 65 (37\%) & 17 (19\%) & 0 (0\%) & --- & --- & 32 (58\%) & 5 (15\%) & 0 (0\%) & 1 (5\%) \\
Endocrine therapy, no chemo & 4 (2\%) & 1 (1\%) & 0 (0\%) & --- & --- & 0 (0\%) & 1 (3\%) & 0 (0\%) & 5 (26\%) \\
Unknown & 103 (58\%) & 2 (2\%) & 0 (0\%) & --- & --- & 0 (0\%) & 0 (0\%) & 28 (100\%) & 1 (5\%) \\
\hline
\end{tabular}
\end{table}

\subsubsection{TCGA and CPTAC cohorts}
\label{appendix:morpheus_cohorts}

MORPHEUS training and validation cohorts are summarized by TCGA project in \Cref{tab:morpheus_breakdown_tcga}; independent evaluation cohorts are summarized by CPTAC cancer type in \Cref{tab:morpheus_breakdown_cptac}. Filtering and slide--RNA pairing are described in \Cref{subsec:rna_expression_prediction_dataset}. Expression fidelity was evaluated per patient in TCGA and per primary-tumor RNA-seq sample in CPTAC (\Cref{sec:fidelity_metric}).

TCGA training cohort sizes ranged from 28 patients (CHOL) to 835 (BRCA), with a median of 172. The 13 projects with at least 250 patients contributed 5,041 of the 6,994 training patients (72\%); BRCA alone contributed 12\%.
Validation cohort sizes ranged from 6 to 212 patients (median 44). Nine projects had fewer than 20 validation patients (ACC, CHOL, DLBC, KICH, MESO, OV, SKCM, UCS, and UVM) and were excluded from the per-type fidelity summaries and pathway analyses (\Cref{fig:ed_morpheus_fidelity}; \Cref{fig:morpheus}a).

CPTAC cohort sizes ranged from 61 patients (OV) to 236 (UCEC). CCRCC contributed 314 RNA-seq samples from 222 patients and accounted for 40 of the 41 patients with multiple primary-tumor samples.

No patient in the fourteen NAT cohorts appears in TCGA or CPTAC.

\begin{table}[htbp]
    \centering
    \footnotesize
    \caption{
    \scriptsize
    \textbf{MORPHEUS training-validation data composition by TCGA project.} Patients and distinct WSIs in the MORPHEUS training and TCGA validation subsets derived from the 80:20 patient-level split of \Cref{subsec:rna_expression_prediction_dataset}.  $^{\dagger}$Fewer than 20 validation patients, not visualized in \Cref{fig:ed_morpheus_fidelity} or \Cref{fig:morpheus}a.}
    \label{tab:morpheus_breakdown_tcga}
    \begin{tabular}{lrrrr}
        \hline
        & \multicolumn{2}{c}{Training} & \multicolumn{2}{c}{Validation} \\
        Project & Patients & WSIs & Patients & WSIs \\
        \hline
        ACC$^{\dagger}$ & 46 & 183 & 9 & 40 \\
        BLCA & 293 & 337 & 84 & 108 \\
        BRCA & 835 & 886 & 212 & 227 \\
        CESC & 213 & 220 & 54 & 57 \\
        CHOL$^{\dagger}$ & 28 & 28 & 7 & 7 \\
        COAD & 335 & 338 & 85 & 89 \\
        DLBC$^{\dagger}$ & 38 & 38 & 6 & 6 \\
        ESCA & 123 & 125 & 32 & 32 \\
        GBM & 102 & 267 & 25 & 57 \\
        HNSC & 372 & 392 & 70 & 72 \\
        KICH$^{\dagger}$ & 58 & 58 & 8 & 8 \\
        KIRC & 409 & 414 & 97 & 98 \\
        KIRP & 223 & 246 & 49 & 50 \\
        LGG & 395 & 656 & 96 & 187 \\
        LIHC & 291 & 297 & 67 & 69 \\
        LUAD & 362 & 421 & 89 & 93 \\
        LUSC & 366 & 390 & 100 & 110 \\
        MESO$^{\dagger}$ & 64 & 74 & 11 & 13 \\
        OV$^{\dagger}$ & 67 & 68 & 16 & 16 \\
        PAAD & 145 & 167 & 31 & 35 \\
        PCPG & 141 & 160 & 35 & 36 \\
        PRAD & 305 & 339 & 92 & 104 \\
        READ & 120 & 120 & 33 & 33 \\
        SARC & 199 & 486 & 53 & 110 \\
        SKCM$^{\dagger}$ & 79 & 90 & 18 & 19 \\
        STAD & 276 & 296 & 77 & 81 \\
        TGCT & 110 & 202 & 39 & 50 \\
        THCA & 399 & 410 & 105 & 107 \\
        THYM & 95 & 144 & 22 & 32 \\
        UCEC & 403 & 448 & 95 & 107 \\
        UCS$^{\dagger}$ & 37 & 66 & 16 & 21 \\
        UVM$^{\dagger}$ & 65 & 65 & 15 & 15 \\
        \hline
        \textbf{Total} & 6,994 & 8,431 & 1,748 & 2,089 \\
        \hline
    \end{tabular}
\end{table}

\begin{table}[htbp]
    \centering
    \footnotesize
    \caption{
    \scriptsize
    \textbf{CPTAC data composition by cancer type.} Patients, primary-tumor RNA-seq samples, and WSIs, after restriction to primary tumor. Fidelity is evaluated per RNA-seq sample rather than per patient, because each sample carries its own slides (\Cref{sec:fidelity_metric}); 41 patients contribute more than one sample and therefore more than one case. Every cohort exceeds the 20-case threshold, so all 10 appear individually in \Cref{fig:ed_morpheus_fidelity}.}
    \label{tab:morpheus_breakdown_cptac}
    \begin{tabular}{lrrr}
        \hline
        Cancer type & Patients & RNA-seq samples & WSIs \\
        \hline
        BRCA & 106 & 106 & 106 \\
        CCRCC & 222 & 314 & 340 \\
        COAD & 104 & 104 & 104 \\
        GBM & 157 & 157 & 210 \\
        HNSCC & 110 & 110 & 142 \\
        LSCC & 108 & 109 & 132 \\
        LUAD & 224 & 224 & 277 \\
        OV & 61 & 61 & 61 \\
        PDA & 145 & 145 & 212 \\
        UCEC & 236 & 236 & 261 \\
        \hline
        \textbf{Total} & 1,473 & 1,566 & 1,845 \\
        \hline
    \end{tabular}
\end{table}

\subsubsection{Expression preprocessing}\label{sec:expression_preprocessing}

Genes were retained if they had more than 0.5 transcripts per million (TPM) in at least 35\% of TCGA training samples, leaving 14,773 out of 19,962 protein-coding genes.
A variance-stabilizing transformation (VST; \cite{love2014moderated}) was fitted separately to the unstranded count matrices of TCGA training subset, TCGA validation subset, and CPTAC data.
VST reduces the dependence of RNA-seq count variance on mean expression.
Separate fits accommodate differences in sequencing protocols and preprocessing without making one set's transformation depend on another set's counts.

MORPHEUS predictions are on the transformed scale of the TCGA training subset. Because the targets use separately fitted transformations across datasets, fidelity was assessed by within-cohort correlation rather than absolute agreement (\Cref{sec:fidelity_metric}).
After VST, each gene was standardized using its TCGA training-subset mean and standard deviation.
The standardized values were used as MORPHEUS training targets.

\subsection{Model development}\label{sec:supp_model_development}\label{sec:architectures}

\begin{figure}[H]
    \centering
    \includegraphics[width=\linewidth]{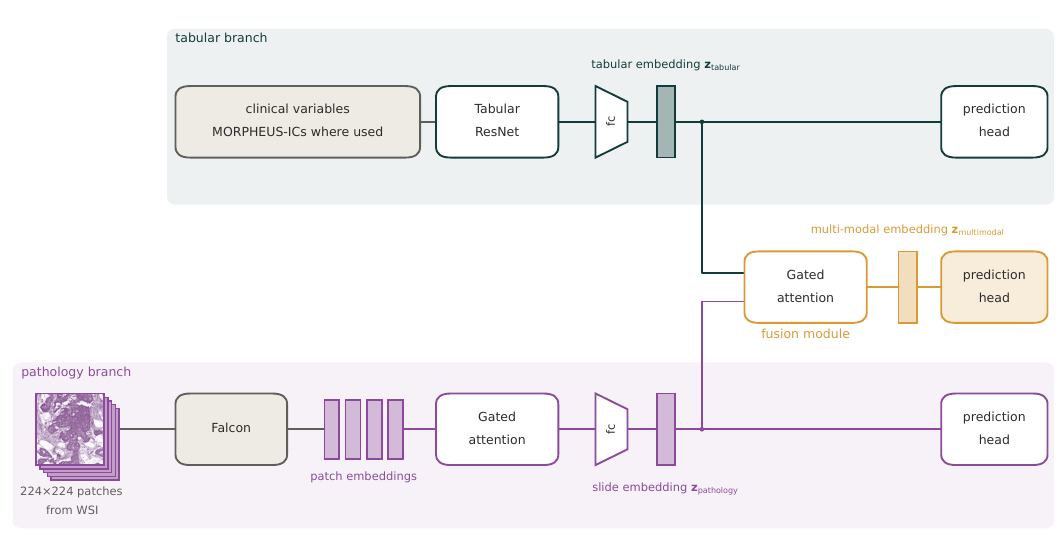}
    \caption{
    \scriptsize
    \textbf{Components of the neural-network-based pCR prediction models and MORPHEUS.}
    Inputs flow from left to right through the color-coded branches to the prediction heads. The pathology branch (lower, purple) begins with the whole-slide image, tiled into $224\times224$ patches and encoded by the frozen Falcon encoder. The pathology branch then aggregates the resulting patch-level embeddings by gated attention (\Cref{eq:gated_attention}). The attention-weighted average passes through dropout and linear projections (\Cref{sec:gated_attention_detail}) to width $d_{\mathrm{int}}$, shared between modalities, giving the slide embedding $\mathbf{z}_{\mathrm{pathology}}$. The tabular branch (upper lane, green) passes its inputs through a tabular ResNet and a fully-connected layer to the tabular embedding $\mathbf{z}_{\mathrm{tabular}}$ of the same width $d_{\mathrm{int}}$. Those inputs are the baseline clinical variables, the five MORPHEUS-ICs of the H\&E-inferred expression, or both, depending on the model. The fusion module (between the two lanes, orange) belongs to neither branch: it applies a second gated-attention block over the two embeddings and outputs the multi-modal embedding $\mathbf{z}_{\mathrm{multimodal}}$.
    A classification prediction head contains one linear layer per binary subtask.
    The multi-modal direct pCR prediction model uses both branches together with the fusion module, and has prediction heads on each branch and on the fused representation (\Cref{sec:multitask_detail}).
    MORPHEUS follows the pathology branch shown here, with a linear expression-regression head in place of the classification head.
    NEO uses MORPHEUS-derived features together with clinical variables, calibrates each per-fold tabular ResNet and EBM, and averages the calibrated outputs (\Cref{sec:response_model,sec:calibration}).
    Hyperparameter ranges for each component, and where dropout sits within each, are given in \Cref{sec:hyperparameters} and \Cref{tab:hyperparameters}.}
    \label{fig:model_components}
\end{figure}

\subsubsection{Clinical-input encoding}\label{sec:clinical_encoding}
Clinical inputs were processed identically during training and evaluation.
For the tabular ResNet, categorical variables were one-hot encoded, with an additional category for missing values of each variable.
Missing age values were imputed using the development cohort mean.
The EBM used categorical variables directly and handled missing clinical values through its built-in missing-value bins, without age imputation.

\subsubsection{Training and inference algorithms}\label{sec:algorithms}
\Cref{alg:mscv} is the multiple-source cross-validation protocol (\Cref{subsec:mscv-main}), applied to one model class at a time: it fits models on all but one cross-validation cohort at a time, uses the held-out cohorts for early stopping and configuration selection, and fits each fold's calibrator on that fold's held-out cohort.
\Cref{alg:atxneo} assembles NEO itself, invoking \Cref{alg:mscv} once per Falcon variant and model class, and averaging the calibrated predictions into the reported probability of pCR (\Cref{sec:response_model,sec:backbone_combination}).
The direct pCR prediction models use the same MSCV protocol (\Cref{sec:direct_path_model}).

\begin{algorithm}[H]
\caption{
    \scriptsize
\textbf{\textsc{MSCV}}$(\mathcal{D}^T, \mathcal{D}^V, q, \mathcal{H}_q, R)$: multiple-source cross-validated training, selection, and calibration of one model class}
\label{alg:mscv}
\begin{algorithmic}[1]
\REQUIRE Train-only cohorts $\mathcal{D}^T = \{D_1, \ldots, D_J\}$ and cross-validation cohorts $\mathcal{D}^V = \{D_{J+1}, \ldots, D_{J+K}\}$, where $K = |\mathcal{D}^V|$,
         with input features and pCR labels for every patient;
         model class $q$ with candidate hyperparameter configurations $\mathcal{H}_q$;
         number of configurations to retain $R \le |\mathcal{H}_q|$
\ENSURE  A fitted model $f_{q,k}$ and its calibrator $\gamma_{q,k}$ for each fold $k = 1, \ldots, K$
\FOR{$k = 1, \ldots, K$}
  \STATE $\mathcal{D}^{\mathrm{val}}_{k} \leftarrow D_{J+k}$, \quad $\mathcal{D}^{\mathrm{tr}}_{k} \leftarrow (\mathcal{D}^T \cup \mathcal{D}^V) \setminus D_{J+k}$ \COMMENT{one held-out cohort per fold}
  \STATE Train every configuration $\eta \in \mathcal{H}_q$ on $\mathcal{D}^{\mathrm{tr}}_{k}$ and score it on $\mathcal{D}^{\mathrm{val}}_{k}$ \COMMENT{$|\mathcal{H}_q|K$ models in total}
\ENDFOR
\STATE $\mathcal{H}_q^\star \leftarrow$ the $R$ configurations with the highest AUROC averaged over $\mathcal{D}^{\mathrm{val}}_{1}, \ldots, \mathcal{D}^{\mathrm{val}}_{K}$
\FOR{$k = 1, \ldots, K$}
  \STATE $f_{q,k} \leftarrow$ mean prediction of the configurations in $\mathcal{H}_q^\star$ trained on $\mathcal{D}^{\mathrm{tr}}_{k}$
  \STATE $\gamma_{q,k} \leftarrow$ beta calibrator fitted on the predictions of $f_{q,k}$ for $\mathcal{D}^{\mathrm{val}}_{k}$ \COMMENT{\Cref{sec:calibration}}
\ENDFOR
\RETURN $\{(f_{q,k}, \gamma_{q,k})\}_{k=1}^{K}$
\end{algorithmic}
\end{algorithm}

\begin{algorithm}[H]
\caption{
    \scriptsize
\textbf{NEO}: feature construction, cross-validated training, and calibrated ensembling}
\label{alg:atxneo}
\begin{algorithmic}[1]
\REQUIRE Patient $p$ with biopsy slides $\mathcal{S}_p$ and baseline clinical variables $c_p$ (\Cref{sec:clinical_inputs});
         Falcon variants $\mathcal{B} = \{\text{g}, \text{L}\}$;
         model classes $\mathcal{Q} = \{\text{EBM}, \text{ResNet}\}$ with retained-configuration counts $R_{\text{EBM}} = 1$ (fixed settings; \Cref{sec:ebm_mscv_integration}) and $R_{\text{ResNet}} = 10$;
         development cohorts $\mathcal{D}^T$, $\mathcal{D}^V$, where $K = |\mathcal{D}^V|$, and candidate configurations $\mathcal{H}_q$ for each model class, as in \Cref{alg:mscv};
         fitted $\textsc{Morpheus}_b$ and $\mathrm{ICA}_b$ for each Falcon variant $b$
\ENSURE  Calibrated probability $\hat{p}(\text{pCR} \mid \text{NAT})$
\STATE \textbf{Feature construction} \COMMENT{apply to each development patient and each patient to be scored}
\FORALL{branches $b \in \mathcal{B}$}
  \STATE $z_{s,b} \leftarrow \textsc{Morpheus}_b(s)$ for every slide $s \in \mathcal{S}_p$ \COMMENT{14,773 RNA values per slide}
  \STATE $\bar{z}_{p,b} \leftarrow \frac{1}{|\mathcal{S}_p|}\sum_{s \in \mathcal{S}_p} z_{s,b}$ \COMMENT{patient level}
  \STATE $g_{p,b} \leftarrow \mathrm{ICA}_b\!\left(\bar{z}_{p,b}[\,\mathcal{G}\,]\right)$ \COMMENT{$\mathcal{G}$: 2,427 MSigDB-family genes; ICA fit on TCGA validation subset}
  \STATE $x_{p,b} \leftarrow [\,c_p \,;\, g_{p,b}\,]$ \COMMENT{model input; categorical variables encoded per model class}
  \STATE Form $\mathcal{D}^T_b$ and $\mathcal{D}^V_b$ from the development cohorts using features $x_{p,b}$ and the corresponding outcome labels
\ENDFOR
\STATE \textbf{Training and selection} \COMMENT{development cohorts only; no evaluation cohort is used}
\FORALL{$b \in \mathcal{B}$, $q \in \mathcal{Q}$}
  \STATE $\{(f_{b,q,k},\, \gamma_{b,q,k})\}_{k=1}^{K} \leftarrow \textsc{MSCV}(\mathcal{D}^T_b, \mathcal{D}^V_b, q, \mathcal{H}_q, R_q)$ \COMMENT{\Cref{alg:mscv}}
\ENDFOR
\STATE \textbf{Inference}
\STATE $\hat{p} \leftarrow \dfrac{1}{|\mathcal{B}||\mathcal{Q}|K} \displaystyle\sum_{b \in \mathcal{B}} \sum_{q \in \mathcal{Q}} \sum_{k=1}^{K} \gamma_{b,q,k}\!\left(f_{b,q,k}(x_{p,b})\right)$
\RETURN $\hat{p}$
\end{algorithmic}
\end{algorithm}

\subsubsection{Slide-level aggregation by gated attention}\label{sec:gated_attention_detail}
For a slide containing $M$ patches, let $h_1,\dots,h_M \in \mathbb{R}^{d}$ denote the foundation-model embeddings, with $d=1536$ for Falcon-g and $d=1024$ for Falcon-L. Gated attention assigns each patch a normalized weight
\begin{equation}
a_j = \frac{\exp\!\big(w^{\top}\left[\tanh(V h_j) \odot \sigma(U h_j)\right]\big)}
           {\sum_{j'=1}^{M} \exp\!\big(w^{\top}\left[\tanh(V h_{j'}) \odot \sigma(U h_{j'})\right]\big)},
\label{eq:gated_attention}
\end{equation}
with $V, U \in \mathbb{R}^{d_a \times d}$, $w \in \mathbb{R}^{d_a}$, $\odot$ the element-wise product, and $\sigma$ the logistic function.
The attention width was $d_a=1$ for patch aggregation.
The weighted mean, $Z = \sum_j a_j h_j \in \mathbb{R}^{d}$, passes through dropout, a $d \times d$ linear projection, and a second linear projection to the intermediate feature dimension $d_{\mathrm{int}}$, which is tuned for each model (\Cref{tab:hyperparameters}).
In MORPHEUS the resulting slide-level embedding is passed through a single linear layer to jointly predict all 14,773 VST-normalized gene expression values (\Cref{sec:gene_expression_model}). In the uni-modal direct pCR prediction model the same aggregator feeds the multitask classification head instead (\Cref{sec:direct_path_model,sec:multitask_detail}).

\subsubsection{Fusion of the clinical and pathology branches}\label{sec:fusion_detail}
The multi-modal direct pCR prediction model combines a pathology embedding, $\mathbf{z}_{\mathrm{pathology}} \in \mathbb{R}^{d_{\mathrm{int}}}$, with a clinical embedding, $\mathbf{z}_{\mathrm{tabular}} \in \mathbb{R}^{d_{\mathrm{int}}}$, produced by a tabular ResNet (\Cref{fig:model_components,sec:direct_path_model}).
The embeddings are stacked into a two-element sequence and weighted by the gated-attention function in \Cref{eq:gated_attention}, with $M=2$ and $d=d_{\mathrm{int}}$.
Their weighted sum, $a_{\mathrm{tabular}}\mathbf{z}_{\mathrm{tabular}} + a_{\mathrm{pathology}}\mathbf{z}_{\mathrm{pathology}}$, passes through dropout and a $d_{\mathrm{int}} \times d_{\mathrm{int}}$ linear projection to produce the multi-modal embedding $\mathbf{z}_{\mathrm{multimodal}}$.
The fusion attention width is $d_a=\lfloor\gamma_{\mathrm{att}}d_{\mathrm{int}}\rfloor$; both the multiplier $\gamma_{\mathrm{att}}$ and the dropout rate were tuned (\Cref{tab:hyperparameters}).

Prediction heads are attached to the clinical, pathology, and fused representations. Each head contains one linear layer per binary subtask (\Cref{sec:multitask_detail}), and their losses are combined using the branch weights in \Cref{eq:loss}.
The reported pCR probability is the output of the fused head.

\subsubsection{Optimization and model selection}\label{sec:optimization_detail}
MORPHEUS and the neural-network-based pCR prediction models (\Cref{fig:model_components}) were trained using Adam~\citep{kingma2014adam}, with a batch size of 1,100. MORPHEUS minimized the regularized expression-regression objective in \Cref{sec:morpheus_objective}. Each slide was a separate training example paired with its patient's RNA-seq profile. The TCGA training subset was used for fitting and the validation subset for early stopping and model selection; CPTAC was reserved for independent evaluation.

The neural-network-based pCR prediction models used the multitask objective in \Cref{sec:multitask_detail} under the MSCV protocol (\Cref{subsec:mscv-main}). From each model's hyperparameter search, we retained the ten configurations with the highest mean validation pCR AUROC across the $K$ folds. Within each fold, their predictions were averaged and then calibrated on that fold's held-out cohort (\Cref{sec:calibration}).

Search budgets, stopping criteria, and parameter ranges for MORPHEUS and the neural-network-based pCR prediction models are specified in \Cref{sec:hyperparameters}.

EBMs~\citep{nori2019interpretml,caruana2015intelligible} were fitted without hyperparameter search, using library defaults except for the bagging and weighting settings in \Cref{sec:ebm_mscv_integration}.

\subsubsection{Auxiliary target definitions}\label{sec:auxiliary_targets}
Post-treatment stage and downstaging provided auxiliary supervision for the neural-network-based pCR prediction models. These targets were used only during training; their predictions were not evaluated as clinical endpoints.

\paragraph{Post-treatment pathological staging.}
Beyond binary pCR status (ypT0/Tis and ypN0), we modeled post-treatment pathological tumor and nodal stage (ypT and ypN) as auxiliary targets. These variables reflect the anatomical extent of residual invasive disease in the breast and axillary lymph nodes following NAT.

Pathological T stage was encoded as ypT0/Tis, ypT1, ypT2, ypT3, or ypT4, and pathological N stage as ypN0, ypN1, ypN2 or ypN3. Alphabetic subcategories were collapsed into their parent numeric categories.

\paragraph{TN downstaging.}

Downstaging was defined by comparing baseline clinical stage (cT, cN) with post-treatment pathological stage (ypT, ypN).
T-stage and N-stage downstaging were separate binary targets: ypT $<$ cT and ypN $<$ cN, respectively.
For each target, stable stage or upstaging was classified as no downstaging.
Unavailable or indeterminate labels were masked as described in \Cref{sec:multitask_detail}.

These variables capture reductions in stage, including partial responses that do not meet pCR criteria, and complement the residual-stage targets.

\subsubsection{Multitask learning formulation and training objective}\label{sec:multitask_detail}

\begin{figure}
    \centering
    \includegraphics[width=\linewidth]{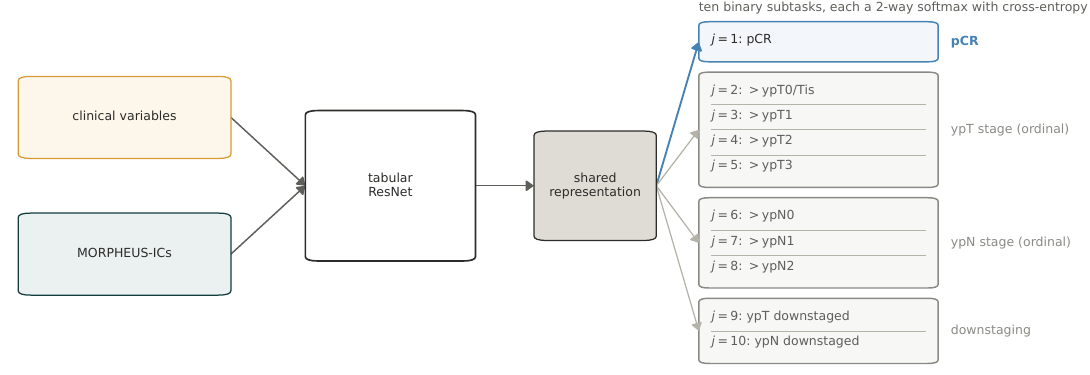}
    \caption{
    \scriptsize
    \textbf{Multitask learning.} During training, the tabular ResNet was optimized with auxiliary tasks to provide additional supervision. In addition to pCR prediction, the model jointly predicts pathological staging (as an ordinal classification task) and T/N downstaging. The ordinal outcomes are decomposed into cumulative binary subtasks; together with pCR and the two downstaging targets, these yield ten binary subtasks, each trained using two-class softmax outputs and cross-entropy loss.}
    \label{fig:multitask}
\end{figure}

The neural-network-based pCR prediction models were trained by multitask classification (\Cref{fig:multitask,sec:direct_path_model}).

The five jointly predicted outcomes were pCR, pathological T stage, pathological N stage, T-stage downstaging, and N-stage downstaging.

Ordinal pathological staging tasks were implemented using ordinal regression~\cite{frank2001simple} via multiple binary classification subtasks.
Pathological T stage (ypT0/Tis, ypT1, ypT2, ypT3, ypT4) was represented using four cumulative binary tasks ($>$ypT0/Tis, $>$ypT1, $>$ypT2, $>$ypT3).
Pathological N stage (ypN0, ypN1, ypN2, ypN3) was represented using three cumulative binary tasks ($>$ypN0, $>$ypN1, $>$ypN2).
Together with pCR and T/N downstaging, this yields 10 binary classification subtasks indexed by $j \in \{1,\ldots,10\}$:
\begin{itemize}
\item $j=1$: pCR;
\item $j=2,\dots,5$: ordinal pathological T staging thresholds ($>$ypT0/Tis, $>$ypT1, $>$ypT2, $>$ypT3);
\item $j=6,\dots,8$: ordinal pathological N staging thresholds ($>$ypN0, $>$ypN1, $>$ypN2);
\item $j=9,10$: ypT and ypN downstaging.
\end{itemize}

\paragraph{Per-task cross-entropy loss.}

Prediction heads are indexed by $m$.
NEO's tabular ResNet has one head.
The multi-modal direct pCR prediction model has three, attached to the tabular, pathology, and fused representations ($m=\mathrm{tabular},\mathrm{pathology},\mathrm{multimodal}$). Each head predicts the same ten binary subtasks using shared outcome labels.
For sample $i$ and task $j$, let $y_{i,j,c}$ denote the one-hot label for class $c$, and $\hat{p}^{(m)}_{i,j,c}$ the corresponding softmax probability from prediction head $m$.

A binary mask $M_{i,j} \in \{0,1\}$ indicates whether the label for sample $i$ and task $j$ is available: $M_{i,j}=1$ if the label is observed and contributes to the loss, $M_{i,j}=0$ otherwise.
This mask handles missing labels and ordinal thresholds that cannot be determined from partially specified staging categories (e.g., for a patient labeled ypN$+$ without subclassification, only the $>$ypN0 target is assigned; the $>$ypN1 and $>$ypN2 thresholds are masked).

The task-level cross-entropy loss for task $j$, evaluated on the prediction head of branch $m$, is
\begin{align}
\mathcal{L}^{(m)}_{\mathrm{CE},j}
=
\frac{
\sum_{i=1}^{N} M_{i,j}
\left[
-\sum_{c=1}^{2}
y_{i,j,c}
\log \hat{p}^{(m)}_{i,j,c}
\right]
}{
\sum_{i=1}^{N} M_{i,j}
},
\end{align}
where $N$ is the batch size. Normalization by $\sum_i M_{i,j}$ averages each task's loss over its observed labels. This excludes unavailable targets from the loss.

\paragraph{Overall training objective.}
The overall training objective $\mathcal{L}$ aggregates the per-task cross-entropy losses of every prediction head with elastic-net regularization on model parameters $\boldsymbol{\theta}$:
\begin{equation}
\begin{aligned}
\mathcal{L}
&= (1-\alpha) \Bigg(
    \sum_{m} \lambda_{m} \sum_{j=1}^{10} \mathcal{L}^{(m)}_{\mathrm{CE},j}
 \Bigg) \;+\; \alpha \left(\gamma \lVert \boldsymbol{\theta} \rVert_{1}
 \;+\; \tfrac{1-\gamma}{2} \lVert \boldsymbol{\theta} \rVert_{2}^{2} \right)
\end{aligned}
\label{eq:loss}
\end{equation}
where $\lambda_{m}$ are the branch weights over the prediction heads defined above, $\gamma \in [0,1]$ balances $L_1$ and $L_2$ penalties, and $\alpha \in [0,1]$ trades off predictive loss against regularization. All ten binary subtasks have equal weight within a head; no task-specific weights are fitted or tuned.
For NEO's neural-network-based pCR prediction model and the uni-modal direct pCR prediction model, the outer sum contains one term with $\lambda_{m}=1$.
The separate regression objective for MORPHEUS is given below.
For the three-head fusion models the weights are parameterized by two free values $\lambda_1$ and $\lambda_2$ as $\lambda_{\mathrm{multimodal}} = \lambda_1$, $\lambda_{\mathrm{tabular}} = (1-\lambda_1)\lambda_2$, and $\lambda_{\mathrm{pathology}} = (1-\lambda_1)(1-\lambda_2)$, so that the three weights sum to one.
$\alpha$, $\gamma$, and, where the architecture has more than one head, $\lambda_1$ and $\lambda_2$ were tuned by Bayesian hyperparameter search (\Cref{tab:hyperparameters,sec:hyperparameters}).

\subsubsection{MORPHEUS training objective}\label{sec:morpheus_objective}

MORPHEUS (\Cref{sec:gene_expression_model}) uses the same regularization structure as \Cref{eq:loss}, with the classification losses replaced by mean squared error over the measured expression targets:
\begin{equation}
\begin{aligned}
\mathcal{L}_{\mathrm{MORPHEUS}}
&= (1-\alpha) \Bigg( \sum_{m} \lambda_{m}\, \mathcal{L}^{(m)}_{\mathrm{MSE}} \Bigg)
 \;+\; \alpha \left(\gamma \lVert \boldsymbol{\theta} \rVert_{1}
 \;+\; \tfrac{1-\gamma}{2} \lVert \boldsymbol{\theta} \rVert_{2}^{2} \right),
\\[2pt]
\mathcal{L}^{(m)}_{\mathrm{MSE}}
&= \frac{1}{N G} \sum_{i=1}^{N} \sum_{g=1}^{G}
   \left( \hat{y}^{(m)}_{i,g} - y_{i,g} \right)^{2},
\end{aligned}
\label{eq:loss_morpheus}
\end{equation}
where $N$ is the batch size in slides, $G=14{,}773$ is the number of protein-coding genes, $y_{i,g}$ is the VST-normalized expression of gene $g$ (standardized using each gene's TCGA training-subset mean and standard deviation) in the patient contributing slide $i$, and $m$, $\lambda_{m}$, $\alpha$, and $\gamma$ carry the same meaning as in \Cref{eq:loss}.
MORPHEUS instantiates only the pathology branch, so the outer sum contains a single term, with $\lambda_{\mathrm{pathology}}=1/3$, a fixed scaling of the data-fitting term relative to the regularization term.
$\alpha$ and $\gamma$ were tuned over the same ranges as for the pCR prediction models (\Cref{tab:hyperparameters}).

\subsubsection{EBM training within MSCV}\label{sec:ebm_mscv_integration}

EBM training used the same cohort partitions as the tabular ResNet (\Cref{subsec:mscv-main}). To impose these partitions, we set the number of outer bags to one and supplied a custom bag-assignment vector for each fold. Training observations were assigned $+1$ and observations from the held-out cohort $-1$, reserving the latter for internal early stopping. One EBM was fitted per MSCV fold.

\subsubsection{Hyperparameter tuning}\label{sec:hyperparameters}

Hyperparameter search covered MORPHEUS and every neural-network-based pCR prediction model. Each search space combined the shared parameters with those of the model's components (\Cref{fig:model_components,tab:hyperparameters}). The first half of the trials used random sampling; the remainder used Bayesian hyperparameter search with BoTorch~\citep{balandat2020botorch}.

All neural-network-based pCR prediction models used identical ranges for shared parameters, a log-uniform learning-rate range of $(10^{-3},3\times10^{-2})$, and 100 trials. Each trial ran for at most 300 epochs, with early stopping after 20 epochs without improvement in validation pCR AUROC. Configuration selection used mean AUROC across the MSCV validation cohorts (\Cref{sec:optimization_detail}). These matched settings applied to the transcriptome-informed and direct pCR prediction models and to the clinical-only baseline (\Cref{fig:tech_interpretability}a).

MORPHEUS used 50 trials, at most 100 epochs per trial, and an early-stopping patience of 15 epochs. Configuration selection maximized validation Kendall's $\tau$ averaged over all protein-coding genes. Learning rates were sampled log-uniformly from $(10^{-5},10^{-2})$ for Falcon-g and $(10^{-3},3\times10^{-2})$ for Falcon-L. MORPHEUS was tuned on TCGA and then frozen before training pCR prediction models. Its expression-learning objective and search budget were separate from those of the pCR prediction models.

In the tabular ResNet, block size is the residual block's input and output width; the internal hidden width is the integer part of block size multiplied by the hidden dimension multiplier.

EBMs were not tuned; their fixed settings are described in \Cref{sec:ebm_mscv_integration}.

\begin{table}[t]
\centering
\caption{
    \scriptsize
\textbf{Hyperparameter search space, organized by model component.} Each model's search space is the union of the shared block and the blocks for the components it instantiates (\Cref{sec:hyperparameters}). MORPHEUS and the uni-modal direct pCR prediction model use the pathology branch, the transcriptome-informed pCR prediction models use the tabular branch, and the multi-modal direct pCR prediction model uses both branches together with the fusion module. During the random-search phase, parameters were sampled independently from the distributions below; subsequent configurations were selected by Bayesian hyperparameter search. $\mathcal{U}$ denotes a uniform and $\log\mathcal{U}$ a log-uniform distribution over the stated range. Draws for integer-valued parameters (intermediate feature dimension, block size) were rounded to the nearest integer, and discrete sets were sampled uniformly. Learning rate and search budget differ between models and are given in the text.
}
\label{tab:hyperparameters}
\begin{tabular}{ll}
\toprule
\textbf{Hyperparameter} & \textbf{Range} \\
\midrule
\multicolumn{2}{l}{\textit{Shared by all tuned neural networks}} \\
Elastic-net regularization weight ($\alpha$) & $\log\mathcal{U}(10^{-6}, 10^{-3})$ \\
Elastic-net mixing weight ($\gamma$) & $\mathcal{U}(0, 1)$ \\
Intermediate feature dimension ($d_{\mathrm{int}}$) & $\log\mathcal{U}(4, 512)$ \\
\midrule
\multicolumn{2}{l}{\textit{Pathology branch}} \\
Pathology-branch dropout & $\mathcal{U}(0, 0.3)$ \\
\midrule
\multicolumn{2}{l}{\textit{Tabular branch}} \\
Number of residual blocks ($n_{\mathrm{blocks}}$) & $\{1,\ldots,6\}$ \\
Block size & $\log\mathcal{U}(4, 32)$ \\
Hidden dimension multiplier & $\mathcal{U}(1.0, 3.0)$ \\
Tabular-branch dropout & $\mathcal{U}(0, 0.3)$ \\
\midrule
\multicolumn{2}{l}{\textit{Fusion module}} \\
Fusion attention hidden dimension multiplier ($\gamma_{\mathrm{att}}$) & $\mathcal{U}(0.1, 3.0)$ \\
Fusion-module dropout & $\mathcal{U}(0, 0.3)$ \\
Branch weights ($\lambda_1$, $\lambda_2$) & $\mathcal{U}(0, 1)$ \\
\bottomrule
\end{tabular}
\end{table}

\subsubsection{Calibration of pCR probabilities}\label{sec:calibration}

Predicted pCR probabilities were calibrated using beta calibration~\citep{kull2017beta}.
A separate calibrator was fitted for each MSCV fold using predictions for its held-out cohort $D_{J+k}$. The inputs were the mean predictions of the models with top-10 hyperparameters or the predictions of the single EBM fitted with fixed settings.
Evaluation cohorts were excluded from fitting, model selection, and calibration.

For each model class and fold, the calibrator was applied to the corresponding model's predictions on the evaluation cohorts (\Cref{alg:atxneo}).
Calibrated probabilities were then averaged with equal weight across folds, model classes, and Falcon variants (\Cref{sec:response_model,sec:backbone_combination}).
Calibration in the evaluation cohorts was assessed as described in \Cref{sec:calibration_assessment} and shown in \Cref{fig:performance}b.

\subsubsection{Ensembling Falcon-g and Falcon-L}\label{sec:backbone_combination}

Each model using Falcon features was fitted separately for Falcon-g and Falcon-L.
For NEO and the pCR ablations, predictions were calibrated within each variant and MSCV fold, then averaged across variants at the probability level.

For gene-subset selection (\Cref{sec:gene_list_resolution}), patient-level inferred expression was averaged across the two Falcon variants before ranking genes.
This produced a common gene ranking for the ablations, which used both variants.

Descriptive expression analyses used Falcon-g: per-gene fidelity (\Cref{fig:ed_morpheus_fidelity,sec:fidelity_metric}), pathway enrichment (\Cref{fig:morpheus}a, \Cref{fig:ic_pathways}), patch-level pathway scores (\Cref{fig:morpheus}b), and spatial examples (\Cref{fig:morpheus}c).
The EBM importance and shape-function plots also used Falcon-g (\Cref{fig:tech_interpretability}b).

\subsubsection{Direct pCR prediction models}\label{sec:direct_path_model}

To evaluate the contribution of inferred expression, we trained two direct pCR prediction models that predict pCR directly from Falcon embeddings, bypassing MORPHEUS and ICA (\Cref{fig:tech_interpretability}a).
The uni-modal one used a MIL aggregator alone.
The multi-modal one combined its output with clinical variables through gated-attention fusion (\Cref{sec:fusion_detail,fig:model_components}).
Multi-slide aggregation also differs from NEO: the direct pCR prediction models average slide-level pCR probabilities, whereas transcriptome-informed pCR prediction models average slide-level inferred expression before ICA and response prediction (\Cref{sec:ica_readout}).

These neural-network-based pCR prediction models used the same NAT development patients, MSCV partitions, evaluation cohorts, and tuning budgets (\Cref{sec:hyperparameters}). Models were fitted separately for each Falcon variant and combined by averaging calibrated probabilities (\Cref{sec:backbone_combination}).

This ablation covers only the neural-network component of NEO so that the transcriptome-informed and direct pCR prediction models are matched in architecture family.
The EBM has no counterpart that aggregates patch embeddings by MIL.
The full NEO ensemble, which adds the EBM, achieved a pooled AUROC of 0.792 (\Cref{fig:performance}), compared with 0.785 for the multi-modal neural-network-based pCR prediction model alone (\Cref{fig:tech_interpretability}a).

\subsection{Evaluation and statistical analysis}\label{sec:supp_evaluation}

\subsubsection{Pooling across cohorts}\label{sec:pooling_detail}

AUROCs and paired AUROC differences were pooled across cohorts using DerSimonian--Laird random-effects models~\cite{dersimonian1986meta} (\Cref{sec:method_eval,sec:paired_diff}). Within each cohort, confidence intervals were estimated by patient-level bootstrap with 2,000 resamples; resamples containing only one outcome class were discarded. The sample standard deviation of the retained bootstrap replicates supplied the cohort-level standard error. Paired comparisons used the same resampled patients for both predictors.
Adjusted odds ratios used the same pooling framework on the log-odds-ratio scale, with analytic Wald standard errors from Firth-penalized logistic regression instead of bootstrap estimates (\Cref{sec:adjusted_analysis}).

The number of contributing cohorts is reported with each pooled estimate. For primary discrimination and adjusted association, heterogeneity is summarized by $\tau^2$, the estimated between-cohort variance on the analysis scale, and $I^2$, the estimated percentage of variation attributable to between-cohort differences rather than sampling error. An $I^2$ of zero indicates no detected excess variation beyond sampling error.

Pooled confidence intervals and tests used the Hartung--Knapp adjustment~\cite{inthout2014hartung} to account for uncertainty with few cohorts. The adjustment changes the interval and test but leaves the pooled estimate and cohort weights unchanged. Two-sided tests used a $t$ distribution with $k-1$ degrees of freedom, where $k$ is the number of cohorts, against the stated null.

For the pooled ROC curves (\Cref{fig:performance}a), each cohort's empirical curve was interpolated onto a common false-positive-rate grid. True-positive rates were then averaged using the random-effects cohort weights.

\subsubsection{Paired comparisons between scores}\label{sec:paired_diff}

Paired AUROC comparisons were used for NEO versus computational TIL biomarkers and Ki-67 (\Cref{fig:performance}d,e; \Cref{fig:paired_comparisons}a--e), and for the model ablations (\Cref{fig:tech_interpretability}a; \Cref{fig:paired_comparisons}f--h).
Within each cohort, both scores were evaluated on the same bootstrap resample of patients and their AUROC difference recorded. Resamples with only one outcome class were discarded. Percentile intervals and the standard deviation of the replicates supplied the cohort-level confidence interval and standard error.

Cohort-level differences were pooled on the difference scale using the random-effects model and Hartung--Knapp adjustment in \Cref{sec:pooling_detail}, with a two-sided test against zero.
Cohort eligibility followed \Cref{sec:method_eval}.
Because the model fitted to paired differences assigns different weights from models fitted to each AUROC separately, the pooled difference need not equal the difference between pooled AUROCs.

\subsubsection{Covariate coding and fitting for the adjusted analysis}\label{sec:adjusted_detail}

The adjusted association analysis used the pre-specified covariates in \Cref{sec:adjusted_analysis}.
Unresolved equivocal HER2 status and unspecified nodal stage (N$+$) were treated as missing.
Within each cohort, constant, duplicated, or otherwise linearly dependent covariates were dropped to obtain a full-rank design.

We used Firth-penalized logistic regressions~\citep{heinze2002solution}.
Each cohort contributed its NEO-score log odds ratio and the analytic Wald standard error derived from the fitted model's information matrix.
These entered the random-effects model as the effect estimate and standard error.

\subsection{Benchmark biomarkers}\label{sec:benchmark_biomarkers}

NEO's AUROC was compared with that of four computational TIL biomarkers (cTILs, LTR, ITR, and muTILs) and pathologist-assessed Ki-67 (\Cref{fig:performance}d,e), and separately within subtype (\Cref{fig:subtype_biomarkers}).
The computational TIL biomarkers were also compared for robustness to tumor sampling bias (\Cref{fig:sampling_bias}c).

The cTILs, LTR, and ITR implementations of \citet{aswolinskiy2023proacting} were distributed as a container on grand-challenge.org, with access granted on request.
An attempt to run the container in August 2025 failed with an unresolved runtime error; a subsequent access request in April 2026 remained pending at submission.
We therefore re-implemented these metrics from their published definitions using a publicly available breast tissue segmentation model.
The published approach permits alternative models that segment tumor, stroma, and lymphocytes, or manual segmentation.
Our substitute provides these classes (\Cref{sec:ctils}), but differences in segmentation can affect performance (\Cref{sec:reproduction_check}).
We did not reproduce the mitotic rate (MTR) biomarker because its required mitosis-detection network was not publicly distributed.

For muTILs~\citep{liu2024panoptic}, we used the released trained models (\Cref{sec:mutils}).
Computational TIL comparisons used shared slides (\Cref{sec:shared_slides}). The Ki-67 comparison used a subset of patients in the evaluation set with recorded Ki-67 (\Cref{sec:Ki-67}).

\subsubsection{cTILs (computational tumor-infiltrating lymphocytes)}
\label{sec:ctils}

cTILs~\citep{aswolinskiy2023proacting} is the ratio of lymphocyte area to lymphocyte-plus-stroma area within a tumor-bulk envelope.
We derived the required tissue masks using TIAToolbox~\citep{Pocock2022} and its \texttt{fcn\_resnet50\_unet-bcss} model, trained on the Breast Cancer Semantic Segmentation dataset~\citep{amgad2019structured}.
The model classifies pixels as tumor, stroma, inflammatory tissue, necrosis, or other; the inflammatory class was used as the lymphocyte mask.

Input and output resolution were set to 0.25\,$\mu$m per pixel.
Tile predictions were stitched and stored at 2\,$\mu$m per pixel, an eightfold downsampling in each dimension.
The resulting $(H\times W\times5)$ class probabilities were converted to labels by channel-wise argmax.
All subsequent morphological operations used this 2\,$\mu$m resolution.

The tumor-bulk envelope was constructed in three steps.
First, the tumor mask was closed with a disk-shaped structuring element of radius 100\,$\mu$m to connect nearby regions and fill small gaps.
Second, connected components smaller than 0.1\,mm\textsuperscript{2} were removed.
Third, retained regions were dilated by 50\,$\mu$m to include surrounding tumor-associated stroma.
Within this envelope, cTILs was computed as inflammatory area divided by inflammatory-plus-stromal area.
Scores were undefined if no tumor was detected, no component survived the area threshold, or the envelope contained neither stroma nor lymphocytes.
These failures were excluded from the shared-slide set (\Cref{sec:shared_slides}).

\subsubsection{LTR (lymphocytes-to-tumor ratio) and ITR (inflamed tumor ratio)}
\label{sec:ltr_itr}

LTR and ITR were computed from the same BCSS segmentation as cTILs, following the definitions of \citet{aswolinskiy2023proacting}.
LTR is the slide-wide ratio of lymphocyte area to lymphocyte-plus-tumor area.
ITR is the fraction of total tumor area within 80\,$\mu$m of any lymphocyte region.

\subsubsection{muTILs}
\label{sec:mutils}

muTILs~\citep{liu2024panoptic} jointly segments tissue regions and individual nuclei to derive a stromal-TIL score.
We applied the released model to all evaluation cohorts and reported its ``saliency-weighted nTnS'' score, identified by the authors as the most prognostic variant in their validation.
For each tissue-containing tile, the score is the number of TIL cells divided by the total number of stromal-region cells.
Tile scores are averaged with weights proportional to the fraction of tile area occupied by peritumoral stroma.

\subsubsection{Reproduction of performance of the re-implemented biomarkers}\label{sec:reproduction_check}
IMPRESS was the only evaluation cohort with published performance for cTILs, LTR and ITR~\citep{aswolinskiy2023proacting}.
We compared the published and recomputed AUROCs on the same 126 patients, stratified by subtype (\Cref{tab:impress_reproduction}).
Our cTILs implementation reproduced the published results, whereas we were not able to reproduce the results reported for LTR and ITR.
NEO's comparisons with these re-implemented markers should therefore be interpreted in light of the different segmentation models.

\begin{table}[H]
\centering
\footnotesize
\caption{
    \scriptsize
\textbf{Reproduction of the re-implemented computational TIL biomarkers on IMPRESS.} AUROC for pCR (95\% bootstrap confidence interval) among the IMPRESS patients of each stratum. \textbf{As published} is the value the originating study reports on these same patients: \citet{aswolinskiy2023proacting} for cTILs, LTR, ITR, and MTR; NEO is shown for reference. \textbf{Recomputed} is our implementation (\Cref{sec:ctils,sec:ltr_itr}), scored on the identical patients with the bootstrap of \Cref{sec:method_eval}.}
\label{tab:impress_reproduction}
\begin{tabular}{lcc}
\hline
\textbf{Method} & \textbf{As published} & \textbf{Recomputed} \\
\hline
\multicolumn{3}{c}{\textbf{TNBC} $N = 64$ ($N_\text{pCR} = 27$)} \\
\hline
cTILs & 0.708 (0.574--0.841) & 0.691 (0.553--0.820) \\
LTR & 0.686 (0.550--0.821) & 0.583 (0.439--0.727) \\
ITR & 0.715 (0.584--0.845) & 0.568 (0.430--0.713) \\
\hline
\multicolumn{3}{c}{\textbf{HER2$+$} $N = 62$ ($N_\text{pCR} = 38$)} \\
\hline
cTILs & 0.679 (0.543--0.814) & 0.623 (0.481--0.760) \\
LTR & 0.682 (0.544--0.820) & 0.598 (0.448--0.745) \\
ITR & 0.664 (0.527--0.802) & 0.434 (0.284--0.590) \\
\hline
\end{tabular}
\end{table}

\subsubsection{Shared-slide evaluation set}
\label{sec:shared_slides}
Comparisons between NEO and computational TIL biomarkers used only slides scored by all five methods (\Cref{fig:performance}d; \Cref{fig:subtype_biomarkers}).
This shared-slide set comprised 2,381 slides from 1,386 patients across the nine evaluation cohorts.
For patient-level AUROC analyses, computational TIL scores were averaged across each patient's retained slides.
The sampling-stability comparison used slide-level predictions for the 242 UChicago patients with at least two shared slides (\Cref{fig:sampling_bias}c; \Cref{sec:sampling_robustness}).

Of 2,489 evaluation slides, LTR and ITR returned scores for all slides, whereas cTILs was undefined for 12 slides without detectable tumor.
muTILs returned usable scores for 2,388 slides (95.9\%).
Its 101 failures comprised metric-summarization errors in slides without a salient region of interest ($n=43$), no detectable tissue ($n=48$), and other technical errors ($n=10$).
Five slides failed both cTILs and muTILs, giving 108 excluded slides in total.
Twenty-six patients had no slide successfully scored by all methods and were therefore excluded.

\subsubsection{Ki-67}
\label{sec:Ki-67}

Ki-67 proliferation index was recorded for 237 patients across the Jagiellonian, Shaare Zedek, and \L{}\'od\'z cohorts. NEO and Ki-67 were compared in these patients (\Cref{fig:performance}e), independently of the shared-slide restriction for computational TIL biomarkers. Ki-67 values were used as reported in the source datasets.

\subsection{Gene list and dimensionality ablations}\label{sec:supp_additional}

\subsubsection{Gene list selection and expression-only ablation}
\label{sec:gene_list_resolution}

The gene list ablation compared random subsets, subsets ranked by expression fidelity or pCR discrimination, and published signatures (\Cref{fig:tech_interpretability}c).
For each subset, inferred expression was reduced to five ICA components fitted on the TCGA validation subset.
EBMs were trained on these components alone under MSCV, separately for each Falcon variant, and their calibrated probabilities were averaged (\Cref{subsec:mscv-main,sec:backbone_combination}).

Random gene lists were sampled 25 times at each size.
To construct the ranked subsets, genes were scored using expression averaged across Falcon variants (\Cref{sec:backbone_combination}).
Expression fidelity rankings used per-gene Kendall's $\tau$ in the TCGA validation subset (\Cref{sec:fidelity_metric}).
pCR discrimination rankings used $|\mathrm{AUROC}-0.5|$, calculated separately in each of the three NAT cross-validation cohorts and then averaged.

Published signatures came from nine sources spanning clinical assays, immune-cell markers, tumor evolution and molecular pathways (\Cref{tab:gene_list_summary}).
Gene symbols were mapped to Ensembl identifiers~\citep{yates2026ensembl} and intersected with the 14,773-gene MORPHEUS output. Legacy symbols were updated to HUGO Gene Nomenclature Committee (HGNC) names for MammaPrint (for example, \textit{KNTC2}$\to$\textit{NDC80}, \textit{ORC6L}$\to$\textit{ORC6}, and \textit{WISP1}$\to$\textit{CCN4}), Oncotype DX (\textit{STK15}$\to$\textit{AURKA} and \textit{CTSL2}$\to$\textit{CTSV}), and HER2DX (\textit{IGJ}$\to$\textit{JCHAIN} and \textit{GAPD}$\to$\textit{GAPDH}).
Other signatures used the published symbols.

MammaPrint's 70 microarray reporters mapped to 52 unique protein-coding genes before expression filtering: some reporters duplicated genes, while others represented expressed sequence tags without an assigned protein-coding gene.
HER2DX lists 27 targets and five normalization genes~\citep{prat2022development}.
Its IGL symbol denotes an immunoglobulin locus without a unique Ensembl identifier, leaving 31 symbols for resolution.
Across gene lists, retained counts reflect symbol resolution, deduplication, exclusion of reporters without protein-coding assignments, and the expression filter in \Cref{sec:expression_preprocessing}.

\begin{figure}[H]
    \centering
    \includegraphics[width=0.45\linewidth]{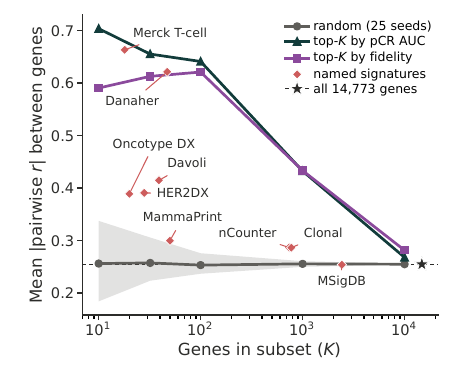}
    \caption{
    \scriptsize
    \textbf{Between-gene redundancy of the gene subsets of \Cref{fig:tech_interpretability}c.}
    Redundancy is measured as the mean absolute pairwise Pearson correlation in H\&E-inferred expression from the TCGA validation subset (\Cref{sec:backbone_combination}).
    Gene lists ranked by expression fidelity (purple) or per-gene pCR AUROC (green) have higher correlations than random gene lists of the same size (gray; mean and range across 25 seeds), with differences narrowing as gene lists approach all 14,773 genes.
    Red diamonds indicate published signatures; the star and dashed line indicate the full gene set.
    The ranked subsets accumulate mutually correlated genes, which may help explain their limited gains as gene list size increases (\Cref{fig:tech_interpretability}c).}
    \label{fig:gene_redundancy}
\end{figure}

To characterize gene redundancy, we calculated the mean absolute pairwise Pearson correlation in inferred expression from the TCGA validation subset for each gene list (\Cref{fig:gene_redundancy}).

\begin{table}[ht]
\centering
\footnotesize
\caption{
    \scriptsize
\textbf{Gene signature sources and retained gene counts.} ``Total'' is the published gene list size, including controls where applicable; for MammaPrint it counts reporters rather than unique genes. ``Retained'' counts unique genes in the 14,773-gene MORPHEUS output after gene symbol mapping, deduplication, and expression filtering.}
\label{tab:gene_list_summary}
\begin{tabular}{l l r r }
\hline
\textbf{Signature} & \textbf{Source} & \textbf{Total} & \textbf{Retained} \\
\hline
Merck T-cell-inflamed GEP & \citet{ayers2017ifn} & 18 & 18 \\
Oncotype DX & \citet{paik2004multigene} & 21 & 20 \\
HER2DX & \citet{prat2022development} & 32 & 28 \\
Davoli immune evasion & \citet{davoli2017tumor} & 42 & 39 \\
Danaher immune cell markers & \citet{danaher2017gene} & 60 & 47 \\
MammaPrint & \citet{van2002gene} & 70 & 50 \\
nCounter BC360 & (commercial gene list) & 778 & 737 \\
Clonal & \citet{biswas2019clonal} & 897 & 775 \\
MSigDB  & \citet{chen2021multimodal}, \citet{liberzon2011molecular} & 3,047 & 2,427 \\
\hline
\end{tabular}
\end{table}

\subsubsection{Reduction method and dimensionality}\label{sec:reduction_ablation}
To assess sensitivity to the reduction method and representation size (\Cref{fig:tech_interpretability}d), the 2,427-gene MSigDB-family subset of the 14,773 inferred genes (\Cref{sec:ica_readout}) was reduced to 1--200 components using ICA, principal component analysis, or sparse principal component analysis. Each transformation was fitted on the TCGA validation subset. An EBM was then trained on the resulting components together with clinical variables.
This analysis is descriptive.
It was run on the evaluation cohorts and did not inform selection of NEO's five-component ICA representation, which was fixed during development (\Cref{subsec:mscv-main,sec:ica_readout}).

\subsection{Biological interpretation}

\subsubsection{Interpreting an EBM}\label{sec:ebm_interpretation}
The EBM's predicted log-odds are the sum of feature-specific main effects, an intercept, and fitted interaction terms.
The interpretation plots in \Cref{fig:tech_interpretability}b summarize its thirteen main effects.
For each patient, every feature contributes one value to the predicted log-odds.
Feature importance is the mean absolute value of that contribution over the development cohorts.

Shape functions plot that contribution against the value of the feature.
Each function is centered vertically so that its mean over the development cohorts is zero.
For the Falcon-g models, shape functions and feature importance values were averaged across the three cross-validation folds.
Shaded bands indicate $\pm$1 standard deviation across folds.
Pairwise interaction terms are omitted from both the importance ranking and the shape-function plots.

Individual feature importance does not show how much a set of related features moves the prediction together.
We therefore also report the importance of two groups: the five MORPHEUS-ICs, and the eight clinical variables (\Cref{fig:tech_interpretability}b).
For each patient, the contributions of a group's members are first summed.
Group importance is the mean absolute value of that sum over the development cohorts~\citep{bosschieter2026most}.
Summing before taking the absolute value lets one feature's positive contribution cancel another's negative one, so group importance captures how far the group of features in aggregate moves a patient's predicted log-odds.
It is not a mere sum of individual feature importance values and can be considerably smaller.

\subsubsection{Spatially resolved examples of the H\&E-inferred expression}\label{sec:spatial_methods}

Patch-level expression predictions were compared with measurements from two public breast cancer sections released by 10x Genomics (\Cref{fig:morpheus}c). One section~\cite{janesick2025biomarker} was profiled with a Xenium v1 280-target custom breast gene list; the other used a Xenium Prime gene list with 5,101 targets.

Each section was tiled on a grid of 224$\times$224-pixel patches, corresponding to 112\,$\mu$m squares (\Cref{sec:falcon}).
Patches were retained if they lay entirely within the transcript footprint and contained at least 80\% H\&E foreground.
MORPHEUS was applied independently to each retained patch. Neither section nor any spatial transcriptomic data were used to train, select, or calibrate MORPHEUS.

For each gene and patch, we counted transcripts with quality value $\ge20$ whose centers lay within the patch.
Measured expression was computed as $\log(1+10^4 c_{i,g}/\max(L_i,1))$, where $c_{i,g}$ is the count for gene $g$ in patch $i$ and $L_i$ is the patch's total count across the measured gene list.
Patches with no transcripts had zero measured expression. Prediction and measurement represented the same tissue area. Agreement was assessed by Spearman correlation across unsmoothed patch-level values.

One gene per program is displayed in \Cref{fig:morpheus}c.
Genes were chosen from the members driving the corresponding hallmark set's enrichment in TCGA-BRCA: those preceding the peak of its running enrichment score (\Cref{fig:morpheus}a).

\subsubsection{Pathway analyses of the MORPHEUS-ICs}\label{sec:ic_pathways}
\begin{figure}[H]
    \centering
    \includegraphics[width=\linewidth]{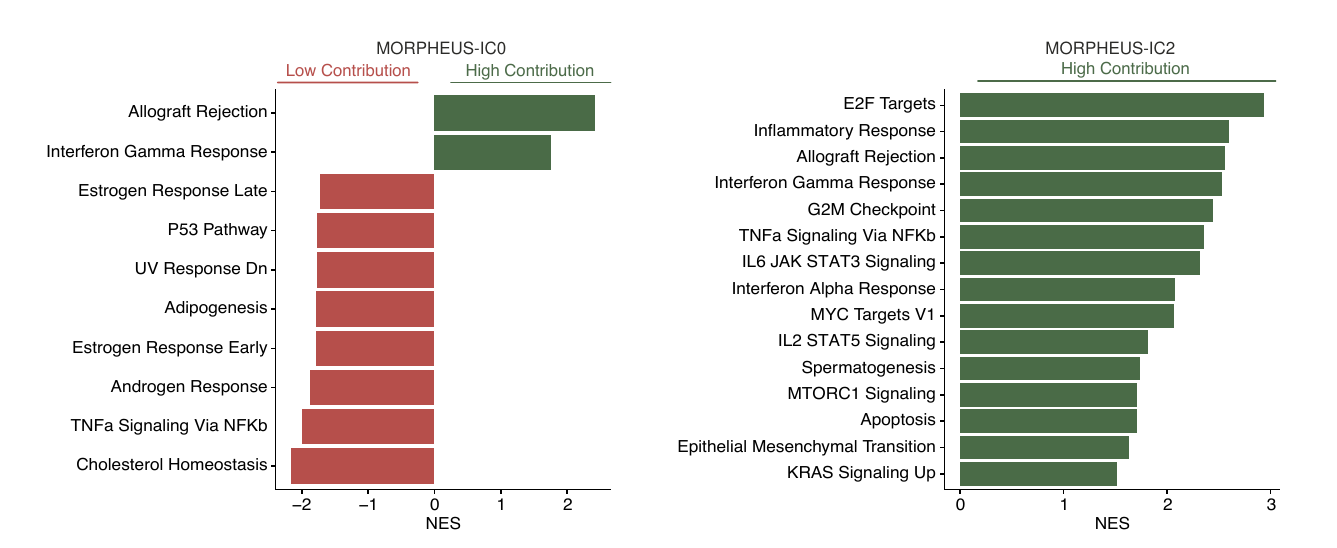}
    \caption{
    \scriptsize
    \textbf{Pathway analysis of the two most important MORPHEUS-ICs.}
    Gene set enrichment analysis of the hallmark pathway collection over all genes ranked by their weight in the ICA mixing-matrix column of MORPHEUS-IC0 (left) and MORPHEUS-IC2 (right) (\Cref{sec:ica_readout,sec:bio_correlates_methods}). Bars are the normalized enrichment scores (NES) of the significantly enriched pathways: green, pathways whose genes have a high contribution to the component; red, a low contribution.}
    \label{fig:ic_pathways}
\end{figure}

To annotate the two MORPHEUS-ICs most important to NEO's EBM (\Cref{fig:ic_pathways}), genes were ranked by their weight in each component (\Cref{sec:ica_readout}) and GSEA was performed against the Hallmark pathways as described in \Cref{sec:bio_correlates_methods}.
A positive NES indicates that a pathway's genes are concentrated among the component's most positively weighted genes, and a negative NES among its more negatively weighted genes.
MORPHEUS-IC0 was enriched for interferon gamma response, consistent with an enriched adaptive immune response.
MORPHEUS-IC2 was instead enriched for proliferation and growth programs with elevated cell turnover: E2F targets, G2M checkpoint, MYC targets V1, MTORC1 signaling, and apoptosis.

\subsubsection{Molecular correlates of NEO predictions}\label{sec:neo_correlates_methods}

\begin{figure}[H]
    \centering
    \includegraphics[width=0.9\linewidth]{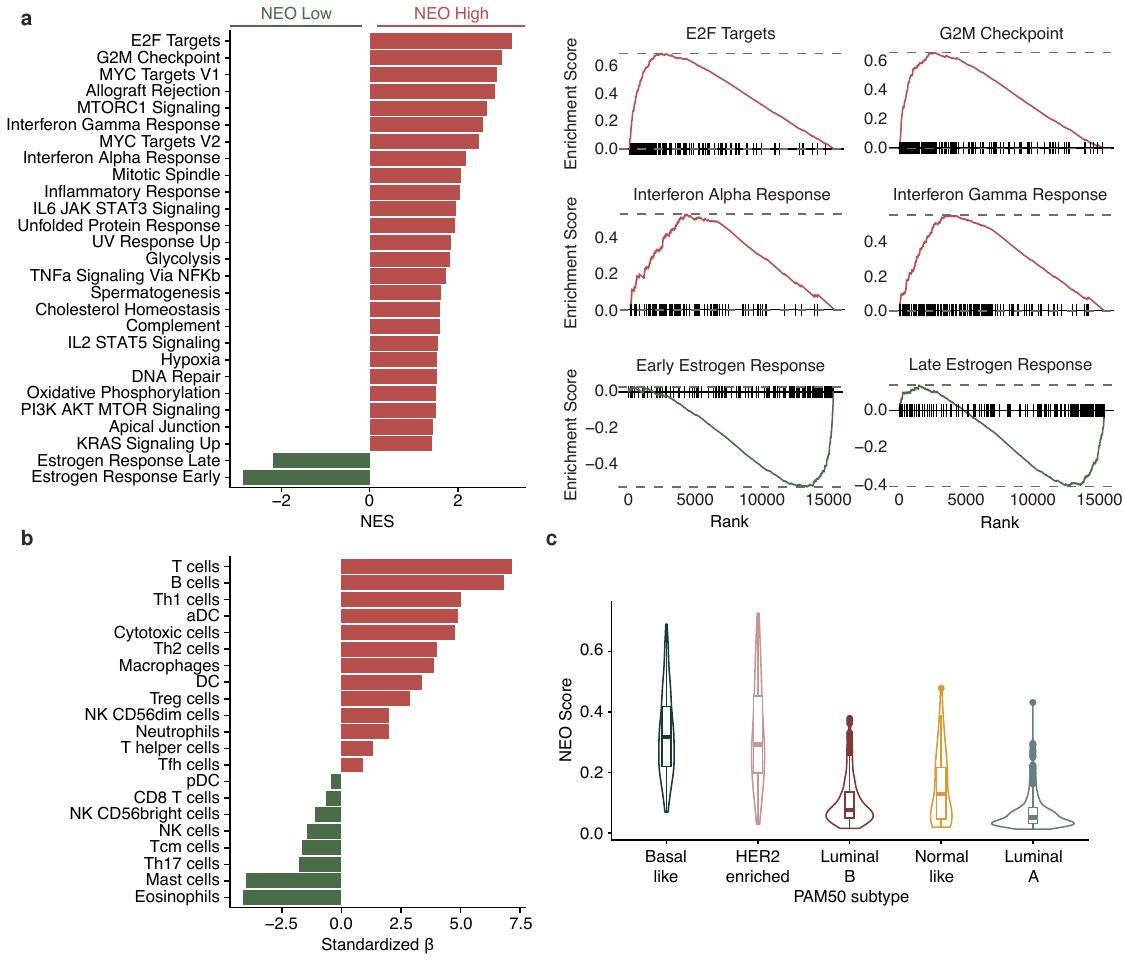}
    \caption{
    \scriptsize
    \textbf{Biological correlates of NEO predictions.}
    All analyses use the TCGA-BRCA cohort with MORPHEUS training patients included (\Cref{sec:neo_correlates_methods}).
    \textbf{a}, GSEA of the hallmark pathway collection over genes ranked by their association with the NEO score. \emph{Left}, NES of the significantly enriched pathways; red, enriched in NEO-high tumors; green, in NEO-low tumors. \emph{Right}, enrichment plots for representative pathways: proliferation (E2F targets, G2M checkpoint) and interferon response, enriched in NEO-high tumors, and early and late estrogen response, enriched in NEO-low tumors.
    \textbf{b}, Association between immune cell-type enrichment scores and the NEO score (standardized $\beta$ per cell type); red, positively associated cell types; green, negatively associated.
    \textbf{c}, Distribution of NEO scores by PAM50 intrinsic subtype.}
    \label{fig:neo_bio_correlates}
\end{figure}

NEO scores were related to measured bulk expression in 1,053 TCGA-BRCA patients with matched profiles from cBioPortal~\cite{cerami2012cbio,gao2013integrative,de2023analysis} (\Cref{fig:neo_bio_correlates}).
This cohort slightly differs from the 1,047 BRCA patients used for MORPHEUS (\Cref{tab:morpheus_breakdown_tcga}): patients whose slides could not be assigned to a single primary-tumor RNA-seq sample were retained here, whereas patients without expression data in the cBioPortal study were excluded.
The analysis included MORPHEUS training patients, although no TCGA patient contributed NAT outcome labels to the pCR prediction model (\Cref{sec:response_model}).
Closer agreement between inferred and measured expression in MORPHEUS training patients may overstate association strengths relative to an independent cohort.

To identify transcriptional programs associated with NEO scores (\Cref{fig:neo_bio_correlates}a), we performed differential expression with DESeq2~\cite{love2014moderated} and used the resulting test-statistic ranking for GSEA of hallmark pathways as in \Cref{sec:bio_correlates_methods}. To profile the immune microenvironment (\Cref{fig:neo_bio_correlates}b), we computed single-sample GSEA scores for the 23 immune cell-type signatures of \citet{bindea2013spatiotemporal} and fitted a linear model with the NEO scores as the dependent variable. Benjamini--Hochberg correction~\cite{benjamini1995controlling} was applied across tested signatures.
We also summarized NEO scores by PAM50 intrinsic subtype (\Cref{fig:neo_bio_correlates}c).

\subsection{Exploratory secondary endpoints}

\subsubsection{Residual cancer burden}\label{sec:rcb_ordinal}

NEO's overall-pCR score was evaluated without retraining against residual cancer burden (RCB), which grades residual disease from RCB-0 (no invasive disease) to RCB-III (extensive disease). RCB was recorded in 364 of 1,412 evaluation patients: 126/126 in IMPRESS, 107/116 in Jagiellonian, 99/153 in Nexus, 29/51 in \L{}\'od\'z, and 3/146 in Dordrecht. All estimates are conditional on RCB availability.

For the ordinal analysis (\Cref{fig:secondary_endpoints}a), the minimum-pair rule in \Cref{sec:method_eval} was applied to pairs of patients in different RCB classes. Dordrecht contributed only two such pairs (one RCB-0 patient and two RCB-II patients) and was excluded, leaving 361 patients in four cohorts.
The trend across RCB classes was assessed by a cohort-stratified Jonckheere--Terpstra test.
The test statistic, null mean, and null variance were computed within each cohort and summed across cohorts, so that all patient comparisons were within cohort. The $p$ value is one-sided to test whether NEO scores decreased with increasing residual burden.

For the binary contrast in \Cref{fig:secondary_endpoints}b, RCB-0/I was treated as the favorable outcome and RCB-II/III as the unfavorable outcome. Each of the four retained cohorts also contributes at least thirty favorable--unfavorable pairs, satisfying the AUROC inclusion rule in \Cref{sec:method_eval}. AUROCs were estimated within cohort and pooled as described in \Cref{sec:pooling_detail}. Because RCB-0 is pCR, this endpoint partly overlaps the primary endpoint.

\subsubsection{Nodal pCR}\label{sec:nodal_pcr}
Nodal pCR was evaluated among patients with documented clinical nodal involvement at baseline (cN1, cN2, cN3, or unspecified N$+$).
Patients with cN0, cNX, or missing clinical nodal stage were excluded.
IMPRESS recorded no clinical nodal stage and contributed no patients, leaving 712 eligible patients across eight cohorts (\Cref{fig:secondary_endpoints}c).

Nodal pCR was assigned when ypN0 or overall pCR was documented; residual nodal disease was assigned when ypN was positive.
Outcomes were indeterminate for 114 patients without recorded ypN who had not achieved overall pCR; these patients were excluded, leaving 598 with determinable nodal outcomes.
Nexus ($n=45$) and Providence ($n=17$) were also excluded because all determinable outcomes were nodal pCR, so neither cohort met the inclusion rule in \Cref{sec:method_eval}.
This left 536 patients in six cohorts for the nodal pCR analysis.

The overall-pCR score was used without retraining or recalibration.
AUROCs were computed and pooled as for the primary endpoint (\Cref{sec:method_eval}).
Comparisons of nodal and overall pCR used the same 536 patients.

\subsection{Robustness to biopsy sampling}\label{sec:sampling_estimators}

\subsubsection{Intraclass correlation}
For the slide-consistency analysis (\Cref{fig:sampling_bias}a; \Cref{sec:sampling_robustness}), let $x_{ij}$ denote the prediction for slide $j$ of patient $i$, with $k_i$ slides per patient, $n$ patients, and $N=\sum_i k_i$ slides in total.
The one-way random-effects intraclass correlation coefficient ICC(1,1)~\citep{shrout1979intraclass} estimates the proportion of slide-level variance attributable to differences between patients:
\begin{equation}
\mathrm{ICC}(1,1) = \frac{\mathrm{MS}_{\mathrm{b}} - \mathrm{MS}_{\mathrm{w}}}{\mathrm{MS}_{\mathrm{b}} + (k_0 - 1)\,\mathrm{MS}_{\mathrm{w}}},
\label{eq:icc}
\end{equation}
where $\mathrm{MS}_{\mathrm{b}} = \tfrac{1}{n-1}\sum_i k_i (\bar{x}_i - \bar{x})^2$ and $\mathrm{MS}_{\mathrm{w}} = \tfrac{1}{N-n}\sum_i \sum_j (x_{ij} - \bar{x}_i)^2$ are the between- and within-patient mean squares.
Here, $\bar{x}_i$ is patient $i$'s mean prediction and $\bar{x}=N^{-1}\sum_i k_i\bar{x}_i$ is the slide-weighted grand mean.
Unequal slide counts are accounted for by the effective count $k_0 = \tfrac{1}{n-1}\!\left(N - \tfrac{\sum_i k_i^2}{N}\right)$, following the analysis-of-variance estimator for unequal group sizes of \citet{donner1980estimation}.
For equal slide counts between patients, $k_0=k$ and \Cref{eq:icc} reduces to the usual ICC(1,1) estimator.
The 95\% confidence interval was obtained by resampling patients, retaining all slides from each sampled patient together.

\subsubsection{Variance decomposition}
For the sampling-variance comparison (\Cref{fig:sampling_bias}c), we computed each patient $i$'s within-tumor variance $\sigma^2_{w,i}$, the variance of that patient's $k_i$ slide-level scores about their mean $\bar{x}_i$, and averaged these across the $n$ patients, $\overline{\sigma^2_w} = \tfrac{1}{n}\sum_{i=1}^{n}\sigma^2_{w,i}$, the estimator specified by \citet{biswas2025prospective}.
The sampling variance of an average over $k$ slides is then $W(k)=\overline{\sigma^2_w}/k$, by the $1/k$ variance-of-a-mean relation of \citet{bachtiary2006gene}.
Here $k$ is a hypothetical number of slides to be averaged, varied over $k=1,\dots,10$ following~\citet{biswas2025prospective}.
This $k$ differs from the observed slide count $k_i$ for patient $i$.
The between-tumor variance $B=\sigma^2_b$ was estimated as the variance of the $n$ per-patient mean scores $\bar{x}_i$, and the total is $T(k)=\sigma^2_b+W(k)$, so that $W/T$ is the fraction of total variance attributable to sampling.
The ratio $W(k)/T(k)$ was evaluated using the fixed estimates $\overline{\sigma^2_w}$ and $\sigma^2_b$, rather than an empirical resampling at each $k$.
Solving $W(k)/T(k)=0.15$ gives $k^{*}=(\overline{\sigma^2_w}/\sigma^2_b)\,(1-0.15)/0.15$, the number of slides at which the estimated sampling fraction reaches 0.15.

\subsection{Software and hardware}\label{sec:methods:eval:stats}
Analyses were performed in Python (version 3.10.19) and R (version 4.5.2).
Models were trained and run with \texttt{PyTorch} (version 2.7.1) and their hyperparameters tuned by Bayesian hyperparameter search using \texttt{BoTorch} (version 0.14.0); explainable boosting machines were fitted with \texttt{interpret} (version 0.7.6).
Independent component analysis and the evaluation metrics were computed with \texttt{scikit-learn} (version 1.6.1); bootstrap resampling used \texttt{NumPy} (version 1.26.4), and the Jonckheere--Terpstra test used \texttt{statsmodels.stats.nonparametric.jonckheere\_terpstra} (version 0.15.0).
The Firth-penalized per-cohort logistic regressions of \Cref{sec:adjusted_analysis} were fitted with the R package \texttt{logistf} (version 1.26.1).
Random-effects models were computed using the R package \texttt{meta} (version 8.2.1) and gene set enrichment analysis using \texttt{fgsea} (version 1.36.2); single-sample gene set enrichment scores were computed with the \texttt{ssgsea} function of the R package \texttt{corto}, and differential expression and the variance-stabilizing transformation with \texttt{DESeq2}.
The tissue segmentation underlying cTILs, LTR, and ITR (\Cref{sec:ctils}) used \texttt{TIAToolbox} (version 2.0.1 for the Protege cohort, version 1.6.0 for the rest).
Model training was performed on a single computing node equipped with eight NVIDIA H200 GPUs.

\end{document}